\documentclass{article}

\usepackage{PRIMEarxiv}

\usepackage[utf8]{inputenc} 
\usepackage[T1]{fontenc}    
\usepackage{hyperref}       
\usepackage{url}            
\usepackage{booktabs}       
\usepackage{amsfonts}       
\usepackage{nicefrac}       
\usepackage{microtype}      
\usepackage{lipsum}
\usepackage{fancyhdr}       
\usepackage{graphicx}       
\graphicspath{{media/}}     

\usepackage[numbers]{natbib}

\usepackage{multirow}
\usepackage{float}
\usepackage{subfig} 
\usepackage{amsmath}
\usepackage{mathtools}
\usepackage{bm}
\usepackage{multirow}
\usepackage{enumitem}
\usepackage[ruled,vlined,linesnumbered,noend]{algorithm2e}
\usepackage[table,xcdraw]{xcolor}

\usepackage{combelow}

\usepackage{xcolor}

\newcommand{\resultlowclr}{\cellcolor[HTML]{FEF9C3}}

\title{DANES: \underline{D}eep Neural Network Ensemble \underline{A}rchitecture for Social and Textual Context-aware Fake \underline{Ne}w\underline{s} Detection}

\author{
  Ciprian-Octavian Truic{\u{a}}$^1$, Elena-Simona Apostol$^1$, Panagiotis Karras$^2$ \\
  $^1$University Politehnica of Bucharest, Bucharest, Romania \\
  $^2$Aarhus University, Aarhus, Denmark \\
  \texttt{ciprian.truica@upb.ro, elena.apostol@upb.ro, panos@au.cs.dk}
}

\begin{document}
\maketitle

\begin{abstract}
The growing popularity of social media platforms has simplified the creation and distribution of news articles but also creates a conduit for spreading \emph{fake news}. In consequence, the need arises for effective context-aware fake news detection mechanisms, where the contextual information can be built either from the textual content of posts or from available social data (e.g., information about the users, reactions to posts, or the social network). In this paper, we propose DANES, a Deep Neural Network Ensemble Architecture for Social and Textual Context-aware Fake News Detection. DANES comprises a Text Branch for a textual content-based context and a Social Branch for the social context. These two branches are used to create a novel Network Embedding. 
Preliminary ablation results on 3 real-world datasets, i.e., BuzzFace, Twitter15, and Twitter16, are promising, with an accuracy that outperforms state-of-the-art solutions when employing both social and textual content features.
\end{abstract}


\maketitle

\section{Introduction}

The current media landscape has shifted from mass media to more personalized social media. This personalization of news creation and exchange brings advantages in terms of convenience and relevance, yet also increases the risk of misinformation \cite{Ruths2019} in the form of fake news, propaganda, and conspiracy theories, with detrimental consequences to society. The conventional definition of fake news is news articles that are \emph{verifiably} and usually \emph{intentionally} false \cite{Zhou2019}. Fake news on social media poses unique challenges considering that social media is a large-scale environment that provides users with the capacity for low-cost content creation and the option to post anonymously \cite{Shu2019}. Due to the importance of the problem, a variety of fake news detection (FND) solutions for social media have been developed \cite{Elhadad2019}.

Several solutions enhance the detection task using textual context \cite{Ilie2021,Ni2021,Pan2018,Truica2022,Truica2023} by employing contextual features obtained through the use of linguistic analysis or different embeddings. A few solutions consider the social context \cite{Bani2020,Shu2019beyond,Zhou2019network}, i.e., information about the social environment in which the news is disseminated, such as online users' interactions, their behaviors via posts and comments with respect to the news, and the network structure.

In this paper, we answer the following research questions (RQs):
\begin{itemize}
    \item[\textit{RQ1:}] Do textual and social context-enhanced solutions improve fake news detection?
    \item[\textit{RQ2:}] Do such solutions work well with limited training data?
\end{itemize}

To address these questions, we propose DANES, a Deep Neural Network Ensemble Architecture for Social and Textual Context-aware Fake News Detection.
The DANES model employs a Text Branch for textual content-based context and a Social Branch for social context. 
We concatenate the output of the Text Branch and Social Branch to obtain a novel Network Embedding. 
We use ablation testing to determine which layer combination offers the create best Network Embedding that improves the performance of the Fake News detection task.
For the Text Branch, we train six word embeddings:
\begin{itemize}
    \item[\textit{(1)}] \textsc{Word2Vec} \textsc{CBOW} (Continuous Bag-of-Words) \cite{Mikolov2013}; 
    \item[\textit{(2)}] \textsc{Word2Vec} \textsc{Skip-Gram} \cite{Mikolov2013}; 
    \item[\textit{(3)}] \textsc{FastText} \textsc{CBOW} \cite{Bojanowski2017}; 
    \item[\textit{(4)}] \textsc{FastText} \textsc{Skip-Gram} \cite{Bojanowski2017};
    \item[\textit{(5)}] \textsc{GloVe} \cite{Pennington2014}; and
    \item[\textit{(6)}] \textsc{Mittens} \cite{Dingwall2018}.
\end{itemize}

For the Social Branch, we did not consider Transformers embeddings, as they are used only pre-trained, and are extremely expensive to train or fine-tune.

For the Social Branch, we consider social networks platform dependent post and user information.
As post information, we study user engagement with the post, such as the number of reactions, comments, shares, etc., it received. 
As user information, we examine information regarding the number of followers, followees, friends, etc.
Thus, we propose a new network information embedding that takes into account the social network status of a user as well as its posts' influence.

We test DANES with different layer configurations with respect to each word embedding and neural network setup to determine the best performing model. The preliminary results obtained using ablation on the BuzzFace dataset \cite{Santia2018}, Twitter15, and Twitter16 \cite{Ma2017} are promising.
We attain the best-performing layer configuration when employing both social and textual features in classification.

The main contributions of this paper are as follows:
\begin{itemize}
    \item[\textit{C1:}] Propose DANES, a novel Deep Neural Network Ensemble Architecture for Social and Textual Context-aware Fake News Detection;
    \item[\textit{C2:}] Propose a new Network Embedding that considers both the Social and Textual Context of Social Networks;
    \item[\textit{C3:}] Perform extensive ablation tests on 3 different real-world datasets to show the efficiency of our proposed architecture DANES and Network Embedding and compare our results with the current state-of-the-art results.
\end{itemize}

The rest of this paper is structured as follows. 
Section~\ref{sec:sota} discusses current research related to context-based fake news detection. 
Section~\ref{sec:methodology} introduces the social and textual context-aware solution. 
Section~\ref{sec:results} presents the experimental results.
Section~\ref{sec:discussion} discusses the findings and hints at current challenges. 
Lastly, Section~\ref{sec:conclusions} presents the conclusions and future research directions.

\section{Related Work}\label{sec:sota}

Fake news detection (FND) solutions that use social context consider information about users' behavior \cite{Chowdhury2020,Shu2019beyond}, generated posts \cite{Vosoughi2018,Giachanou2019}, or the network \cite{Zhou2019network,Nguyen2020}.

\textbf{User-oriented} solutions analyze the behavior of users that create or spread fake news in contradistinction to those that spread real news. Such methods use credibility-based techniques that create links between news articles, posts, and users or publishers \cite{Chowdhury2020}. Thereby, news content originating from an unreliable user has more chance to be fake than one from credible users. 
In~\cite{Hamdi2020}, the authors propose a hybrid user-network approach on Twitter that considers both user characteristics (e.g., created at, name, favorites count, statuses count) and their social graph (i.e., followers/followees graph). For each user, a credibility score is calculated based on the credibility of the topics he/she tweets about, while also using the node2vec \cite{Grover2016node2vec} graph embedding to extract features from the social graph.

\textbf{Post-oriented} solutions focus on the opinions and emotions of users with regard to news, e.g., supporting or opposing. Such detection models consider that fake news can prompt tremendous controversial views among users, such as negative sentiments and denial \cite{Vosoughi2018}. User views are determined directly through explicit indications of emotion in messages, e.g., like, or indirectly by extracting this information from posts.
EmoCred \cite{Giachanou2019} is a post-oriented solution that incorporates emotions into a Long Short Term Memory (LSTM) neural network to distinguish between real and fake claims. The emotional signals are generated using three approaches: lexicon-based, intensity-based, and emotional intensity prediction using a neural network. Experiments on 2 \href{https://www.politifact.com/}{PolitiFact} datasets indicate the effectiveness of the solution.

\textbf{Network-oriented} solutions analyze information from various networks that users form on social media, e.g., interaction, friendship, and diffusion networks.
In~\cite{Zhou2019network}, the authors propose a network-oriented pattern-driven fake news detection solution that considers the friendship network and also provides an analysis of the fake news patterns in social networks. Factual News Graph (FANG) \cite{Nguyen2020} builds an interaction network using a graph representation that models social actors (user, source, news) and their interactions, while labeling each user-news link (e.g., neutral, negative, supportive, denial, or report).

We observe that the majority of context-oriented FND solutions focus on either social or textual context. The interaction between these two types of context has been scarcely studied. EchoFakeD \cite{Kaliyar2021} is one of the few works that considers both contexts. It uses textual context via word embeddings and social context (i.e., number of shares per article and the relationships between the users), training on 145 articles extracted from FakeNewsNet \cite{Shu2020}. However, the social context employed is limited.

\section{Methodology}\label{sec:methodology}

\begin{figure*}[!htbp]
\centering
\includegraphics[width=1\textwidth]{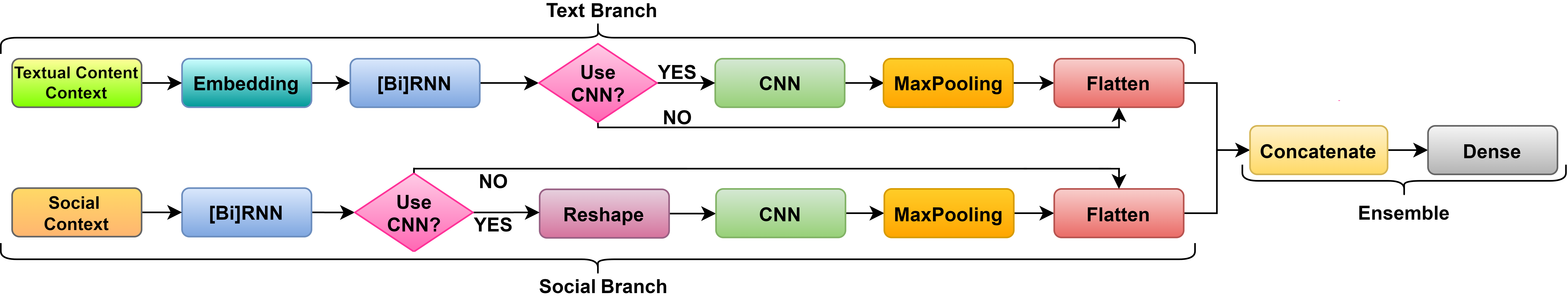}
\caption{DANES: Deep Neural Network Ensemble Architecture for Social and Textual Context-aware Fake News Detection}
\label{fig:architecture}
\end{figure*}

Figure~\ref{fig:architecture} presents the proposed Deep Neural Network Ensemble Architecture for Social and Textual Context-aware Fake News Detection (DANES). DANES uses the Text Branch for the textual context and the Social Branch for the social context information. The Content Branch employs different word embeddings, while the Social Branch uses an embedding that encodes the social context of the post (i.e., likes, loves, wows, hahas, sads, angrys, etc.). Both branches use recurrent or convolutional layers. For the baseline, we remove the Social Branch. We first use the social network context and the textual content to create embeddings and then apply DANES to determine the veracity of social media posts.

\subsection{Feature Representation}

\subsubsection{Text Preprocessing and Word Embeddings}

Prior to feeding the data to the proposed Deep Neural Network Ensemble Architecture, we preprocess the text to extract the token identifiers (ids), build a document-to-token-id matrix, and compute the word embeddings. The preprocessing involves the following steps used in literature for text analysis tasks~\cite{Truica2016,Truica2017,Truica2021}: 
\textit{(1)} expand contractions;
\textit{(2)} lemmatize each word to minimize the vocabulary;
\textit{(3)} remove punctuation;
\textit{(4)} tokenize the documents and assign each lemma an unique integer identifier (id) starting with id $1$; and
\textit{(5)} reconstruct the documents using the tokens' ids to obtain the 2-dimensional document-to-token-id matrix $D$.

The matrix $D$ is defined on $\mathbb{N}^{n \times k}$, where $n$ is the number of documents in the corpus and $k$ is the maximum size of a document in the corpus.
For documents that are smaller than the maximum size $k$, we use left zero padding, i.e., we add $0$ on the left side of the document to achieve the maximum length $k$.

Using the tokenized text, we train six word embeddings: 
\begin{itemize}
    \item[\textit{(1)}] \textsc{Word2Vec} \textsc{CBOW}~\cite{Mikolov2013}, a model that takes as input the context (the words on the left~$w_{-n}, \cdots, w_{-1}$  and the words on the right side~$w_{1}, \cdots, w_{n}$) of a word~$w$ to predict~$w$;
    \item[\textit{(2)}] \textsc{Word2Vec} \textsc{Skip-Gram}~\cite{Mikolov2013}, a model that takes as input the word~$w$ and tries to detect the context (the words on the left and right sides of~$w$);
    \item[\textit{(3)}] \textsc{FastText} \textsc{CBOW}~\cite{Bojanowski2017}, an extension of \textsc{Word2Vec} \textsc{CBOW} that uses character n-grams derived from the word~$w$, instead of~$w$;
    \item[\textit{(4)}] \textsc{FastText} \textsc{Skip-Gram}~\cite{Bojanowski2017} a similar extension of \textsc{Word2Vec} \textsc{Skip-Gram} that also uses the character n-grams derived from $w$, instead of $w$;
    \item[\textit{(5)}] \textsc{GloVe} model~\cite{Pennington2014}, which uses the word co-occurrence matrices obtained from the corpus to create vector representations;
    \item[\textit{(6)}] \textsc{Mittens} model~\cite{Dingwall2018}, an extension to \textsc{GloVe} for learning domain-specific representations.
\end{itemize}

For each word embedding model, we create a 2-dimensional matrix $W \in \mathbb{R}^{m \times s}$ that stores the embedding vector of each token id, where $m$ is the size of the vocabulary (i.e., the number of unique lemmas in the corpus plus the padding element) and $s$ is the size of the word embedding (i.e., the number of the embedding components).
For the padding element, the word embedding is represented as a vector containing only the value 0.

\subsubsection{Social Context Embedding}

We use social network-dependent metadata to build the social context such as the number of comments, shares, and reactions (e.g., likes, loves, wows, hahas, sads, angrys) for each post.
We also consider the post's author information for the platforms that provide such information (i.e., number of followers, followees, friends, public lists they belong to, etc.).
The social information is then vectorized as a feature vector $x_i=\{x_{ij} | i=1..n, j=1..f\}$, where $n$ is the number of posts and $f$ is the number of features taken into account.
Thus, each element $x_{ij}$ represents a normalized value for the reaction or user information.
We normalize each element of the feature vector $x_i$ using standard scaling $x'_{ij} = \frac{x_{ij} - \mu_{x_{j}}}{\sigma_{x_{j}}}$, with $\mu_{x_{j}}=\frac{1}{n}\sum_{i=1}^{n}x_{ij}$ and $\sigma_{x_{j}} = \sqrt{\frac{1}{n} \sum_{i=1}^{n}(x_{ij} -\mu_{x_{j}})}$ ($j=1..f$), to center the values independently on each feature by computing the relevant statistics on the samples in the training set. Using these normalized values, we create a social context embedding.
In the end, we obtain a 2-dimensional matrix $X \in \mathbb{R}^{n \times f}$ where each line of the matrix is the normalized feature vector $x'_i = \{ x'_{ij} | j=1..f \}$, i.e., $X = \{ x'_{i} | i=1..n\}$.

\subsection{DANES Architecture}

\subsubsection{Text Branch}

The input of DANES' Text Branch comprises the 2-dimensional matrices obtained during the Text Preprocessing and Word Embedding steps, i.e., the document-to-token-id matrix $D \in \mathbb{N}^{n \times k}$ and the word embedding matrix $W \in \mathbb{R}^{m \times s}$.
We use the Embedding layer to map each document's token id to its respective embedding.
Following the Embedding Layer, we employ either a Unidirectional Recurrent Neural Network (RNN) layer or a Bidirectional Recurrent Network (BiRNN) layer.
We call this layer [Bi]RNN, to denote that we discuss both RNN and BiRNN at the same time.
In the implementation for the [Bi]RNN layer, we employ either Long Short-Term Memory (LSTM) units or Gated Recurrent Units (GRU).
We note that the RNN is replaced by the actual cell employed by DANES in Section~\ref{sec:results}, i.e., LSTM or GRU.
As a unidirectional RNN only preserves information from inputs that pass through it using the hidden state in a feed-forward manner, we test whether the model's performance increases if the network receives the sequence of information in both directions: backward and forward.
After the [Bi]RNN layer, we use either a CNN (Convolutional Neural Network) with MaxPooling followed by a Flatten Layer, or just the Flatten Layer.
The CNN layer slides across the input text and detects all textual features, thus extracting complex features that are then used by the model to improve prediction.

\paragraph{\textbf{Text Content Embeddings using [Bi]RNN.}}

The first set of Text Content Embeddings uses only [Bi]RNN cells. 
For this set of embedding, the RNN cells are either LSTM or GRU.
Regardless of the word embedding used as input, we create 4 subtypes of Text Embeddings using [Bi]RNN: 
\begin{itemize}
    \item[(1)] Text Content Embedding using GRU: the word embedding is passed through a GRU layer and then a Flatten layer to obtain the final embedding.
    \item[(2)] Text Content Embedding using BiGRU: the word embedding is passed through a BiGRU layer and then a Flatten layer to obtain the final embedding.
    \item[(3)] Text Content Embedding using LSTM: the word embedding is passed through an LSTM layer and then a Flatten layer to obtain the final embedding.
    \item[(4)] Text Content Embedding using BiLSTM: the word embedding is passed through a BiLSTM layer and then a Flatten layer to obtain the final embedding.    
\end{itemize}

\paragraph{\textbf{Text Content Embeddings using [Bi]RNN-CNN.}}

The second set of Text Content Embeddings uses [Bi]RNN and CNN cells. 
The RNN cells are either LSTM or GRU.
Following the [Bi]RNN layer, there is a CNN layer followed by a MaxPool layer.
The final Text Content Embedding is obtained after the Flatten layer. 
Regardless of the word embedding used as input, we create 4 subtypes of Text Embeddings using [Bi]RNN-CNN: 
\begin{itemize}
    \item[(1)] Text Content Embedding using GRU-CNN: the word embedding is passed through a GRU layer, a CNN layer, a MaxPoling Layer, and a Flatten layer to obtain the final embedding.
    \item[(2)] Text Content Embedding using BiGRU-CNN: the word embedding is passed through a BiGRU layer, a CNN layer, a MaxPoling Layer, and a Flatten layer to obtain the final embedding.
    \item[(3)] Text Content Embedding using LSTM-CNN: the word embedding is passed through an LSTM layer, a CNN layer, a MaxPoling Layer, and a Flatten layer to obtain the final embedding.
    \item[(4)] Text Content Embedding using BiLSTM-CNN: the word embedding is passed through a BiLSTM layer, a CNN layer, a MaxPoling Layer, and a Flatten layer to obtain the final embedding.
\end{itemize}

\subsubsection{Social Branch}

DANES' Social Branch uses an input layer for the social context embedding.
That is the 2-dimensional matrix $X \in \mathbb{R}^{n \times f}$ representing the normalized feature vector of the platform-dependent social metadata.
Thereafter, we use an [Bi]RNN layer followed by either a Flatten Layer or Reshape, CNN, MaxPooling, and Flatten layers similar to the Content Branch.
The only modification is the Reshape Layer which is required to transform a 3-dimensional tensor (the output of the [Bi]RNN layer) to a 2-dimensional tensor as input to the CNN.
When using the CNN layer, the network may learn complex features extracted by the sliding window that passes over the 2-dimensional tensor representation and is then aggregated by the MaxPooling layer.

\paragraph{\textbf{Social Context Embeddings using [Bi]RNN.}}

The first set of Social Context Embeddings uses only [Bi]RNN cells. 
For this set of embeddings, the RNN cells are either LSTM or GRU.
We create 4 subtypes of Social Context Embeddings using [Bi]RNN: 
\begin{itemize}
    \item[(1)] Social Context Embedding using GRU: the word embedding is passed through a GRU layer and then a Flatten layer to obtain the final embedding.
    \item[(2)] Social Context Embedding using BiGRU: the word embedding is passed through a BiGRU layer and then a Flatten layer to obtain the final embedding.
    \item[(3)] Social Context Embedding using LSTM: the word embedding is passed through an LSTM layer and then a Flatten layer to obtain the final embedding.
    \item[(4)] Social Context Embedding using BiLSTM: the word embedding is passed through a BiLSTM layer and then a Flatten layer to obtain the final embedding.    
\end{itemize}

\paragraph{\textbf{Social Context Embeddings using [Bi]RNN-CNN.}}

The second set of Social Context Embeddings uses [Bi]RNN and CNN cells. 
The RNN cells are either LSTM or GRU.
Following the [Bi]RNN layer, there is a CNN layer followed by a MaxPool layer.
The final Text Embedding is obtained after the Flatten layer. 
We create 4 subtypes of Social Context Embeddings using [Bi]RNN-CNN: 
\begin{itemize}   
    \item[(1)] Social Context Embedding using GRU-CNN: the word embedding is passed through a GRU layer, a CNN layer, a MaxPoling Layer, and a Flatten layer to obtain the final embedding.
    \item[(2)] Social Context Embedding using BiGRU-CNN: the word embedding is passed through a BiGRU layer, a CNN layer, a MaxPoling Layer, and a Flatten layer to obtain the final embedding.
    \item[(3)] Social Context Embedding using LSTM-CNN: the word embedding is passed through an LSTM layer, a CNN layer, a MaxPoling Layer, and a Flatten layer to obtain the final embedding.
    \item[(4)] Social Context Embedding using BiLSTM-CNN: the word embedding is passed through a BiLSTM layer, a CNN layer, a MaxPoling Layer, and a Flatten layer to obtain the final embedding.
\end{itemize}

\subsubsection{Ensemble}

DANES' Ensemble concatenates the output of the Content and Social Branches into one tensor, which is passed to a Dense layer used for classification.
During the concatenation, we create a new \textit{Network Embedding} that takes into account both the textual content and the social context, i.e., it combines the Text Content Embedding with the Social Context Embedding.
This novel Network Embedding is then used by the Dense layer for classification. 

The Dense layer employs Perceptron units to determine the veracity of a post using a softmax activation function.

\section{Results}\label{sec:results}

In this section, we present the datasets used for the experiments, the experimental setup and implementation, and the classification results.
We conclude this section with a comparison between our results and the results obtained by state-of-the-art models.

\subsection{Dataset}

We perform our experiments on 3 datasets to see if the models generalize well: \href{https://github.com/BuzzFeedNews/2016-10-facebook-fact-check}{BuzzFace} \cite{Santia2018}, \href{https://github.com/majingCUHK/Rumor_RvNN}{Twitter15}, and \href{https://github.com/majingCUHK/Rumor_RvNN}{Twitter16} \cite{Ma2017}.
Table \ref{tab:data_stats} presents the statistics for these datasets.
The in-depth analysis of the BuzzFace dataset is presented in \cite{Santia2018}, while for the two Twitter datasets is presented in \cite{Ma2017}.

\begin{table}[!htbp]
\centering
\caption{Datasets statistics}\label{tab:data_stats}
\begin{tabular}{|l|r|c|c|l|l|}
\hline
\textbf{Dataset}                                                & \multicolumn{1}{c|}{\textbf{\begin{tabular}[c]{@{}c@{}}Corpus\\ Size\end{tabular}}} & \textbf{\begin{tabular}[c]{@{}c@{}}Document\\ Size\end{tabular}} & \textbf{\begin{tabular}[c]{@{}c@{}}Vocabulary\\ Size\end{tabular}} & \textbf{\begin{tabular}[c]{@{}l@{}}Classes labels\\ (Class: encoding)\end{tabular}}                                             & \textbf{Network features}                                                                                                                                                          \\ \hline
BuzzFace & 2\,282                                                                              & 76                                                               & 5\,556                                                             & \begin{tabular}[c]{@{}l@{}}mostly true: 0\\ mixture of \\true and false: 1\\ no factual content: 2\\ mostly false: 3\end{tabular} & \begin{tabular}[c]{@{}l@{}}comments count\\ shares count\\ likes count\\ love emoji count\\ wow emoji count\\ haha emoji count\\ sad  emoji count\\ angry emoji count\end{tabular} \\ \hline
\multirow{3}{*}{Twitter15}                                      & \multirow{3}{*}{1\,490}                                                             & \multirow{3}{*}{52}                                              & \multirow{3}{*}{4\,258}                                            & \multirow{6}{*}{\begin{tabular}[c]{@{}l@{}}true: 0\\ false: 1\\ unverified: 2\\ non-rumor: 3\end{tabular}}                      & \multirow{6}{*}{\begin{tabular}[c]{@{}l@{}}likes count\\ retweet count\\ user followers count\\ user friends count\\ user lists count\\ user favorites count\end{tabular}}      \\
                                                                &                                                                                     &                                                                  &                                                                    &                                                                                                                                 &                                                                                                                                                                                    \\
                                                                &                                                                                     &                                                                  &                                                                    &                                                                                                                                 &                                                                                                                                                                                    \\ \cline{1-4}
\multirow{3}{*}{Twitter16}                                      & \multirow{3}{*}{818}                                                                & \multirow{3}{*}{27}                                              & \multirow{3}{*}{2\,795}                                            &                                                                                                                                 &                                                                                                                                                                                    \\
                                                                &                                                                                     &                                                                  &                                                                    &                                                                                                                                 &                                                                                                                                                                                    \\
                                                                &                                                                                     &                                                                  &                                                                    &                                                                                                                                 &                                                                                                                                                                                    \\ \hline
\end{tabular}
\end{table}

The BuzzFace dataset \cite{Santia2018}, an extension of BuzzFeed News dataset \cite{Horne2017,Silverman2016}, consists of 2\,282 labeled and verified Facebook posts along with the corresponding Facebook engagement numbers (i.e., shares, comments, reactions, etc.).
Each post is labeled with one of the following 4 classes ``mostly true'', ``mixture of true and false'', ``mostly false'', or ``no factual content'' for posts that are satirical or opinion-driven.
For the social context embedding, we consider the number of shares, likes, and each individual reaction (i.e., love, wow, haha, sad, and angry emojis).
After preprocessing, the maximum document length is 76 and the vocabulary contains 5\,556 lemmas.

Twitter15 contains 1\,490 tweets, while Twitter16 contains 818 tweets.
Each tweet is labeled with one of the following 4 classes ``true'', ``false'', ``unverified'', or ``non-rumor''.
For the social context, we considered tweet related data (e.g., number of retweets, number of likes) as well as user related information (i.e., number of followers, number of friends, etc.)
For Twitter15, after preprocessing, the maximum document length is 52 for Twitter and the vocabulary contains 4\,258 lemmas.
While for Twitter16, after preprocessing, the maximum document length is 27 for Twitter and the vocabulary contains 2\,795 lemmas.

\subsection{Experimental Setup and Implementation}

To test whether performance improves depending on the word embedding used, we train six word embeddings for each dataset to evaluate the performance of the proposed ensemble architecture:
\textsc{Word2Vec} CBOW and \textsc{Skip-Gram}, \textsc{FastText} CBOW and \textsc{Skip-Gram}, \textsc{GloVe}, and \textsc{Mittens}. For each embedding, we use the following training configuration: window size 10, number of components 128, training over 100 epochs, and learning rate 0.05.

The proposed DANES model uses the LSTM implementation in \cite{Hochreiter1997} and the GRU implementation in \cite{Cho2014} without altering them. Each RNN layer has 128 units, while the BiRNN layers contain 256 units each. The [Bi]RNN layers use a feed-forward dropout rate of 0.2 and a recurrent dropout rate of 0.2 to prevent overfitting.
For the CNN with the BuzzFace dataset, we use a kernel size of 64 when employing the RNN and 128 for the BiRNN as the length of the documents is large.
For the CNN with the Twitter15 dataset, we use a kernel size of 32 when employing the RNN and 64 for the BiRNN because the length of the documents is medium.
Finally, for the CNN with the Twitter16 dataset, we use a kernel size of 16 when employing the RNN and 32 for the BiRNN because the length of the documents is small.
The hyperparameters' values were selected based on the work results presented in~\cite{Truica2022,Ilie2021}.
Lastly, the dense layer uses 4 units equal to the number of classes and a softmax activation function.

We run the experiments on an NVIDIA\textsuperscript{\textregistered} DGX Station\textsuperscript{\texttrademark} with an Ubuntu 20.04 LTS operating system. 
The station has the following configuration: 4 NVIDIA\textsuperscript{\textregistered} Tesla\textsuperscript{\textregistered} V100-DGXS GPU with 32GB VRAM each, Intel\textsuperscript{\textregistered} Xeon\textsuperscript{\textregistered} E5-2698 v4 @ 2.20GHz CPU with 16 Cores, 32 Threads, and 50MB Intel\textsuperscript{\textregistered} Smart Cache, and 264GB RAM.
The NVidia\textsuperscript{\textregistered} Driver version is 470.57.02 and the CUDA\textsuperscript{\textregistered} version is 11.4.

We implemented the preprocessing elements of DANES in Python version 3.9 using \href{https://spacy.io/}{SpaCy} \cite{SpaCy} for text preprocessing.
To train word embeddings on each of the 3 employed datasets, we used the following packages: 
\href{https://radimrehurek.com/gensim/}{gensim} \cite{Rehurek2010} for the the \textsc{Word2Vec} and \textsc{FastText} models, \href{https://github.com/maciejkula/glove-python}{pyglove} \cite{PyGloVe} for the \textsc{GloVe} model, and \href{https://github.com/roamanalytics/mittens}{Mittens} \cite{Dingwall2018} for the \textsc{Mittens} model. We implemented the neural elements of DANES using \href{https://www.tensorflow.org/}{TensorFlow} \cite{tensorflow2015-whitepaper} and \href{https://keras.io/}{Keras} \cite{chollet2015keras}.
The code is freely available on GitHub at \url{https://github.com/DS4AI-UPB/DANES}.

\subsection{Classification Results}

To perform testing, we split the dataset using 70\% as the training set and 30\% as the testing set. We also use 20\% of the training set for validation. When splitting, we shuffle the data but keep the same samples' percentage of each target class as the complete set.

To determine which architectural choice as well as input word embedding influences the performance of the Network Embedding and model overall, we perform 10 runs and measure the mean of each evaluation metric $\pm$ its standard deviation.
We also register the runtime required for the model to converge.
To avoid overfitting and improve model generalization, we use for each layer both feed-forward and recurrent dropout rates of 0.2 and an early stopping mechanism that monitors the validation loss over 20 epochs and stops the training if the loss is no longer decreasing.

As a baseline for our experiment w.r.t each word embedding, we consider the model that uses only the Text Branch without the Social Branch.
Note that when both the Text Branch and the Social Branch are used, it means that the model uses the novel Network Embedding.
At the end of this section, we will compare our results with the results obtained by state-of-the-art models.

\subsubsection{BuzzFace results}
Tables \ref{tab:results_gru_buzzface} and \ref{tab:results_lstm_buzzface} present
the ablation results when using [Bi]GRU and [Bi]LSTM as the [Bi]RNN element in our Network Embedding architecture.
Regardless of whether we employ [Bi]GRU or [Bi]LSTM, we obtain the overall best accuracy with the \textsc{Mittens} word embeddings along with the social context embedding. This embedding puts more weight on the domain-specific terms, thus DANES learns to discriminate based on these complex and specific features.

For the case when we use GRU as the RNN variant, we obtain the highest accuracy when using only one GRU layer for the Text Branch and the GRU and CNN for the Social Branch. With the GRU configuration, we obtain an accuracy of 79.62\%, while when replacing the GRU with a BiGRU layer we obtain an accuracy of 79.34\%.
When using LSTM cells, we obtain the best result by employing the same model configuration as for the GRU, i.e., [Bi]LSTM for Text Branch and [Bi]LSTM and CNN for the Social Branch.
The models that uses only LSTM obtains an accuracy of 79.74\%, while the models that use BiLSTM obtain an accuracy of 79.63\%.


\begin{table*}[!htbp]
\centering
\caption{Ablation testing on the BuzzFace dataset~\cite{Santia2018} using [Bi]GRU as [Bi]RNN units (bold marks the best accuracy w.r.t. word embedding)}
\label{tab:results_gru_buzzface}
\resizebox{1\textwidth}{!}{%
\begin{tabular}{|c|cc|cc|llll|llll|}
\hline
\multicolumn{1}{|l|}{\multirow{3}{*}{\textbf{\begin{tabular}[c]{@{}c@{}}Word\\ Embedding\end{tabular}}}} & \multicolumn{4}{c|}{\textbf{Network Embedding}}  & \multicolumn{4}{c|}{\multirow{2}{*}{\textbf{GRU Cell}}}   & \multicolumn{4}{c|}{\multirow{2}{*}{\textbf{BiGRU Cell}}}                                                                                                                                           \\ \cline{2-5}
\multicolumn{1}{|l|}{}                               & \multicolumn{2}{c|}{\textbf{Text Branch}}     & \multicolumn{2}{c|}{\textbf{Social Branch}}     & \multicolumn{4}{c|}{} & \multicolumn{4}{c|}{}                                                                                                                                       \\ \cline{2-13} 
\multicolumn{1}{|l|}{}                                                                                                                 & \multicolumn{1}{c|}{\textbf{RNN}} & \textbf{CNN} & \multicolumn{1}{c|}{\textbf{RNN}} & \textbf{CNN} & \multicolumn{1}{c|}{\textbf{Accuracy}}         & \multicolumn{1}{c|}{\textbf{Precision}} & \multicolumn{1}{c|}{\textbf{Recall}}  & \multicolumn{1}{c|}{\textbf{Runtime(s)}} & \multicolumn{1}{c|}{\textbf{Accuracy}}         & \multicolumn{1}{c|}{\textbf{Precision}} & \multicolumn{1}{c|}{\textbf{Recall}}  & \multicolumn{1}{c|}{\textbf{Runtime(s)}} \\ \hline
\multirow{6}{*}{\begin{tabular}[c]{@{}c@{}}\textsc{\textbf{Word2Vec}}\\ \textbf{CBOW}\end{tabular}}               & \multicolumn{1}{c|}{\checkmark}   &              & \multicolumn{2}{c|}{Branch N/A}                  & \multicolumn{1}{l|}{78.98 $\pm$ 0.99}          & \multicolumn{1}{l|}{70.81 $\pm$ 3.01}   & \multicolumn{1}{l|}{78.98 $\pm$ 0.99} & 14.33 $\pm$ 1.12                              & \multicolumn{1}{l|}{79.04 $\pm$ 0.84}          & \multicolumn{1}{l|}{71.02 $\pm$ 2.63}   & \multicolumn{1}{l|}{79.04 $\pm$ 0.84} & 22.45 $\pm$ 1.75                              \\ \cline{2-13} 
                                                                                                                  & \multicolumn{1}{c|}{\checkmark}   & \checkmark   & \multicolumn{2}{c|}{Branch N/A}                  & \multicolumn{1}{l|}{78.78 $\pm$ 0.91}          & \multicolumn{1}{l|}{68.98 $\pm$ 2.50}   & \multicolumn{1}{l|}{78.78 $\pm$ 0.91} & 13.77 $\pm$ 1.04                              & \multicolumn{1}{l|}{77.76 $\pm$ 0.96}          & \multicolumn{1}{l|}{69.40 $\pm$ 1.25}   & \multicolumn{1}{l|}{77.76 $\pm$ 0.96} & 20.96 $\pm$ 1.44                              \\ \cline{2-13} 
                                                                                                                  & \multicolumn{1}{c|}{\checkmark}   &              & \multicolumn{1}{c|}{\checkmark}   &              & \multicolumn{1}{l|}{78.98 $\pm$ 0.92}          & \multicolumn{1}{l|}{71.64 $\pm$ 2.96}   & \multicolumn{1}{l|}{78.98 $\pm$ 0.92} & 16.81 $\pm$ 1.21                              & \multicolumn{1}{l|}{78.64 $\pm$ 1.08}          & \multicolumn{1}{l|}{70.77 $\pm$ 1.90}   & \multicolumn{1}{l|}{78.64 $\pm$ 1.08} & 27.93 $\pm$ 0.87                              \\ \cline{2-13} 
                                                                                                                  & \multicolumn{1}{c|}{\checkmark}   &              & \multicolumn{1}{c|}{\checkmark}   & \checkmark   & \multicolumn{1}{l|}{78.90 $\pm$ 1.36}          & \multicolumn{1}{l|}{72.31 $\pm$ 2.39}   & \multicolumn{1}{l|}{78.90 $\pm$ 1.36} & 17.13 $\pm$ 1.75                              & \multicolumn{1}{l|}{79.04 $\pm$ 1.19}          & \multicolumn{1}{l|}{71.49 $\pm$ 2.85}   & \multicolumn{1}{l|}{79.04 $\pm$ 1.19} & 27.98 $\pm$ 0.89                              \\ \cline{2-13} 
                                                                                                                  & \multicolumn{1}{c|}{\checkmark}   & \checkmark   & \multicolumn{1}{c|}{\checkmark}   &              & \multicolumn{1}{l|}{78.78 $\pm$ 0.88}          & \multicolumn{1}{l|}{69.76 $\pm$ 1.91}   & \multicolumn{1}{l|}{78.78 $\pm$ 0.88} & 15.87 $\pm$ 0.25                              & \multicolumn{1}{l|}{78.08 $\pm$ 1.25}          & \multicolumn{1}{l|}{70.24 $\pm$ 2.22}   & \multicolumn{1}{l|}{78.08 $\pm$ 1.25} & 26.88 $\pm$ 1.19                              \\ \cline{2-13} 
                                                                                                                  & \multicolumn{1}{c|}{\checkmark}   & \checkmark   & \multicolumn{1}{c|}{\checkmark}   & \checkmark   & \multicolumn{1}{l|}{78.68 $\pm$ 0.95}          & \multicolumn{1}{l|}{69.38 $\pm$ 2.30}   & \multicolumn{1}{l|}{78.68 $\pm$ 0.95} & 17.21 $\pm$ 1.71                              & \multicolumn{1}{l|}{78.80 $\pm$ 1.04}          & \multicolumn{1}{l|}{70.00 $\pm$ 1.78}   & \multicolumn{1}{l|}{78.80 $\pm$ 1.04} & 29.76 $\pm$ 1.34                              \\ \hline
\multirow{6}{*}{\begin{tabular}[c]{@{}c@{}}\textsc{\textbf{Word2Vec}}\\ \textsc{\textbf{Skip-Gran}}\end{tabular}} & \multicolumn{1}{c|}{\checkmark}   &              & \multicolumn{2}{c|}{Branch N/A}                  & \multicolumn{1}{l|}{78.02 $\pm$ 0.86}          & \multicolumn{1}{l|}{71.54 $\pm$ 1.84}   & \multicolumn{1}{l|}{78.02 $\pm$ 0.86} & 17.68 $\pm$ 1.65                              & \multicolumn{1}{l|}{78.52 $\pm$ 0.98}          & \multicolumn{1}{l|}{73.27 $\pm$ 2.49}   & \multicolumn{1}{l|}{78.52 $\pm$ 0.98} & 27.75 $\pm$ 2.89                              \\ \cline{2-13} 
                                                                                                                  & \multicolumn{1}{c|}{\checkmark}   & \checkmark   & \multicolumn{2}{c|}{Branch N/A}                  & \multicolumn{1}{l|}{77.86 $\pm$ 0.71}          & \multicolumn{1}{l|}{69.50 $\pm$ 1.63}   & \multicolumn{1}{l|}{77.86 $\pm$ 0.71} & 15.89 $\pm$ 1.42                              & \multicolumn{1}{l|}{78.16 $\pm$ 0.93}          & \multicolumn{1}{l|}{70.45 $\pm$ 1.44}   & \multicolumn{1}{l|}{78.16 $\pm$ 0.93} & 25.51 $\pm$ 2.02                              \\ \cline{2-13} 
                                                                                                                  & \multicolumn{1}{c|}{\checkmark}   &              & \multicolumn{1}{c|}{\checkmark}   &              & \multicolumn{1}{l|}{78.56 $\pm$ 1.31}          & \multicolumn{1}{l|}{72.95 $\pm$ 2.04}   & \multicolumn{1}{l|}{78.56 $\pm$ 1.31} & 20.05 $\pm$ 1.47                              & \multicolumn{1}{l|}{78.68 $\pm$ 0.93}          & \multicolumn{1}{l|}{72.48 $\pm$ 1.76}   & \multicolumn{1}{l|}{78.68 $\pm$ 0.93} & 32.49 $\pm$ 0.64                              \\ \cline{2-13} 
                                                                                                                  & \multicolumn{1}{c|}{\checkmark}   &              & \multicolumn{1}{c|}{\checkmark}   & \checkmark   & \multicolumn{1}{l|}{78.60 $\pm$ 1.02}          & \multicolumn{1}{l|}{73.01 $\pm$ 1.59}   & \multicolumn{1}{l|}{78.60 $\pm$ 1.02} & 20.38 $\pm$ 1.71                              & \multicolumn{1}{l|}{78.34 $\pm$ 1.47}          & \multicolumn{1}{l|}{72.23 $\pm$ 2.35}   & \multicolumn{1}{l|}{78.34 $\pm$ 1.47} & 32.88 $\pm$ 2.43                              \\ \cline{2-13} 
                                                                                                                  & \multicolumn{1}{c|}{\checkmark}   & \checkmark   & \multicolumn{1}{c|}{\checkmark}   &              & \multicolumn{1}{l|}{78.56 $\pm$ 1.00}          & \multicolumn{1}{l|}{70.84 $\pm$ 2.15}   & \multicolumn{1}{l|}{78.56 $\pm$ 1.00} & 18.56 $\pm$ 1.66                              & \multicolumn{1}{l|}{78.42 $\pm$ 1.20}          & \multicolumn{1}{l|}{71.26 $\pm$ 1.80}   & \multicolumn{1}{l|}{78.42 $\pm$ 1.20} & 32.06 $\pm$ 3.18                              \\ \cline{2-13} 
                                                                                                                  & \multicolumn{1}{c|}{\checkmark}   & \checkmark   & \multicolumn{1}{c|}{\checkmark}   & \checkmark   & \multicolumn{1}{l|}{77.82 $\pm$ 1.25}          & \multicolumn{1}{l|}{70.29 $\pm$ 2.12}   & \multicolumn{1}{l|}{77.82 $\pm$ 1.25} & 20.61 $\pm$ 1.35                              & \multicolumn{1}{l|}{78.26 $\pm$ 1.34}          & \multicolumn{1}{l|}{70.69 $\pm$ 1.92}   & \multicolumn{1}{l|}{78.26 $\pm$ 1.34} & 37.49 $\pm$ 2.22                              \\ \hline
\multirow{6}{*}{\begin{tabular}[c]{@{}c@{}}\textsc{\textbf{FastText}}\\ \textbf{CBOW}\end{tabular}}               & \multicolumn{1}{c|}{\checkmark}   &              & \multicolumn{2}{c|}{Branch N/A}                  & \multicolumn{1}{l|}{79.04 $\pm$ 0.47}          & \multicolumn{1}{l|}{70.90 $\pm$ 1.81}   & \multicolumn{1}{l|}{79.04 $\pm$ 0.47} & 13.99 $\pm$ 1.27                              & \multicolumn{1}{l|}{79.12 $\pm$ 0.82}          & \multicolumn{1}{l|}{72.50 $\pm$ 2.41}   & \multicolumn{1}{l|}{79.12 $\pm$ 0.82} & 22.58 $\pm$ 0.44                              \\ \cline{2-13} 
                                                                                                                  & \multicolumn{1}{c|}{\checkmark}   & \checkmark   & \multicolumn{2}{c|}{Branch N/A}                  & \multicolumn{1}{l|}{78.04 $\pm$ 0.94}          & \multicolumn{1}{l|}{67.61 $\pm$ 1.13}   & \multicolumn{1}{l|}{78.04 $\pm$ 0.94} & 13.58 $\pm$ 1.14                              & \multicolumn{1}{l|}{77.76 $\pm$ 1.55}          & \multicolumn{1}{l|}{69.98 $\pm$ 2.08}   & \multicolumn{1}{l|}{77.76 $\pm$ 1.55} & 20.78 $\pm$ 1.07                              \\ \cline{2-13} 
                                                                                                                  & \multicolumn{1}{c|}{\checkmark}   &              & \multicolumn{1}{c|}{\checkmark}   &              & \multicolumn{1}{l|}{79.26 $\pm$ 0.94}          & \multicolumn{1}{l|}{71.30 $\pm$ 1.63}   & \multicolumn{1}{l|}{79.26 $\pm$ 0.94} & 16.59 $\pm$ 1.35                              & \multicolumn{1}{l|}{79.14 $\pm$ 0.87}          & \multicolumn{1}{l|}{72.20 $\pm$ 2.88}   & \multicolumn{1}{l|}{79.14 $\pm$ 0.87} & 27.83 $\pm$ 0.32                              \\ \cline{2-13} 
                                                                                                                  & \multicolumn{1}{c|}{\checkmark}   &              & \multicolumn{1}{c|}{\checkmark}   & \checkmark   & \multicolumn{1}{l|}{79.44 $\pm$ 0.61}          & \multicolumn{1}{l|}{71.47 $\pm$ 2.51}   & \multicolumn{1}{l|}{79.44 $\pm$ 0.61} & 15.98 $\pm$ 0.16                              & \multicolumn{1}{l|}{79.20 $\pm$ 0.74}          & \multicolumn{1}{l|}{71.49 $\pm$ 2.71}   & \multicolumn{1}{l|}{79.20 $\pm$ 0.74} & 27.99 $\pm$ 0.40                              \\ \cline{2-13} 
                                                                                                                  & \multicolumn{1}{c|}{\checkmark}   & \checkmark   & \multicolumn{1}{c|}{\checkmark}   &              & \multicolumn{1}{l|}{79.10 $\pm$ 0.60}          & \multicolumn{1}{l|}{69.74 $\pm$ 1.97}   & \multicolumn{1}{l|}{79.10 $\pm$ 0.60} & 15.72 $\pm$ 0.82                              & \multicolumn{1}{l|}{78.48 $\pm$ 0.69}          & \multicolumn{1}{l|}{69.77 $\pm$ 1.03}   & \multicolumn{1}{l|}{78.48 $\pm$ 0.69} & 26.90 $\pm$ 0.96                              \\ \cline{2-13} 
                                                                                                                  & \multicolumn{1}{c|}{\checkmark}   & \checkmark   & \multicolumn{1}{c|}{\checkmark}   & \checkmark   & \multicolumn{1}{l|}{78.66 $\pm$ 0.96}          & \multicolumn{1}{l|}{69.10 $\pm$ 1.53}   & \multicolumn{1}{l|}{78.66 $\pm$ 0.96} & 17.78 $\pm$ 1.42                              & \multicolumn{1}{l|}{78.90 $\pm$ 0.84}          & \multicolumn{1}{l|}{69.93 $\pm$ 1.56}   & \multicolumn{1}{l|}{78.90 $\pm$ 0.84} & 30.94 $\pm$ 2.55                              \\ \hline
\multirow{6}{*}{\begin{tabular}[c]{@{}c@{}}\textsc{\textbf{FastText}}\\ \textsc{\textbf{Skip-Gran}}\end{tabular}} & \multicolumn{1}{c|}{\checkmark}   &              & \multicolumn{2}{c|}{Branch N/A}                  & \multicolumn{1}{l|}{78.28 $\pm$ 1.18}          & \multicolumn{1}{l|}{71.32 $\pm$ 2.36}   & \multicolumn{1}{l|}{78.28 $\pm$ 1.18} & 17.43 $\pm$ 1.53                              & \multicolumn{1}{l|}{78.38 $\pm$ 0.54}          & \multicolumn{1}{l|}{71.91 $\pm$ 1.52}   & \multicolumn{1}{l|}{78.38 $\pm$ 0.54} & 28.05 $\pm$ 2.87                              \\ \cline{2-13} 
                                                                                                                  & \multicolumn{1}{c|}{\checkmark}   & \checkmark   & \multicolumn{2}{c|}{Branch N/A}                  & \multicolumn{1}{l|}{78.02 $\pm$ 1.21}          & \multicolumn{1}{l|}{70.23 $\pm$ 2.34}   & \multicolumn{1}{l|}{78.02 $\pm$ 1.21} & 16.10 $\pm$ 1.03                              & \multicolumn{1}{l|}{78.94 $\pm$ 0.81}          & \multicolumn{1}{l|}{71.08 $\pm$ 2.71}   & \multicolumn{1}{l|}{78.94 $\pm$ 0.81} & 24.24 $\pm$ 1.83                              \\ \cline{2-13} 
                                                                                                                  & \multicolumn{1}{c|}{\checkmark}   &              & \multicolumn{1}{c|}{\checkmark}   &              & \multicolumn{1}{l|}{78.74 $\pm$ 0.76}          & \multicolumn{1}{l|}{73.02 $\pm$ 2.68}   & \multicolumn{1}{l|}{78.74 $\pm$ 0.76} & 19.62 $\pm$ 1.32                              & \multicolumn{1}{l|}{78.84 $\pm$ 0.97}          & \multicolumn{1}{l|}{72.37 $\pm$ 2.43}   & \multicolumn{1}{l|}{78.84 $\pm$ 0.97} & 32.52 $\pm$ 2.69                              \\ \cline{2-13} 
                                                                                                                  & \multicolumn{1}{c|}{\checkmark}   &              & \multicolumn{1}{c|}{\checkmark}   & \checkmark   & \multicolumn{1}{l|}{78.34 $\pm$ 0.90}          & \multicolumn{1}{l|}{72.76 $\pm$ 2.84}   & \multicolumn{1}{l|}{78.34 $\pm$ 0.90} & 20.08 $\pm$ 1.52                              & \multicolumn{1}{l|}{78.78 $\pm$ 1.01}          & \multicolumn{1}{l|}{72.92 $\pm$ 3.42}   & \multicolumn{1}{l|}{78.78 $\pm$ 1.01} & 33.46 $\pm$ 2.82                              \\ \cline{2-13} 
                                                                                                                  & \multicolumn{1}{c|}{\checkmark}   & \checkmark   & \multicolumn{1}{c|}{\checkmark}   &              & \multicolumn{1}{l|}{78.72 $\pm$ 0.88}          & \multicolumn{1}{l|}{69.80 $\pm$ 1.46}   & \multicolumn{1}{l|}{78.72 $\pm$ 0.88} & 17.91 $\pm$ 1.86                              & \multicolumn{1}{l|}{78.68 $\pm$ 1.12}          & \multicolumn{1}{l|}{70.64 $\pm$ 3.52}   & \multicolumn{1}{l|}{78.68 $\pm$ 1.12} & 29.99 $\pm$ 2.67                              \\ \cline{2-13} 
                                                                                                                  & \multicolumn{1}{c|}{\checkmark}   & \checkmark   & \multicolumn{1}{c|}{\checkmark}   & \checkmark   & \multicolumn{1}{l|}{78.54 $\pm$ 1.28}          & \multicolumn{1}{l|}{70.06 $\pm$ 1.63}   & \multicolumn{1}{l|}{78.54 $\pm$ 1.28} & 19.61 $\pm$ 1.52                              & \multicolumn{1}{l|}{78.78 $\pm$ 1.14}          & \multicolumn{1}{l|}{70.47 $\pm$ 2.24}   & \multicolumn{1}{l|}{78.78 $\pm$ 1.14} & 35.49 $\pm$ 3.24                              \\ \hline
\multirow{6}{*}{\textsc{\textbf{GloVe}}}                                                                          & \multicolumn{1}{c|}{\checkmark}   &              & \multicolumn{2}{c|}{Branch N/A}                  & \multicolumn{1}{l|}{78.34 $\pm$ 1.17}          & \multicolumn{1}{l|}{72.10 $\pm$ 1.95}   & \multicolumn{1}{l|}{78.34 $\pm$ 1.17} & 19.29 $\pm$ 1.54                              & \multicolumn{1}{l|}{78.18 $\pm$ 1.41}          & \multicolumn{1}{l|}{72.17 $\pm$ 1.90}   & \multicolumn{1}{l|}{78.18 $\pm$ 1.41} & 31.27 $\pm$ 2.28                              \\ \cline{2-13} 
                                                                                                                  & \multicolumn{1}{c|}{\checkmark}   & \checkmark   & \multicolumn{2}{c|}{Branch N/A}                  & \multicolumn{1}{l|}{78.88 $\pm$ 1.04}          & \multicolumn{1}{l|}{71.28 $\pm$ 1.53}   & \multicolumn{1}{l|}{78.88 $\pm$ 1.04} & 17.45 $\pm$ 1.85                              & \multicolumn{1}{l|}{77.86 $\pm$ 1.33}          & \multicolumn{1}{l|}{71.25 $\pm$ 2.80}   & \multicolumn{1}{l|}{77.86 $\pm$ 1.33} & 25.96 $\pm$ 2.19                              \\ \cline{2-13} 
                                                                                                                  & \multicolumn{1}{c|}{\checkmark}   &              & \multicolumn{1}{c|}{\checkmark}   &              & \multicolumn{1}{l|}{78.76 $\pm$ 0.95}          & \multicolumn{1}{l|}{73.34 $\pm$ 1.44}   & \multicolumn{1}{l|}{78.76 $\pm$ 0.95} & 22.84 $\pm$ 1.41                              & \multicolumn{1}{l|}{78.10 $\pm$ 0.82}          & \multicolumn{1}{l|}{73.09 $\pm$ 1.20}   & \multicolumn{1}{l|}{78.10 $\pm$ 0.82} & 36.42 $\pm$ 2.19                              \\ \cline{2-13} 
                                                                                                                  & \multicolumn{1}{c|}{\checkmark}   &              & \multicolumn{1}{c|}{\checkmark}   & \checkmark   & \multicolumn{1}{l|}{77.94 $\pm$ 1.16}          & \multicolumn{1}{l|}{72.74 $\pm$ 1.99}   & \multicolumn{1}{l|}{77.94 $\pm$ 1.16} & 22.83 $\pm$ 1.41                              & \multicolumn{1}{l|}{77.94 $\pm$ 0.97}          & \multicolumn{1}{l|}{72.70 $\pm$ 1.40}   & \multicolumn{1}{l|}{77.94 $\pm$ 0.97} & 36.96 $\pm$ 2.05                              \\ \cline{2-13} 
                                                                                                                  & \multicolumn{1}{c|}{\checkmark}   & \checkmark   & \multicolumn{1}{c|}{\checkmark}   &              & \multicolumn{1}{l|}{78.82 $\pm$ 1.45}          & \multicolumn{1}{l|}{72.02 $\pm$ 1.81}   & \multicolumn{1}{l|}{78.82 $\pm$ 1.45} & 20.39 $\pm$ 1.81                              & \multicolumn{1}{l|}{77.78 $\pm$ 1.39}          & \multicolumn{1}{l|}{70.48 $\pm$ 1.92}   & \multicolumn{1}{l|}{77.78 $\pm$ 1.39} & 34.64 $\pm$ 2.30                              \\ \cline{2-13} 
                                                                                                                  & \multicolumn{1}{c|}{\checkmark}   & \checkmark   & \multicolumn{1}{c|}{\checkmark}   & \checkmark   & \multicolumn{1}{l|}{78.02 $\pm$ 1.14}          & \multicolumn{1}{l|}{71.45 $\pm$ 1.93}   & \multicolumn{1}{l|}{78.02 $\pm$ 1.14} & 22.28 $\pm$ 1.93                              & \multicolumn{1}{l|}{78.14 $\pm$ 0.95}          & \multicolumn{1}{l|}{71.74 $\pm$ 1.45}   & \multicolumn{1}{l|}{78.14 $\pm$ 0.95} & 39.49 $\pm$ 3.09                              \\ \hline
\multirow{6}{*}{\textsc{\textbf{Mittens}}}                                                                        & \multicolumn{1}{c|}{\checkmark}   &              & \multicolumn{2}{c|}{Branch N/A}                  & \multicolumn{1}{l|}{78.68 $\pm$ 0.78}          & \multicolumn{1}{l|}{72.43 $\pm$ 2.11}   & \multicolumn{1}{l|}{78.68 $\pm$ 0.78} & 19.33 $\pm$ 1.42                              & \multicolumn{1}{l|}{78.48 $\pm$ 1.05}          & \multicolumn{1}{l|}{72.04 $\pm$ 1.58}   & \multicolumn{1}{l|}{78.48 $\pm$ 1.05} & 29.49 $\pm$ 2.22                              \\ \cline{2-13} 
                                                                                                                  & \multicolumn{1}{c|}{\checkmark}   & \checkmark   & \multicolumn{2}{c|}{Branch N/A}                  & \multicolumn{1}{l|}{78.16 $\pm$ 0.85}          & \multicolumn{1}{l|}{69.95 $\pm$ 2.27}   & \multicolumn{1}{l|}{78.16 $\pm$ 0.85} & 17.49 $\pm$ 1.84                              & \multicolumn{1}{l|}{77.70 $\pm$ 1.35}          & \multicolumn{1}{l|}{69.93 $\pm$ 2.44}   & \multicolumn{1}{l|}{77.70 $\pm$ 1.35} & 24.15 $\pm$ 1.99                              \\ \cline{2-13} 
                                                                                                                  & \multicolumn{1}{c|}{\checkmark}   &              & \multicolumn{1}{c|}{\checkmark}   &              & \multicolumn{1}{l|}{79.18 $\pm$ 1.04}          & \multicolumn{1}{l|}{73.06 $\pm$ 2.11}   & \multicolumn{1}{l|}{79.18 $\pm$ 1.04} & 21.49 $\pm$ 1.53                              & \multicolumn{1}{l|}{79.00 $\pm$ 0.73}          & \multicolumn{1}{l|}{72.98 $\pm$ 1.01}   & \multicolumn{1}{l|}{79.00 $\pm$ 0.73} & 34.97 $\pm$ 2.27                              \\ \cline{2-13} 
                                                                                                                  & \multicolumn{1}{c|}{\checkmark}   &              & \multicolumn{1}{c|}{\checkmark}   & \checkmark   & \multicolumn{1}{l|}{\resultlowclr{\textbf{79.62 $\pm$ 0.52}}} & \multicolumn{1}{l|}{73.61 $\pm$ 1.08}   & \multicolumn{1}{l|}{79.62 $\pm$ 0.52} & 21.79 $\pm$ 1.51                              & \multicolumn{1}{l|}{\resultlowclr{\textbf{79.34 $\pm$ 1.13}}} & \multicolumn{1}{l|}{73.74 $\pm$ 1.89}   & \multicolumn{1}{l|}{79.34 $\pm$ 1.13} & 35.64 $\pm$ 3.08                              \\ \cline{2-13} 
                                                                                                                  & \multicolumn{1}{c|}{\checkmark}   & \checkmark   & \multicolumn{1}{c|}{\checkmark}   &              & \multicolumn{1}{l|}{79.14 $\pm$ 0.62}          & \multicolumn{1}{l|}{71.67 $\pm$ 1.08}   & \multicolumn{1}{l|}{79.14 $\pm$ 0.62} & 18.82 $\pm$ 1.04                              & \multicolumn{1}{l|}{78.28 $\pm$ 1.16}          & \multicolumn{1}{l|}{71.81 $\pm$ 1.64}   & \multicolumn{1}{l|}{78.28 $\pm$ 1.16} & 31.04 $\pm$ 2.40                              \\ \cline{2-13} 
                                                                                                                  & \multicolumn{1}{c|}{\checkmark}   & \checkmark   & \multicolumn{1}{c|}{\checkmark}   & \checkmark   & \multicolumn{1}{l|}{79.22 $\pm$ 0.82}          & \multicolumn{1}{l|}{71.48 $\pm$ 1.51}   & \multicolumn{1}{l|}{79.22 $\pm$ 0.82} & 21.45 $\pm$ 1.26                              & \multicolumn{1}{l|}{78.86 $\pm$ 0.64}          & \multicolumn{1}{l|}{72.14 $\pm$ 1.76}   & \multicolumn{1}{l|}{78.86 $\pm$ 0.64} & 37.68 $\pm$ 2.62                              \\ \hline
\end{tabular}
}
\end{table*}


\begin{table*}[!htbp]
\centering

\caption{Ablation testing on the BuzzFace open dataset~\cite{Santia2018} using [Bi]LSTM as [Bi]RNN units (bold marks the best accuracy w.r.t. word embedding)}
\label{tab:results_lstm_buzzface}
\resizebox{1\textwidth}{!}{%
\begin{tabular}{|c|cc|cc|llll|llll|}
\hline
\multicolumn{1}{|l|}{\multirow{3}{*}{\textbf{\begin{tabular}[c]{@{}c@{}}Word\\ Embedding\end{tabular}}}} & \multicolumn{4}{c|}{\textbf{Network Embedding}}  & \multicolumn{4}{c|}{\multirow{2}{*}{\textbf{GRU Cell}}}   & \multicolumn{4}{c|}{\multirow{2}{*}{\textbf{BiGRU Cell}}}                                                                                                                                           \\ \cline{2-5}
\multicolumn{1}{|l|}{}                               & \multicolumn{2}{c|}{\textbf{Text Branch}}     & \multicolumn{2}{c|}{\textbf{Social Branch}}     & \multicolumn{4}{c|}{} & \multicolumn{4}{c|}{}                                                                                                                                       \\ \cline{2-13} 
\multicolumn{1}{|l|}{}                                                                                                                 & \multicolumn{1}{c|}{\textbf{RNN}} & \textbf{CNN} & \multicolumn{1}{c|}{\textbf{RNN}} & \textbf{CNN} & \multicolumn{1}{c|}{\textbf{Accuracy}}         & \multicolumn{1}{c|}{\textbf{Precision}} & \multicolumn{1}{c|}{\textbf{Recall}}  & \multicolumn{1}{c|}{\textbf{Runtime(s)}} & \multicolumn{1}{c|}{\textbf{Accuracy}}         & \multicolumn{1}{c|}{\textbf{Precision}} & \multicolumn{1}{c|}{\textbf{Recall}}  & \multicolumn{1}{c|}{\textbf{Runtime(s)}} \\ \hline
\multirow{6}{*}{\begin{tabular}[c]{@{}c@{}}\textsc{\textbf{Word2Vec}}\\ \textbf{CBOW}\end{tabular}}               & \multicolumn{1}{c|}{\checkmark}   &              & \multicolumn{2}{c|}{Branch N/A}                  & \multicolumn{1}{l|}{78.94 $\pm$ 1.21}          & \multicolumn{1}{l|}{72.02 $\pm$ 2.22}   & \multicolumn{1}{l|}{78.94 $\pm$ 1.21} & 15.39 $\pm$ 1.28                              & \multicolumn{1}{l|}{79.02 $\pm$ 1.43}          & \multicolumn{1}{l|}{71.62 $\pm$ 1.72}   & \multicolumn{1}{l|}{79.02 $\pm$ 1.43} & 25.44 $\pm$ 1.55                              \\ \cline{2-13} 
                                                                                                                  & \multicolumn{1}{c|}{\checkmark}   & \checkmark   & \multicolumn{2}{c|}{Branch N/A}                  & \multicolumn{1}{l|}{77.80 $\pm$ 1.86}          & \multicolumn{1}{l|}{70.82 $\pm$ 1.62}   & \multicolumn{1}{l|}{77.80 $\pm$ 1.86} & 13.89 $\pm$ 0.92                              & \multicolumn{1}{l|}{77.68 $\pm$ 1.55}          & \multicolumn{1}{l|}{70.59 $\pm$ 1.62}   & \multicolumn{1}{l|}{77.68 $\pm$ 1.55} & 22.71 $\pm$ 1.33                              \\ \cline{2-13} 
                                                                                                                  & \multicolumn{1}{c|}{\checkmark}   &              & \multicolumn{1}{c|}{\checkmark}   &              & \multicolumn{1}{l|}{79.14 $\pm$ 1.85}          & \multicolumn{1}{l|}{72.91 $\pm$ 2.64}   & \multicolumn{1}{l|}{79.14 $\pm$ 1.85} & 18.19 $\pm$ 1.51                              & \multicolumn{1}{l|}{79.42 $\pm$ 1.01}          & \multicolumn{1}{l|}{72.52 $\pm$ 2.53}   & \multicolumn{1}{l|}{79.42 $\pm$ 1.01} & 31.15 $\pm$ 2.16                              \\ \cline{2-13} 
                                                                                                                  & \multicolumn{1}{c|}{\checkmark}   &              & \multicolumn{1}{c|}{\checkmark}   & \checkmark   & \multicolumn{1}{l|}{78.94 $\pm$ 1.40}          & \multicolumn{1}{l|}{71.33 $\pm$ 1.37}   & \multicolumn{1}{l|}{78.94 $\pm$ 1.40} & 18.09 $\pm$ 1.47                              & \multicolumn{1}{l|}{79.06 $\pm$ 1.27}          & \multicolumn{1}{l|}{72.48 $\pm$ 2.16}   & \multicolumn{1}{l|}{79.06 $\pm$ 1.27} & 30.55 $\pm$ 0.31                              \\ \cline{2-13} 
                                                                                                                  & \multicolumn{1}{c|}{\checkmark}   & \checkmark   & \multicolumn{1}{c|}{\checkmark}   &              & \multicolumn{1}{l|}{77.58 $\pm$ 1.38}          & \multicolumn{1}{l|}{70.92 $\pm$ 1.81}   & \multicolumn{1}{l|}{77.58 $\pm$ 1.38} & 17.85 $\pm$ 1.26                              & \multicolumn{1}{l|}{78.02 $\pm$ 1.34}          & \multicolumn{1}{l|}{70.45 $\pm$ 1.08}   & \multicolumn{1}{l|}{78.02 $\pm$ 1.34} & 29.17 $\pm$ 0.80                              \\ \cline{2-13} 
                                                                                                                  & \multicolumn{1}{c|}{\checkmark}   & \checkmark   & \multicolumn{1}{c|}{\checkmark}   & \checkmark   & \multicolumn{1}{l|}{77.84 $\pm$ 1.88}          & \multicolumn{1}{l|}{70.82 $\pm$ 2.55}   & \multicolumn{1}{l|}{77.84 $\pm$ 1.88} & 18.00 $\pm$ 1.74                              & \multicolumn{1}{l|}{77.94 $\pm$ 2.56}          & \multicolumn{1}{l|}{70.12 $\pm$ 1.55}   & \multicolumn{1}{l|}{77.94 $\pm$ 2.56} & 33.22 $\pm$ 3.32                              \\ \hline
\multirow{6}{*}{\begin{tabular}[c]{@{}c@{}}\textsc{\textbf{Word2Vec}}\\ \textsc{\textbf{Skip-Gran}}\end{tabular}} & \multicolumn{1}{c|}{\checkmark}   &              & \multicolumn{2}{c|}{Branch N/A}                  & \multicolumn{1}{l|}{76.76 $\pm$ 1.44}          & \multicolumn{1}{l|}{72.14 $\pm$ 1.20}   & \multicolumn{1}{l|}{76.76 $\pm$ 1.44} & 17.13 $\pm$ 1.49                              & \multicolumn{1}{l|}{76.76 $\pm$ 1.59}          & \multicolumn{1}{l|}{71.99 $\pm$ 2.10}   & \multicolumn{1}{l|}{76.76 $\pm$ 1.59} & 29.37 $\pm$ 2.66                              \\ \cline{2-13} 
                                                                                                                  & \multicolumn{1}{c|}{\checkmark}   & \checkmark   & \multicolumn{2}{c|}{Branch N/A}                  & \multicolumn{1}{l|}{76.36 $\pm$ 1.72}          & \multicolumn{1}{l|}{71.12 $\pm$ 0.99}   & \multicolumn{1}{l|}{76.36 $\pm$ 1.72} & 16.63 $\pm$ 1.75                              & \multicolumn{1}{l|}{77.62 $\pm$ 1.23}          & \multicolumn{1}{l|}{71.42 $\pm$ 1.12}   & \multicolumn{1}{l|}{77.62 $\pm$ 1.23} & 25.73 $\pm$ 2.31                              \\ \cline{2-13} 
                                                                                                                  & \multicolumn{1}{c|}{\checkmark}   &              & \multicolumn{1}{c|}{\checkmark}   &              & \multicolumn{1}{l|}{77.04 $\pm$ 1.36}          & \multicolumn{1}{l|}{72.53 $\pm$ 1.60}   & \multicolumn{1}{l|}{77.04 $\pm$ 1.36} & 19.31 $\pm$ 1.52                              & \multicolumn{1}{l|}{77.26 $\pm$ 1.55}          & \multicolumn{1}{l|}{72.88 $\pm$ 0.86}   & \multicolumn{1}{l|}{77.26 $\pm$ 1.55} & 34.25 $\pm$ 2.90                              \\ \cline{2-13} 
                                                                                                                  & \multicolumn{1}{c|}{\checkmark}   &              & \multicolumn{1}{c|}{\checkmark}   & \checkmark   & \multicolumn{1}{l|}{77.28 $\pm$ 1.35}          & \multicolumn{1}{l|}{73.10 $\pm$ 1.50}   & \multicolumn{1}{l|}{77.28 $\pm$ 1.35} & 20.08 $\pm$ 1.92                              & \multicolumn{1}{l|}{77.04 $\pm$ 1.54}          & \multicolumn{1}{l|}{72.30 $\pm$ 1.76}   & \multicolumn{1}{l|}{77.04 $\pm$ 1.54} & 34.46 $\pm$ 3.21                              \\ \cline{2-13} 
                                                                                                                  & \multicolumn{1}{c|}{\checkmark}   & \checkmark   & \multicolumn{1}{c|}{\checkmark}   &              & \multicolumn{1}{l|}{77.92 $\pm$ 1.57}          & \multicolumn{1}{l|}{72.57 $\pm$ 2.72}   & \multicolumn{1}{l|}{77.92 $\pm$ 1.57} & 19.46 $\pm$ 2.47                              & \multicolumn{1}{l|}{77.50 $\pm$ 1.63}          & \multicolumn{1}{l|}{72.11 $\pm$ 2.06}   & \multicolumn{1}{l|}{77.50 $\pm$ 1.63} & 32.78 $\pm$ 3.02                              \\ \cline{2-13} 
                                                                                                                  & \multicolumn{1}{c|}{\checkmark}   & \checkmark   & \multicolumn{1}{c|}{\checkmark}   & \checkmark   & \multicolumn{1}{l|}{77.14 $\pm$ 1.69}          & \multicolumn{1}{l|}{71.07 $\pm$ 2.37}   & \multicolumn{1}{l|}{77.14 $\pm$ 1.69} & 20.44 $\pm$ 2.02                              & \multicolumn{1}{l|}{77.50 $\pm$ 1.57}          & \multicolumn{1}{l|}{71.28 $\pm$ 1.84}   & \multicolumn{1}{l|}{77.50 $\pm$ 1.57} & 36.22 $\pm$ 4.41                              \\ \hline
\multirow{6}{*}{\begin{tabular}[c]{@{}c@{}}\textsc{\textbf{FastText}}\\ \textbf{CBOW}\end{tabular}}               & \multicolumn{1}{c|}{\checkmark}   &              & \multicolumn{2}{c|}{Branch N/A}                  & \multicolumn{1}{l|}{78.94 $\pm$ 1.11}          & \multicolumn{1}{l|}{72.34 $\pm$ 2.33}   & \multicolumn{1}{l|}{78.94 $\pm$ 1.11} & 16.04 $\pm$ 1.79                              & \multicolumn{1}{l|}{79.12 $\pm$ 0.72}          & \multicolumn{1}{l|}{72.03 $\pm$ 2.04}   & \multicolumn{1}{l|}{79.12 $\pm$ 0.72} & 26.09 $\pm$ 2.31                              \\ \cline{2-13} 
                                                                                                                  & \multicolumn{1}{c|}{\checkmark}   & \checkmark   & \multicolumn{2}{c|}{Branch N/A}                  & \multicolumn{1}{l|}{77.18 $\pm$ 0.94}          & \multicolumn{1}{l|}{69.43 $\pm$ 1.13}   & \multicolumn{1}{l|}{77.18 $\pm$ 0.94} & 15.35 $\pm$ 1.51                              & \multicolumn{1}{l|}{77.12 $\pm$ 1.23}          & \multicolumn{1}{l|}{70.28 $\pm$ 2.01}   & \multicolumn{1}{l|}{77.12 $\pm$ 1.23} & 23.77 $\pm$ 1.96                              \\ \cline{2-13} 
                                                                                                                  & \multicolumn{1}{c|}{\checkmark}   &              & \multicolumn{1}{c|}{\checkmark}   &              & \multicolumn{1}{l|}{78.78 $\pm$ 1.13}          & \multicolumn{1}{l|}{72.30 $\pm$ 1.82}   & \multicolumn{1}{l|}{78.78 $\pm$ 1.13} & 19.33 $\pm$ 1.69                              & \multicolumn{1}{l|}{79.14 $\pm$ 0.81}          & \multicolumn{1}{l|}{71.86 $\pm$ 2.34}   & \multicolumn{1}{l|}{79.14 $\pm$ 0.81} & 31.27 $\pm$ 2.45                              \\ \cline{2-13} 
                                                                                                                  & \multicolumn{1}{c|}{\checkmark}   &              & \multicolumn{1}{c|}{\checkmark}   & \checkmark   & \multicolumn{1}{l|}{79.08 $\pm$ 0.65}          & \multicolumn{1}{l|}{72.26 $\pm$ 2.01}   & \multicolumn{1}{l|}{79.08 $\pm$ 0.65} & 18.82 $\pm$ 1.51                              & \multicolumn{1}{l|}{79.24 $\pm$ 0.78}          & \multicolumn{1}{l|}{72.63 $\pm$ 2.09}   & \multicolumn{1}{l|}{79.24 $\pm$ 0.78} & 32.05 $\pm$ 2.33                              \\ \cline{2-13} 
                                                                                                                  & \multicolumn{1}{c|}{\checkmark}   & \checkmark   & \multicolumn{1}{c|}{\checkmark}   &              & \multicolumn{1}{l|}{77.38 $\pm$ 1.75}          & \multicolumn{1}{l|}{70.99 $\pm$ 2.43}   & \multicolumn{1}{l|}{77.38 $\pm$ 1.75} & 17.94 $\pm$ 1.37                              & \multicolumn{1}{l|}{78.04 $\pm$ 0.79}          & \multicolumn{1}{l|}{71.58 $\pm$ 1.15}   & \multicolumn{1}{l|}{78.04 $\pm$ 0.79} & 30.39 $\pm$ 1.77                              \\ \cline{2-13} 
                                                                                                                  & \multicolumn{1}{c|}{\checkmark}   & \checkmark   & \multicolumn{1}{c|}{\checkmark}   & \checkmark   & \multicolumn{1}{l|}{77.20 $\pm$ 0.54}          & \multicolumn{1}{l|}{69.92 $\pm$ 1.27}   & \multicolumn{1}{l|}{77.20 $\pm$ 0.54} & 19.85 $\pm$ 0.66                              & \multicolumn{1}{l|}{77.46 $\pm$ 1.17}          & \multicolumn{1}{l|}{71.40 $\pm$ 1.11}   & \multicolumn{1}{l|}{77.46 $\pm$ 1.17} & 36.17 $\pm$ 1.99                              \\ \hline
\multirow{6}{*}{\begin{tabular}[c]{@{}c@{}}\textsc{\textbf{FastText}}\\ \textsc{\textbf{Skip-Gran}}\end{tabular}} & \multicolumn{1}{c|}{\checkmark}   &              & \multicolumn{2}{c|}{Branch N/A}                  & \multicolumn{1}{l|}{76.28 $\pm$ 0.71}          & \multicolumn{1}{l|}{72.23 $\pm$ 0.82}   & \multicolumn{1}{l|}{76.28 $\pm$ 0.71} & 17.03 $\pm$ 0.56                              & \multicolumn{1}{l|}{76.26 $\pm$ 0.93}          & \multicolumn{1}{l|}{72.19 $\pm$ 1.57}   & \multicolumn{1}{l|}{76.26 $\pm$ 0.93} & 29.03 $\pm$ 0.59                              \\ \cline{2-13} 
                                                                                                                  & \multicolumn{1}{c|}{\checkmark}   & \checkmark   & \multicolumn{2}{c|}{Branch N/A}                  & \multicolumn{1}{l|}{76.40 $\pm$ 1.42}          & \multicolumn{1}{l|}{71.73 $\pm$ 1.02}   & \multicolumn{1}{l|}{76.40 $\pm$ 1.42} & 16.32 $\pm$ 1.25                              & \multicolumn{1}{l|}{76.84 $\pm$ 1.24}          & \multicolumn{1}{l|}{70.91 $\pm$ 1.57}   & \multicolumn{1}{l|}{76.84 $\pm$ 1.24} & 26.37 $\pm$ 1.31                              \\ \cline{2-13} 
                                                                                                                  & \multicolumn{1}{c|}{\checkmark}   &              & \multicolumn{1}{c|}{\checkmark}   &              & \multicolumn{1}{l|}{76.56 $\pm$ 1.26}          & \multicolumn{1}{l|}{71.78 $\pm$ 1.81}   & \multicolumn{1}{l|}{76.56 $\pm$ 1.26} & 19.46 $\pm$ 1.35                              & \multicolumn{1}{l|}{76.22 $\pm$ 0.72}          & \multicolumn{1}{l|}{72.73 $\pm$ 1.06}   & \multicolumn{1}{l|}{76.22 $\pm$ 0.72} & 34.52 $\pm$ 0.41                              \\ \cline{2-13} 
                                                                                                                  & \multicolumn{1}{c|}{\checkmark}   &              & \multicolumn{1}{c|}{\checkmark}   & \checkmark   & \multicolumn{1}{l|}{77.44 $\pm$ 1.09}          & \multicolumn{1}{l|}{73.05 $\pm$ 0.92}   & \multicolumn{1}{l|}{77.44 $\pm$ 1.09} & 19.78 $\pm$ 0.71                              & \multicolumn{1}{l|}{76.36 $\pm$ 1.75}          & \multicolumn{1}{l|}{72.19 $\pm$ 1.32}   & \multicolumn{1}{l|}{76.36 $\pm$ 1.75} & 34.86 $\pm$ 0.84                              \\ \cline{2-13} 
                                                                                                                  & \multicolumn{1}{c|}{\checkmark}   & \checkmark   & \multicolumn{1}{c|}{\checkmark}   &              & \multicolumn{1}{l|}{77.20 $\pm$ 1.25}          & \multicolumn{1}{l|}{72.25 $\pm$ 2.24}   & \multicolumn{1}{l|}{77.20 $\pm$ 1.25} & 19.18 $\pm$ 0.95                              & \multicolumn{1}{l|}{76.82 $\pm$ 1.54}          & \multicolumn{1}{l|}{71.25 $\pm$ 1.22}   & \multicolumn{1}{l|}{76.82 $\pm$ 1.54} & 32.97 $\pm$ 0.96                              \\ \cline{2-13} 
                                                                                                                  & \multicolumn{1}{c|}{\checkmark}   & \checkmark   & \multicolumn{1}{c|}{\checkmark}   & \checkmark   & \multicolumn{1}{l|}{75.98 $\pm$ 1.32}          & \multicolumn{1}{l|}{71.29 $\pm$ 1.89}   & \multicolumn{1}{l|}{75.98 $\pm$ 1.32} & 20.44 $\pm$ 1.00                              & \multicolumn{1}{l|}{77.16 $\pm$ 1.72}          & \multicolumn{1}{l|}{71.77 $\pm$ 2.10}   & \multicolumn{1}{l|}{77.16 $\pm$ 1.72} & 36.71 $\pm$ 0.92                              \\ \hline
\multirow{6}{*}{\textsc{\textbf{GloVe}}}                                                                          & \multicolumn{1}{c|}{\checkmark}   &              & \multicolumn{2}{c|}{Branch N/A}                  & \multicolumn{1}{l|}{77.52 $\pm$ 1.17}          & \multicolumn{1}{l|}{71.74 $\pm$ 1.90}   & \multicolumn{1}{l|}{77.52 $\pm$ 1.17} & 18.26 $\pm$ 1.25                              & \multicolumn{1}{l|}{77.88 $\pm$ 0.96}          & \multicolumn{1}{l|}{72.41 $\pm$ 1.67}   & \multicolumn{1}{l|}{77.88 $\pm$ 0.96} & 33.12 $\pm$ 2.11                              \\ \cline{2-13} 
                                                                                                                  & \multicolumn{1}{c|}{\checkmark}   & \checkmark   & \multicolumn{2}{c|}{Branch N/A}                  & \multicolumn{1}{l|}{77.42 $\pm$ 1.05}          & \multicolumn{1}{l|}{70.80 $\pm$ 1.11}   & \multicolumn{1}{l|}{77.42 $\pm$ 1.05} & 18.51 $\pm$ 1.37                              & \multicolumn{1}{l|}{78.00 $\pm$ 1.29}          & \multicolumn{1}{l|}{71.78 $\pm$ 2.26}   & \multicolumn{1}{l|}{78.00 $\pm$ 1.29} & 29.64 $\pm$ 2.80                              \\ \cline{2-13} 
                                                                                                                  & \multicolumn{1}{c|}{\checkmark}   &              & \multicolumn{1}{c|}{\checkmark}   &              & \multicolumn{1}{l|}{78.24 $\pm$ 1.03}          & \multicolumn{1}{l|}{72.37 $\pm$ 1.82}   & \multicolumn{1}{l|}{78.24 $\pm$ 1.03} & 21.88 $\pm$ 1.17                              & \multicolumn{1}{l|}{78.42 $\pm$ 0.93}          & \multicolumn{1}{l|}{73.09 $\pm$ 1.84}   & \multicolumn{1}{l|}{78.42 $\pm$ 0.93} & 38.58 $\pm$ 2.44                              \\ \cline{2-13} 
                                                                                                                  & \multicolumn{1}{c|}{\checkmark}   &              & \multicolumn{1}{c|}{\checkmark}   & \checkmark   & \multicolumn{1}{l|}{77.88 $\pm$ 1.25}          & \multicolumn{1}{l|}{71.73 $\pm$ 1.67}   & \multicolumn{1}{l|}{77.88 $\pm$ 1.25} & 22.26 $\pm$ 1.50                              & \multicolumn{1}{l|}{78.10 $\pm$ 1.08}          & \multicolumn{1}{l|}{72.47 $\pm$ 2.40}   & \multicolumn{1}{l|}{78.10 $\pm$ 1.08} & 38.38 $\pm$ 3.92                              \\ \cline{2-13} 
                                                                                                                  & \multicolumn{1}{c|}{\checkmark}   & \checkmark   & \multicolumn{1}{c|}{\checkmark}   &              & \multicolumn{1}{l|}{78.24 $\pm$ 1.46}          & \multicolumn{1}{l|}{72.21 $\pm$ 2.71}   & \multicolumn{1}{l|}{78.24 $\pm$ 1.46} & 20.95 $\pm$ 1.84                              & \multicolumn{1}{l|}{77.72 $\pm$ 1.76}          & \multicolumn{1}{l|}{72.71 $\pm$ 1.88}   & \multicolumn{1}{l|}{77.72 $\pm$ 1.76} & 36.36 $\pm$ 2.87                              \\ \cline{2-13} 
                                                                                                                  & \multicolumn{1}{c|}{\checkmark}   & \checkmark   & \multicolumn{1}{c|}{\checkmark}   & \checkmark   & \multicolumn{1}{l|}{78.32 $\pm$ 1.29}          & \multicolumn{1}{l|}{70.93 $\pm$ 1.48}   & \multicolumn{1}{l|}{78.32 $\pm$ 1.29} & 21.70 $\pm$ 1.65                              & \multicolumn{1}{l|}{78.16 $\pm$ 1.01}          & \multicolumn{1}{l|}{71.28 $\pm$ 1.54}   & \multicolumn{1}{l|}{78.16 $\pm$ 1.01} & 40.38 $\pm$ 3.48                              \\ \hline
\multirow{6}{*}{\textsc{\textbf{Mittens}}}                                                                        & \multicolumn{1}{c|}{\checkmark}   &              & \multicolumn{2}{c|}{Branch N/A}                  & \multicolumn{1}{l|}{78.12 $\pm$ 1.32}          & \multicolumn{1}{l|}{72.92 $\pm$ 1.76}   & \multicolumn{1}{l|}{78.12 $\pm$ 1.32} & 18.27 $\pm$ 1.49                              & \multicolumn{1}{l|}{78.04 $\pm$ 1.37}          & \multicolumn{1}{l|}{72.89 $\pm$ 1.78}   & \multicolumn{1}{l|}{78.04 $\pm$ 1.37} & 31.60 $\pm$ 3.38                              \\ \cline{2-13} 
                                                                                                                  & \multicolumn{1}{c|}{\checkmark}   & \checkmark   & \multicolumn{2}{c|}{Branch N/A}                  & \multicolumn{1}{l|}{78.90 $\pm$ 0.91}          & \multicolumn{1}{l|}{72.61 $\pm$ 1.77}   & \multicolumn{1}{l|}{78.90 $\pm$ 0.91} & 17.28 $\pm$ 1.51                              & \multicolumn{1}{l|}{77.64 $\pm$ 1.64}          & \multicolumn{1}{l|}{72.10 $\pm$ 1.53}   & \multicolumn{1}{l|}{77.64 $\pm$ 1.64} & 27.99 $\pm$ 2.44                              \\ \cline{2-13} 
                                                                                                                  & \multicolumn{1}{c|}{\checkmark}   &              & \multicolumn{1}{c|}{\checkmark}   &              & \multicolumn{1}{l|}{78.52 $\pm$ 1.77}          & \multicolumn{1}{l|}{74.13 $\pm$ 2.17}   & \multicolumn{1}{l|}{78.52 $\pm$ 1.77} & 21.28 $\pm$ 1.20                              & \multicolumn{1}{l|}{77.72 $\pm$ 2.06}          & \multicolumn{1}{l|}{73.33 $\pm$ 1.92}   & \multicolumn{1}{l|}{77.72 $\pm$ 2.06} & 36.46 $\pm$ 2.87                              \\ \cline{2-13} 
                                                                                                                  & \multicolumn{1}{c|}{\checkmark}   &              & \multicolumn{1}{c|}{\checkmark}   & \checkmark   & \multicolumn{1}{l|}{\resultlowclr{\textbf{79.74 $\pm$ 1.08}}} & \multicolumn{1}{l|}{74.57 $\pm$ 1.56}   & \multicolumn{1}{l|}{79.74 $\pm$ 1.08} & 21.30 $\pm$ 1.03                              & \multicolumn{1}{l|}{\resultlowclr{\textbf{79.63 $\pm$ 1.30}}} & \multicolumn{1}{l|}{74.61 $\pm$ 1.81}   & \multicolumn{1}{l|}{79.63 $\pm$ 1.30} & 37.44 $\pm$ 2.91                              \\ \cline{2-13} 
                                                                                                                  & \multicolumn{1}{c|}{\checkmark}   & \checkmark   & \multicolumn{1}{c|}{\checkmark}   &              & \multicolumn{1}{l|}{77.84 $\pm$ 1.55}          & \multicolumn{1}{l|}{71.62 $\pm$ 1.27}   & \multicolumn{1}{l|}{77.84 $\pm$ 1.55} & 20.70 $\pm$ 1.36                              & \multicolumn{1}{l|}{78.40 $\pm$ 1.22}          & \multicolumn{1}{l|}{72.56 $\pm$ 1.24}   & \multicolumn{1}{l|}{78.40 $\pm$ 1.22} & 35.32 $\pm$ 2.51                              \\ \cline{2-13} 
                                                                                                                  & \multicolumn{1}{c|}{\checkmark}   & \checkmark   & \multicolumn{1}{c|}{\checkmark}   & \checkmark   & \multicolumn{1}{l|}{77.96 $\pm$ 1.40}          & \multicolumn{1}{l|}{71.46 $\pm$ 1.26}   & \multicolumn{1}{l|}{77.96 $\pm$ 1.40} & 22.28 $\pm$ 1.20                              & \multicolumn{1}{l|}{78.50 $\pm$ 1.46}          & \multicolumn{1}{l|}{72.57 $\pm$ 1.81}   & \multicolumn{1}{l|}{78.50 $\pm$ 1.46} & 39.54 $\pm$ 3.04                              \\ \hline
\end{tabular}
}
\end{table*}

We observe that for some configurations, the Bidirectional layers perform better than the Unidirectional layers. As results are not conclusive regarding this aspect, this should be decided using a data-driven approach on a case-by-case basis. However, based on the ablation tests, we can infer that by adding the social context in the classification process, the performance of fake news detection increases.

\subsubsection{Twitter15 results}

For the Twitter15 dataset (Tables~\ref{tab:results_gru_twitter15} and~\ref{tab:results_lstm_twitter15}), 
we observe that the models trained with \textsc{Mittens} which use the [Bi]RNN for the Text Branch and an [Bi]RNN followed by CNN for the Social Branch to create the novel Network Embedding obtain the best results.


\begin{table*}[!htbp]
\centering

\caption{Ablation testing on the Twitter15 dataset~\cite{Ma2017} using [Bi]GRU as [Bi]RNN units (bold marks the best accuracy w.r.t. word embedding)}
\label{tab:results_gru_twitter15}
\resizebox{1\textwidth}{!}{%
\begin{tabular}{|c|cc|cc|cccc|cccc|}
\hline
\multicolumn{1}{|l|}{\multirow{3}{*}{\textbf{\begin{tabular}[c]{@{}c@{}}Word\\ Embedding\end{tabular}}}} & \multicolumn{4}{c|}{\textbf{Network Embedding}}  & \multicolumn{4}{c|}{\multirow{2}{*}{\textbf{GRU Cell}}}   & \multicolumn{4}{c|}{\multirow{2}{*}{\textbf{BiGRU Cell}}}                                                                                                                                           \\ \cline{2-5}
\multicolumn{1}{|l|}{}                               & \multicolumn{2}{c|}{\textbf{Text Branch}}     & \multicolumn{2}{c|}{\textbf{Social Branch}}     & \multicolumn{4}{c|}{} & \multicolumn{4}{c|}{}                                                                                                                                       \\ \cline{2-13} 
\multicolumn{1}{|l|}{}                                                                                                                 & \multicolumn{1}{c|}{\textbf{RNN}} & \textbf{CNN} & \multicolumn{1}{c|}{\textbf{RNN}} & \textbf{CNN} & \multicolumn{1}{c|}{\textbf{Accuracy}}         & \multicolumn{1}{c|}{\textbf{Precision}} & \multicolumn{1}{c|}{\textbf{Recall}}  & \multicolumn{1}{c|}{\textbf{Runtime(s)}} & \multicolumn{1}{c|}{\textbf{Accuracy}}         & \multicolumn{1}{c|}{\textbf{Precision}} & \multicolumn{1}{c|}{\textbf{Recall}}  & \multicolumn{1}{c|}{\textbf{Runtime(s)}} \\ \hline
\multirow{6}{*}{\begin{tabular}[c]{@{}c@{}}\textsc{\textbf{Word2Vec}}\\ \textbf{CBOW}\end{tabular}}               & \multicolumn{1}{c|}{\checkmark}   &              & \multicolumn{2}{c|}{Branch N/A}                  & \multicolumn{1}{c|}{67.92 $\pm$ 2.05}          & \multicolumn{1}{c|}{68.10 $\pm$ 2.12}   & \multicolumn{1}{c|}{67.92 $\pm$ 2.05} & 13.80 $\pm$ 1.46         & \multicolumn{1}{c|}{65.68 $\pm$ 1.88}          & \multicolumn{1}{c|}{65.70 $\pm$ 2.04}   & \multicolumn{1}{c|}{65.68 $\pm$ 1.88} & 21.88 $\pm$ 2.53         \\ \cline{2-13} 
                                                                                                                  & \multicolumn{1}{c|}{\checkmark}   & \checkmark   & \multicolumn{2}{c|}{Branch N/A}                  & \multicolumn{1}{c|}{65.59 $\pm$ 2.21}          & \multicolumn{1}{c|}{65.81 $\pm$ 1.99}   & \multicolumn{1}{c|}{65.59 $\pm$ 2.21} & 12.60 $\pm$ 1.49         & \multicolumn{1}{c|}{62.89 $\pm$ 1.88}          & \multicolumn{1}{c|}{62.80 $\pm$ 1.84}   & \multicolumn{1}{c|}{62.89 $\pm$ 1.88} & 18.75 $\pm$ 1.43         \\ \cline{2-13} 
                                                                                                                  & \multicolumn{1}{c|}{\checkmark}   &              & \multicolumn{1}{c|}{\checkmark}   &              & \multicolumn{1}{c|}{70.87 $\pm$ 2.85}          & \multicolumn{1}{c|}{71.14 $\pm$ 2.83}   & \multicolumn{1}{c|}{70.87 $\pm$ 2.85} & 16.90 $\pm$ 1.65         & \multicolumn{1}{c|}{69.98 $\pm$ 2.78}          & \multicolumn{1}{c|}{70.14 $\pm$ 2.77}   & \multicolumn{1}{c|}{69.98 $\pm$ 2.78} & 29.00 $\pm$ 2.06         \\ \cline{2-13} 
                                                                                                                  & \multicolumn{1}{c|}{\checkmark}   &              & \multicolumn{1}{c|}{\checkmark}   & \checkmark   & \multicolumn{1}{c|}{73.87 $\pm$ 1.61}          & \multicolumn{1}{c|}{74.05 $\pm$ 1.72}   & \multicolumn{1}{c|}{73.87 $\pm$ 1.61} & 16.46 $\pm$ 0.89         & \multicolumn{1}{c|}{72.46 $\pm$ 2.05}          & \multicolumn{1}{c|}{72.67 $\pm$ 2.03}   & \multicolumn{1}{c|}{72.46 $\pm$ 2.05} & 27.72 $\pm$ 2.15         \\ \cline{2-13} 
                                                                                                                  & \multicolumn{1}{c|}{\checkmark}   & \checkmark   & \multicolumn{1}{c|}{\checkmark}   &              & \multicolumn{1}{c|}{68.97 $\pm$ 1.87}          & \multicolumn{1}{c|}{68.93 $\pm$ 1.91}   & \multicolumn{1}{c|}{68.97 $\pm$ 1.87} & 16.16 $\pm$ 1.05         & \multicolumn{1}{c|}{66.87 $\pm$ 2.36}          & \multicolumn{1}{c|}{67.04 $\pm$ 2.19}   & \multicolumn{1}{c|}{66.87 $\pm$ 2.36} & 24.98 $\pm$ 1.70         \\ \cline{2-13} 
                                                                                                                  & \multicolumn{1}{c|}{\checkmark}   & \checkmark   & \multicolumn{1}{c|}{\checkmark}   & \checkmark   & \multicolumn{1}{c|}{73.33 $\pm$ 2.58}          & \multicolumn{1}{c|}{73.44 $\pm$ 2.50}   & \multicolumn{1}{c|}{73.33 $\pm$ 2.58} & 16.69 $\pm$ 1.17         & \multicolumn{1}{c|}{71.50 $\pm$ 2.00}          & \multicolumn{1}{c|}{71.74 $\pm$ 2.03}   & \multicolumn{1}{c|}{71.50 $\pm$ 2.00} & 27.07 $\pm$ 1.77         \\ \hline
\multirow{6}{*}{\begin{tabular}[c]{@{}c@{}}\textsc{\textbf{Word2Vec}}\\ \textsc{\textbf{Skip-Gran}}\end{tabular}} & \multicolumn{1}{c|}{\checkmark}   &              & \multicolumn{2}{c|}{Branch N/A}                  & \multicolumn{1}{c|}{69.64 $\pm$ 3.16}          & \multicolumn{1}{c|}{69.85 $\pm$ 3.11}   & \multicolumn{1}{c|}{69.64 $\pm$ 3.16} & 12.91 $\pm$ 1.86         & \multicolumn{1}{c|}{68.95 $\pm$ 1.95}          & \multicolumn{1}{c|}{69.16 $\pm$ 2.07}   & \multicolumn{1}{c|}{68.95 $\pm$ 1.95} & 20.61 $\pm$ 2.17         \\ \cline{2-13} 
                                                                                                                  & \multicolumn{1}{c|}{\checkmark}   & \checkmark   & \multicolumn{2}{c|}{Branch N/A}                  & \multicolumn{1}{c|}{67.99 $\pm$ 2.51}          & \multicolumn{1}{c|}{68.16 $\pm$ 2.68}   & \multicolumn{1}{c|}{67.99 $\pm$ 2.51} & 12.34 $\pm$ 1.18         & \multicolumn{1}{c|}{65.91 $\pm$ 2.69}          & \multicolumn{1}{c|}{65.89 $\pm$ 2.81}   & \multicolumn{1}{c|}{65.91 $\pm$ 2.69} & 19.63 $\pm$ 1.35         \\ \cline{2-13} 
                                                                                                                  & \multicolumn{1}{c|}{\checkmark}   &              & \multicolumn{1}{c|}{\checkmark}   &              & \multicolumn{1}{c|}{72.21 $\pm$ 2.59}          & \multicolumn{1}{c|}{72.47 $\pm$ 2.66}   & \multicolumn{1}{c|}{72.21 $\pm$ 2.59} & 15.72 $\pm$ 1.29         & \multicolumn{1}{c|}{72.33 $\pm$ 2.62}          & \multicolumn{1}{c|}{72.62 $\pm$ 2.65}   & \multicolumn{1}{c|}{72.33 $\pm$ 2.62} & 26.80 $\pm$ 1.94         \\ \cline{2-13} 
                                                                                                                  & \multicolumn{1}{c|}{\checkmark}   &              & \multicolumn{1}{c|}{\checkmark}   & \checkmark   & \multicolumn{1}{c|}{75.46 $\pm$ 1.99}          & \multicolumn{1}{c|}{75.74 $\pm$ 2.07}   & \multicolumn{1}{c|}{75.46 $\pm$ 1.99} & 15.85 $\pm$ 0.80         & \multicolumn{1}{c|}{74.54 $\pm$ 1.87}          & \multicolumn{1}{c|}{74.89 $\pm$ 1.86}   & \multicolumn{1}{c|}{74.54 $\pm$ 1.87} & 26.06 $\pm$ 1.87         \\ \cline{2-13} 
                                                                                                                  & \multicolumn{1}{c|}{\checkmark}   & \checkmark   & \multicolumn{1}{c|}{\checkmark}   &              & \multicolumn{1}{c|}{70.98 $\pm$ 2.38}          & \multicolumn{1}{c|}{71.19 $\pm$ 2.42}   & \multicolumn{1}{c|}{70.98 $\pm$ 2.38} & 15.75 $\pm$ 1.12         & \multicolumn{1}{c|}{70.60 $\pm$ 2.48}          & \multicolumn{1}{c|}{71.11 $\pm$ 2.40}   & \multicolumn{1}{c|}{70.60 $\pm$ 2.48} & 25.60 $\pm$ 2.30         \\ \cline{2-13} 
                                                                                                                  & \multicolumn{1}{c|}{\checkmark}   & \checkmark   & \multicolumn{1}{c|}{\checkmark}   & \checkmark   & \multicolumn{1}{c|}{73.98 $\pm$ 2.17}          & \multicolumn{1}{c|}{74.20 $\pm$ 2.27}   & \multicolumn{1}{c|}{73.98 $\pm$ 2.17} & 16.10 $\pm$ 1.07         & \multicolumn{1}{c|}{73.83 $\pm$ 2.19}          & \multicolumn{1}{c|}{74.27 $\pm$ 2.31}   & \multicolumn{1}{c|}{73.83 $\pm$ 2.19} & 26.33 $\pm$ 2.25         \\ \hline
\multirow{6}{*}{\begin{tabular}[c]{@{}c@{}}\textsc{\textbf{FastText}}\\ \textbf{CBOW}\end{tabular}}               & \multicolumn{1}{c|}{\checkmark}   &              & \multicolumn{2}{c|}{Branch N/A}                  & \multicolumn{1}{c|}{64.56 $\pm$ 1.85}          & \multicolumn{1}{c|}{64.62 $\pm$ 1.83}   & \multicolumn{1}{c|}{64.56 $\pm$ 1.85} & 13.65 $\pm$ 1.20         & \multicolumn{1}{c|}{62.66 $\pm$ 1.94}          & \multicolumn{1}{c|}{62.72 $\pm$ 1.96}   & \multicolumn{1}{c|}{62.66 $\pm$ 1.94} & 20.29 $\pm$ 1.48         \\ \cline{2-13} 
                                                                                                                  & \multicolumn{1}{c|}{\checkmark}   & \checkmark   & \multicolumn{2}{c|}{Branch N/A}                  & \multicolumn{1}{c|}{62.04 $\pm$ 1.21}          & \multicolumn{1}{c|}{62.28 $\pm$ 1.07}   & \multicolumn{1}{c|}{62.04 $\pm$ 1.21} & 11.67 $\pm$ 1.22         & \multicolumn{1}{c|}{60.11 $\pm$ 1.53}          & \multicolumn{1}{c|}{60.41 $\pm$ 1.22}   & \multicolumn{1}{c|}{60.11 $\pm$ 1.53} & 17.99 $\pm$ 1.40         \\ \cline{2-13} 
                                                                                                                  & \multicolumn{1}{c|}{\checkmark}   &              & \multicolumn{1}{c|}{\checkmark}   &              & \multicolumn{1}{c|}{68.77 $\pm$ 1.79}          & \multicolumn{1}{c|}{68.81 $\pm$ 1.80}   & \multicolumn{1}{c|}{68.77 $\pm$ 1.79} & 16.77 $\pm$ 1.51         & \multicolumn{1}{c|}{66.96 $\pm$ 2.01}          & \multicolumn{1}{c|}{67.03 $\pm$ 2.04}   & \multicolumn{1}{c|}{66.96 $\pm$ 2.01} & 26.34 $\pm$ 2.18         \\ \cline{2-13} 
                                                                                                                  & \multicolumn{1}{c|}{\checkmark}   &              & \multicolumn{1}{c|}{\checkmark}   & \checkmark   & \multicolumn{1}{c|}{71.95 $\pm$ 2.57}          & \multicolumn{1}{c|}{72.08 $\pm$ 2.60}   & \multicolumn{1}{c|}{71.95 $\pm$ 2.57} & 16.07 $\pm$ 0.94         & \multicolumn{1}{c|}{70.18 $\pm$ 2.58}          & \multicolumn{1}{c|}{70.27 $\pm$ 2.64}   & \multicolumn{1}{c|}{70.18 $\pm$ 2.58} & 25.95 $\pm$ 1.07         \\ \cline{2-13} 
                                                                                                                  & \multicolumn{1}{c|}{\checkmark}   & \checkmark   & \multicolumn{1}{c|}{\checkmark}   &              & \multicolumn{1}{c|}{67.14 $\pm$ 2.78}          & \multicolumn{1}{c|}{67.29 $\pm$ 2.72}   & \multicolumn{1}{c|}{67.14 $\pm$ 2.78} & 15.49 $\pm$ 1.36         & \multicolumn{1}{c|}{65.41 $\pm$ 2.13}          & \multicolumn{1}{c|}{65.46 $\pm$ 2.04}   & \multicolumn{1}{c|}{65.41 $\pm$ 2.13} & 24.19 $\pm$ 1.40         \\ \cline{2-13} 
                                                                                                                  & \multicolumn{1}{c|}{\checkmark}   & \checkmark   & \multicolumn{1}{c|}{\checkmark}   & \checkmark   & \multicolumn{1}{c|}{70.60 $\pm$ 1.53}          & \multicolumn{1}{c|}{71.04 $\pm$ 1.47}   & \multicolumn{1}{c|}{70.60 $\pm$ 1.53} & 15.68 $\pm$ 0.96         & \multicolumn{1}{c|}{70.36 $\pm$ 3.06}          & \multicolumn{1}{c|}{70.47 $\pm$ 3.17}   & \multicolumn{1}{c|}{70.36 $\pm$ 3.06} & 25.88 $\pm$ 1.93         \\ \hline
\multirow{6}{*}{\begin{tabular}[c]{@{}c@{}}\textsc{\textbf{FastText}}\\ \textsc{\textbf{Skip-Gran}}\end{tabular}} & \multicolumn{1}{c|}{\checkmark}   &              & \multicolumn{2}{c|}{Branch N/A}                  & \multicolumn{1}{c|}{69.82 $\pm$ 2.45}          & \multicolumn{1}{c|}{70.11 $\pm$ 2.49}   & \multicolumn{1}{c|}{69.82 $\pm$ 2.45} & 13.48 $\pm$ 1.12         & \multicolumn{1}{c|}{69.15 $\pm$ 2.35}          & \multicolumn{1}{c|}{69.23 $\pm$ 2.42}   & \multicolumn{1}{c|}{69.15 $\pm$ 2.35} & 22.18 $\pm$ 1.86         \\ \cline{2-13} 
                                                                                                                  & \multicolumn{1}{c|}{\checkmark}   & \checkmark   & \multicolumn{2}{c|}{Branch N/A}                  & \multicolumn{1}{c|}{67.83 $\pm$ 3.14}          & \multicolumn{1}{c|}{68.19 $\pm$ 3.19}   & \multicolumn{1}{c|}{67.83 $\pm$ 3.14} & 13.41 $\pm$ 1.53         & \multicolumn{1}{c|}{66.17 $\pm$ 2.00}          & \multicolumn{1}{c|}{66.43 $\pm$ 2.03}   & \multicolumn{1}{c|}{66.17 $\pm$ 2.00} & 20.97 $\pm$ 2.30         \\ \cline{2-13} 
                                                                                                                  & \multicolumn{1}{c|}{\checkmark}   &              & \multicolumn{1}{c|}{\checkmark}   &              & \multicolumn{1}{c|}{72.39 $\pm$ 2.03}          & \multicolumn{1}{c|}{72.55 $\pm$ 1.98}   & \multicolumn{1}{c|}{72.39 $\pm$ 2.03} & 16.38 $\pm$ 1.30         & \multicolumn{1}{c|}{72.44 $\pm$ 1.81}          & \multicolumn{1}{c|}{72.56 $\pm$ 1.68}   & \multicolumn{1}{c|}{72.44 $\pm$ 1.81} & 28.38 $\pm$ 2.12         \\ \cline{2-13} 
                                                                                                                  & \multicolumn{1}{c|}{\checkmark}   &              & \multicolumn{1}{c|}{\checkmark}   & \checkmark   & \multicolumn{1}{c|}{74.99 $\pm$ 2.11}          & \multicolumn{1}{c|}{75.32 $\pm$ 2.25}   & \multicolumn{1}{c|}{74.99 $\pm$ 2.11} & 16.23 $\pm$ 1.21         & \multicolumn{1}{c|}{75.21 $\pm$ 1.86}          & \multicolumn{1}{c|}{75.47 $\pm$ 1.94}   & \multicolumn{1}{c|}{75.21 $\pm$ 1.86} & 28.01 $\pm$ 1.98         \\ \cline{2-13} 
                                                                                                                  & \multicolumn{1}{c|}{\checkmark}   & \checkmark   & \multicolumn{1}{c|}{\checkmark}   &              & \multicolumn{1}{c|}{72.30 $\pm$ 2.20}          & \multicolumn{1}{c|}{72.49 $\pm$ 2.09}   & \multicolumn{1}{c|}{72.30 $\pm$ 2.20} & 16.32 $\pm$ 0.84         & \multicolumn{1}{c|}{70.89 $\pm$ 2.16}          & \multicolumn{1}{c|}{71.04 $\pm$ 2.03}   & \multicolumn{1}{c|}{70.89 $\pm$ 2.16} & 26.72 $\pm$ 2.08         \\ \cline{2-13} 
                                                                                                                  & \multicolumn{1}{c|}{\checkmark}   & \checkmark   & \multicolumn{1}{c|}{\checkmark}   & \checkmark   & \multicolumn{1}{c|}{75.32 $\pm$ 1.51}          & \multicolumn{1}{c|}{75.53 $\pm$ 1.48}   & \multicolumn{1}{c|}{75.32 $\pm$ 1.51} & 16.35 $\pm$ 0.70         & \multicolumn{1}{c|}{74.85 $\pm$ 2.88}          & \multicolumn{1}{c|}{75.05 $\pm$ 3.03}   & \multicolumn{1}{c|}{74.85 $\pm$ 2.88} & 27.49 $\pm$ 1.90         \\ \hline
\multirow{6}{*}{\textsc{\textbf{GloVe}}}                                                                          & \multicolumn{1}{c|}{\checkmark}   &              & \multicolumn{2}{c|}{Branch N/A}                  & \multicolumn{1}{c|}{72.44 $\pm$ 2.72}          & \multicolumn{1}{c|}{73.50 $\pm$ 2.37}   & \multicolumn{1}{c|}{72.44 $\pm$ 2.72} & 14.57 $\pm$ 1.07         & \multicolumn{1}{c|}{71.66 $\pm$ 4.94}          & \multicolumn{1}{c|}{72.70 $\pm$ 4.73}   & \multicolumn{1}{c|}{71.66 $\pm$ 4.94} & 25.30 $\pm$ 2.73         \\ \cline{2-13} 
                                                                                                                  & \multicolumn{1}{c|}{\checkmark}   & \checkmark   & \multicolumn{2}{c|}{Branch N/A}                  & \multicolumn{1}{c|}{71.86 $\pm$ 2.64}          & \multicolumn{1}{c|}{74.17 $\pm$ 2.55}   & \multicolumn{1}{c|}{71.86 $\pm$ 2.64} & 14.88 $\pm$ 0.73         & \multicolumn{1}{c|}{69.04 $\pm$ 4.01}          & \multicolumn{1}{c|}{71.32 $\pm$ 3.39}   & \multicolumn{1}{c|}{69.04 $\pm$ 4.01} & 25.59 $\pm$ 0.85         \\ \cline{2-13} 
                                                                                                                  & \multicolumn{1}{c|}{\checkmark}   &              & \multicolumn{1}{c|}{\checkmark}   &              & \multicolumn{1}{c|}{75.19 $\pm$ 2.26}          & \multicolumn{1}{c|}{76.47 $\pm$ 2.50}   & \multicolumn{1}{c|}{75.19 $\pm$ 2.26} & 17.46 $\pm$ 0.95         & \multicolumn{1}{c|}{75.62 $\pm$ 1.94}          & \multicolumn{1}{c|}{76.77 $\pm$ 1.88}   & \multicolumn{1}{c|}{75.62 $\pm$ 1.94} & 30.97 $\pm$ 1.17         \\ \cline{2-13} 
                                                                                                                  & \multicolumn{1}{c|}{\checkmark}   &              & \multicolumn{1}{c|}{\checkmark}   & \checkmark   & \multicolumn{1}{c|}{77.23 $\pm$ 3.02}          & \multicolumn{1}{c|}{78.52 $\pm$ 2.98}   & \multicolumn{1}{c|}{77.23 $\pm$ 3.02} & 17.98 $\pm$ 0.71         & \multicolumn{1}{c|}{78.28 $\pm$ 1.91}          & \multicolumn{1}{c|}{79.27 $\pm$ 1.51}   & \multicolumn{1}{c|}{78.28 $\pm$ 1.91} & 33.24 $\pm$ 2.58         \\ \cline{2-13} 
                                                                                                                  & \multicolumn{1}{c|}{\checkmark}   & \checkmark   & \multicolumn{1}{c|}{\checkmark}   &              & \multicolumn{1}{c|}{73.20 $\pm$ 3.03}          & \multicolumn{1}{c|}{75.22 $\pm$ 2.99}   & \multicolumn{1}{c|}{73.20 $\pm$ 3.03} & 17.62 $\pm$ 0.98         & \multicolumn{1}{c|}{74.00 $\pm$ 2.60}          & \multicolumn{1}{c|}{75.47 $\pm$ 2.55}   & \multicolumn{1}{c|}{74.00 $\pm$ 2.60} & 30.83 $\pm$ 1.24         \\ \cline{2-13} 
                                                                                                                  & \multicolumn{1}{c|}{\checkmark}   & \checkmark   & \multicolumn{1}{c|}{\checkmark}   & \checkmark   & \multicolumn{1}{c|}{76.22 $\pm$ 2.60}          & \multicolumn{1}{c|}{78.22 $\pm$ 2.40}   & \multicolumn{1}{c|}{76.22 $\pm$ 2.60} & 18.09 $\pm$ 0.94         & \multicolumn{1}{c|}{76.69 $\pm$ 1.80}          & \multicolumn{1}{c|}{78.21 $\pm$ 1.85}   & \multicolumn{1}{c|}{76.69 $\pm$ 1.80} & 32.69 $\pm$ 1.45         \\ \hline
\multirow{6}{*}{\textsc{\textbf{Mittens}}}                                                                        & \multicolumn{1}{c|}{\checkmark}   &              & \multicolumn{2}{c|}{Branch N/A}                  & \multicolumn{1}{c|}{70.89 $\pm$ 5.57}          & \multicolumn{1}{c|}{72.12 $\pm$ 5.29}   & \multicolumn{1}{c|}{70.89 $\pm$ 5.57} & 14.22 $\pm$ 2.33         & \multicolumn{1}{c|}{70.74 $\pm$ 2.78}          & \multicolumn{1}{c|}{71.67 $\pm$ 2.54}   & \multicolumn{1}{c|}{70.74 $\pm$ 2.78} & 25.91 $\pm$ 2.06         \\ \cline{2-13} 
                                                                                                                  & \multicolumn{1}{c|}{\checkmark}   & \checkmark   & \multicolumn{2}{c|}{Branch N/A}                  & \multicolumn{1}{c|}{69.78 $\pm$ 7.26}          & \multicolumn{1}{c|}{71.58 $\pm$ 6.25}   & \multicolumn{1}{c|}{69.78 $\pm$ 7.26} & 13.86 $\pm$ 2.08         & \multicolumn{1}{c|}{63.42 $\pm$ 8.50}          & \multicolumn{1}{c|}{64.21 $\pm$ 8.34}   & \multicolumn{1}{c|}{63.42 $\pm$ 8.50} & 22.79 $\pm$ 4.89         \\ \cline{2-13} 
                                                                                                                  & \multicolumn{1}{c|}{\checkmark}   &              & \multicolumn{1}{c|}{\checkmark}   &              & \multicolumn{1}{c|}{74.92 $\pm$ 3.34}          & \multicolumn{1}{c|}{75.79 $\pm$ 3.08}   & \multicolumn{1}{c|}{74.92 $\pm$ 3.34} & 17.64 $\pm$ 1.93         & \multicolumn{1}{c|}{74.90 $\pm$ 3.98}          & \multicolumn{1}{c|}{75.81 $\pm$ 3.94}   & \multicolumn{1}{c|}{74.90 $\pm$ 3.98} & 30.60 $\pm$ 2.88         \\ \cline{2-13} 
                                                                                                                  & \multicolumn{1}{c|}{\checkmark}   &              & \multicolumn{1}{c|}{\checkmark}   & \checkmark   & \multicolumn{1}{c|}{\resultlowclr{\textbf{77.70 $\pm$ 3.08}}} & \multicolumn{1}{c|}{78.41 $\pm$ 2.94}   & \multicolumn{1}{c|}{77.70 $\pm$ 3.08} & 17.61 $\pm$ 1.95         & \multicolumn{1}{c|}{\resultlowclr{\textbf{78.64 $\pm$ 1.69}}} & \multicolumn{1}{c|}{79.44 $\pm$ 2.05}   & \multicolumn{1}{c|}{78.64 $\pm$ 1.69} & 30.18 $\pm$ 3.58         \\ \cline{2-13} 
                                                                                                                  & \multicolumn{1}{c|}{\checkmark}   & \checkmark   & \multicolumn{1}{c|}{\checkmark}   &              & \multicolumn{1}{c|}{73.65 $\pm$ 5.04}          & \multicolumn{1}{c|}{74.97 $\pm$ 5.41}   & \multicolumn{1}{c|}{73.65 $\pm$ 5.04} & 16.88 $\pm$ 1.90         & \multicolumn{1}{c|}{72.80 $\pm$ 3.26}          & \multicolumn{1}{c|}{73.58 $\pm$ 3.29}   & \multicolumn{1}{c|}{72.80 $\pm$ 3.26} & 30.30 $\pm$ 1.63         \\ \cline{2-13} 
                                                                                                                  & \multicolumn{1}{c|}{\checkmark}   & \checkmark   & \multicolumn{1}{c|}{\checkmark}   & \checkmark   & \multicolumn{1}{c|}{75.68 $\pm$ 3.16}          & \multicolumn{1}{c|}{77.17 $\pm$ 3.14}   & \multicolumn{1}{c|}{75.68 $\pm$ 3.16} & 17.50 $\pm$ 1.36         & \multicolumn{1}{c|}{74.90 $\pm$ 3.50}          & \multicolumn{1}{c|}{76.12 $\pm$ 3.89}   & \multicolumn{1}{c|}{74.90 $\pm$ 3.50} & 30.43 $\pm$ 2.81         \\ \hline
\end{tabular}
}
\end{table*}


\begin{table*}[!htbp]
\centering

\caption{Ablation testing on the Twitter15 dataset~\cite{Ma2017} using [Bi]LSTM as [Bi]RNN units (bold marks the best accuracy w.r.t. word embedding)}
\label{tab:results_lstm_twitter15}
\resizebox{1\textwidth}{!}{%
\begin{tabular}{|c|cc|cc|cccc|cccc|}
\hline
\multicolumn{1}{|l|}{\multirow{3}{*}{\textbf{\begin{tabular}[c]{@{}c@{}}Word\\ Embedding\end{tabular}}}} & \multicolumn{4}{c|}{\textbf{Network Embedding}}  & \multicolumn{4}{c|}{\multirow{2}{*}{\textbf{GRU Cell}}}   & \multicolumn{4}{c|}{\multirow{2}{*}{\textbf{BiGRU Cell}}}                                                                                                                                           \\ \cline{2-5}
\multicolumn{1}{|l|}{}                               & \multicolumn{2}{c|}{\textbf{Text Branch}}     & \multicolumn{2}{c|}{\textbf{Social Branch}}     & \multicolumn{4}{c|}{} & \multicolumn{4}{c|}{}                                                                                                                                       \\ \cline{2-13} 
\multicolumn{1}{|l|}{}                                                                                                                 & \multicolumn{1}{c|}{\textbf{RNN}} & \textbf{CNN} & \multicolumn{1}{c|}{\textbf{RNN}} & \textbf{CNN} & \multicolumn{1}{c|}{\textbf{Accuracy}}         & \multicolumn{1}{c|}{\textbf{Precision}} & \multicolumn{1}{c|}{\textbf{Recall}}  & \multicolumn{1}{c|}{\textbf{Runtime(s)}} & \multicolumn{1}{c|}{\textbf{Accuracy}}         & \multicolumn{1}{c|}{\textbf{Precision}} & \multicolumn{1}{c|}{\textbf{Recall}}  & \multicolumn{1}{c|}{\textbf{Runtime(s)}} \\ \hline
\multirow{6}{*}{\begin{tabular}[c]{@{}c@{}}\textsc{\textbf{Word2Vec}}\\ \textbf{CBOW}\end{tabular}}               & \multicolumn{1}{c|}{\checkmark}   &              & \multicolumn{2}{c|}{Branch N/A}                  & \multicolumn{1}{c|}{73.24 $\pm$ 1.69}          & \multicolumn{1}{c|}{73.50 $\pm$ 1.78}   & \multicolumn{1}{c|}{73.24 $\pm$ 1.69} & 13.39 $\pm$ 0.88         & \multicolumn{1}{c|}{73.15 $\pm$ 1.69}          & \multicolumn{1}{c|}{73.37 $\pm$ 1.97}   & \multicolumn{1}{c|}{73.15 $\pm$ 1.69} & 23.27 $\pm$ 1.39         \\ \cline{2-13} 
                                                                                                                  & \multicolumn{1}{c|}{\checkmark}   & \checkmark   & \multicolumn{2}{c|}{Branch N/A}                  & \multicolumn{1}{c|}{72.26 $\pm$ 2.11}          & \multicolumn{1}{c|}{72.35 $\pm$ 2.24}   & \multicolumn{1}{c|}{72.26 $\pm$ 2.11} & 13.62 $\pm$ 0.75         & \multicolumn{1}{c|}{71.48 $\pm$ 2.07}          & \multicolumn{1}{c|}{71.65 $\pm$ 1.99}   & \multicolumn{1}{c|}{71.48 $\pm$ 2.07} & 22.39 $\pm$ 1.39         \\ \cline{2-13} 
                                                                                                                  & \multicolumn{1}{c|}{\checkmark}   &              & \multicolumn{1}{c|}{\checkmark}   &              & \multicolumn{1}{c|}{74.59 $\pm$ 1.61}          & \multicolumn{1}{c|}{74.98 $\pm$ 1.56}   & \multicolumn{1}{c|}{74.59 $\pm$ 1.61} & 16.19 $\pm$ 0.75         & \multicolumn{1}{c|}{74.74 $\pm$ 1.74}          & \multicolumn{1}{c|}{75.03 $\pm$ 1.82}   & \multicolumn{1}{c|}{74.74 $\pm$ 1.74} & 28.59 $\pm$ 1.40         \\ \cline{2-13} 
                                                                                                                  & \multicolumn{1}{c|}{\checkmark}   &              & \multicolumn{1}{c|}{\checkmark}   & \checkmark   & \multicolumn{1}{c|}{76.62 $\pm$ 2.48}          & \multicolumn{1}{c|}{76.88 $\pm$ 2.43}   & \multicolumn{1}{c|}{76.62 $\pm$ 2.48} & 16.63 $\pm$ 0.51         & \multicolumn{1}{c|}{77.09 $\pm$ 2.37}          & \multicolumn{1}{c|}{77.33 $\pm$ 2.32}   & \multicolumn{1}{c|}{77.09 $\pm$ 2.37} & 29.39 $\pm$ 0.81         \\ \cline{2-13} 
                                                                                                                  & \multicolumn{1}{c|}{\checkmark}   & \checkmark   & \multicolumn{1}{c|}{\checkmark}   &              & \multicolumn{1}{c|}{73.96 $\pm$ 1.67}          & \multicolumn{1}{c|}{74.38 $\pm$ 1.75}   & \multicolumn{1}{c|}{73.96 $\pm$ 1.67} & 15.98 $\pm$ 0.65         & \multicolumn{1}{c|}{73.24 $\pm$ 1.88}          & \multicolumn{1}{c|}{73.53 $\pm$ 1.98}   & \multicolumn{1}{c|}{73.24 $\pm$ 1.88} & 27.86 $\pm$ 0.61         \\ \cline{2-13} 
                                                                                                                  & \multicolumn{1}{c|}{\checkmark}   & \checkmark   & \multicolumn{1}{c|}{\checkmark}   & \checkmark   & \multicolumn{1}{c|}{76.49 $\pm$ 2.08}          & \multicolumn{1}{c|}{76.83 $\pm$ 2.11}   & \multicolumn{1}{c|}{76.49 $\pm$ 2.08} & 16.80 $\pm$ 0.63         & \multicolumn{1}{c|}{76.31 $\pm$ 1.80}          & \multicolumn{1}{c|}{76.71 $\pm$ 1.94}   & \multicolumn{1}{c|}{76.31 $\pm$ 1.80} & 29.46 $\pm$ 0.96         \\ \hline
\multirow{6}{*}{\begin{tabular}[c]{@{}c@{}}\textsc{\textbf{Word2Vec}}\\ \textsc{\textbf{Skip-Gran}}\end{tabular}} & \multicolumn{1}{c|}{\checkmark}   &              & \multicolumn{2}{c|}{Branch N/A}                  & \multicolumn{1}{c|}{70.04 $\pm$ 2.24}          & \multicolumn{1}{c|}{70.37 $\pm$ 2.24}   & \multicolumn{1}{c|}{70.04 $\pm$ 2.24} & 12.16 $\pm$ 0.93         & \multicolumn{1}{c|}{69.51 $\pm$ 2.52}          & \multicolumn{1}{c|}{69.65 $\pm$ 2.69}   & \multicolumn{1}{c|}{69.51 $\pm$ 2.52} & 21.05 $\pm$ 1.42         \\ \cline{2-13} 
                                                                                                                  & \multicolumn{1}{c|}{\checkmark}   & \checkmark   & \multicolumn{2}{c|}{Branch N/A}                  & \multicolumn{1}{c|}{69.60 $\pm$ 2.12}          & \multicolumn{1}{c|}{69.65 $\pm$ 2.14}   & \multicolumn{1}{c|}{69.60 $\pm$ 2.12} & 13.03 $\pm$ 0.81         & \multicolumn{1}{c|}{68.48 $\pm$ 1.98}          & \multicolumn{1}{c|}{68.43 $\pm$ 2.09}   & \multicolumn{1}{c|}{68.48 $\pm$ 1.98} & 21.44 $\pm$ 1.68         \\ \cline{2-13} 
                                                                                                                  & \multicolumn{1}{c|}{\checkmark}   &              & \multicolumn{1}{c|}{\checkmark}   &              & \multicolumn{1}{c|}{71.39 $\pm$ 2.28}          & \multicolumn{1}{c|}{71.79 $\pm$ 2.15}   & \multicolumn{1}{c|}{71.39 $\pm$ 2.28} & 15.16 $\pm$ 0.50         & \multicolumn{1}{c|}{70.18 $\pm$ 2.35}          & \multicolumn{1}{c|}{70.45 $\pm$ 2.47}   & \multicolumn{1}{c|}{70.18 $\pm$ 2.35} & 27.01 $\pm$ 1.61         \\ \cline{2-13} 
                                                                                                                  & \multicolumn{1}{c|}{\checkmark}   &              & \multicolumn{1}{c|}{\checkmark}   & \checkmark   & \multicolumn{1}{c|}{72.95 $\pm$ 2.69}          & \multicolumn{1}{c|}{73.32 $\pm$ 2.61}   & \multicolumn{1}{c|}{72.95 $\pm$ 2.69} & 15.36 $\pm$ 0.79         & \multicolumn{1}{c|}{73.51 $\pm$ 2.42}          & \multicolumn{1}{c|}{73.85 $\pm$ 2.39}   & \multicolumn{1}{c|}{73.51 $\pm$ 2.42} & 27.72 $\pm$ 1.22         \\ \cline{2-13} 
                                                                                                                  & \multicolumn{1}{c|}{\checkmark}   & \checkmark   & \multicolumn{1}{c|}{\checkmark}   &              & \multicolumn{1}{c|}{70.60 $\pm$ 2.87}          & \multicolumn{1}{c|}{70.77 $\pm$ 3.02}   & \multicolumn{1}{c|}{70.60 $\pm$ 2.87} & 15.44 $\pm$ 0.79         & \multicolumn{1}{c|}{70.65 $\pm$ 2.23}          & \multicolumn{1}{c|}{70.83 $\pm$ 2.31}   & \multicolumn{1}{c|}{70.65 $\pm$ 2.23} & 26.73 $\pm$ 1.48         \\ \cline{2-13} 
                                                                                                                  & \multicolumn{1}{c|}{\checkmark}   & \checkmark   & \multicolumn{1}{c|}{\checkmark}   & \checkmark   & \multicolumn{1}{c|}{73.51 $\pm$ 2.30}          & \multicolumn{1}{c|}{73.91 $\pm$ 2.22}   & \multicolumn{1}{c|}{73.51 $\pm$ 2.30} & 16.42 $\pm$ 0.39         & \multicolumn{1}{c|}{73.56 $\pm$ 1.65}          & \multicolumn{1}{c|}{74.01 $\pm$ 1.87}   & \multicolumn{1}{c|}{73.56 $\pm$ 1.65} & 28.97 $\pm$ 0.57         \\ \hline
\multirow{6}{*}{\begin{tabular}[c]{@{}c@{}}\textsc{\textbf{FastText}}\\ \textbf{CBOW}\end{tabular}}               & \multicolumn{1}{c|}{\checkmark}   &              & \multicolumn{2}{c|}{Branch N/A}                  & \multicolumn{1}{c|}{69.44 $\pm$ 2.99}          & \multicolumn{1}{c|}{69.53 $\pm$ 3.00}   & \multicolumn{1}{c|}{69.44 $\pm$ 2.99} & 12.85 $\pm$ 0.97         & \multicolumn{1}{c|}{69.62 $\pm$ 2.64}          & \multicolumn{1}{c|}{69.78 $\pm$ 2.55}   & \multicolumn{1}{c|}{69.62 $\pm$ 2.64} & 22.11 $\pm$ 2.05         \\ \cline{2-13} 
                                                                                                                  & \multicolumn{1}{c|}{\checkmark}   & \checkmark   & \multicolumn{2}{c|}{Branch N/A}                  & \multicolumn{1}{c|}{70.60 $\pm$ 2.87}          & \multicolumn{1}{c|}{71.22 $\pm$ 2.84}   & \multicolumn{1}{c|}{70.60 $\pm$ 2.87} & 13.41 $\pm$ 1.97         & \multicolumn{1}{c|}{67.34 $\pm$ 2.24}          & \multicolumn{1}{c|}{67.59 $\pm$ 2.23}   & \multicolumn{1}{c|}{67.34 $\pm$ 2.24} & 21.19 $\pm$ 1.91         \\ \cline{2-13} 
                                                                                                                  & \multicolumn{1}{c|}{\checkmark}   &              & \multicolumn{1}{c|}{\checkmark}   &              & \multicolumn{1}{c|}{70.43 $\pm$ 2.61}          & \multicolumn{1}{c|}{70.61 $\pm$ 2.63}   & \multicolumn{1}{c|}{70.43 $\pm$ 2.61} & 15.36 $\pm$ 1.22         & \multicolumn{1}{c|}{71.12 $\pm$ 2.71}          & \multicolumn{1}{c|}{71.31 $\pm$ 2.75}   & \multicolumn{1}{c|}{71.12 $\pm$ 2.71} & 28.24 $\pm$ 2.35         \\ \cline{2-13} 
                                                                                                                  & \multicolumn{1}{c|}{\checkmark}   &              & \multicolumn{1}{c|}{\checkmark}   & \checkmark   & \multicolumn{1}{c|}{73.62 $\pm$ 2.11}          & \multicolumn{1}{c|}{73.78 $\pm$ 2.15}   & \multicolumn{1}{c|}{73.62 $\pm$ 2.11} & 15.97 $\pm$ 0.60         & \multicolumn{1}{c|}{74.83 $\pm$ 2.45}          & \multicolumn{1}{c|}{75.06 $\pm$ 2.45}   & \multicolumn{1}{c|}{74.83 $\pm$ 2.45} & 28.61 $\pm$ 1.32         \\ \cline{2-13} 
                                                                                                                  & \multicolumn{1}{c|}{\checkmark}   & \checkmark   & \multicolumn{1}{c|}{\checkmark}   &              & \multicolumn{1}{c|}{70.54 $\pm$ 2.38}          & \multicolumn{1}{c|}{70.82 $\pm$ 2.27}   & \multicolumn{1}{c|}{70.54 $\pm$ 2.38} & 15.84 $\pm$ 1.11         & \multicolumn{1}{c|}{69.31 $\pm$ 3.76}          & \multicolumn{1}{c|}{69.69 $\pm$ 3.64}   & \multicolumn{1}{c|}{69.31 $\pm$ 3.76} & 27.05 $\pm$ 2.04         \\ \cline{2-13} 
                                                                                                                  & \multicolumn{1}{c|}{\checkmark}   & \checkmark   & \multicolumn{1}{c|}{\checkmark}   & \checkmark   & \multicolumn{1}{c|}{73.76 $\pm$ 2.20}          & \multicolumn{1}{c|}{74.00 $\pm$ 2.12}   & \multicolumn{1}{c|}{73.76 $\pm$ 2.20} & 17.12 $\pm$ 1.64         & \multicolumn{1}{c|}{73.78 $\pm$ 2.41}          & \multicolumn{1}{c|}{73.94 $\pm$ 2.31}   & \multicolumn{1}{c|}{73.78 $\pm$ 2.41} & 28.92 $\pm$ 1.38         \\ \hline
\multirow{6}{*}{\begin{tabular}[c]{@{}c@{}}\textsc{\textbf{FastText}}\\ \textsc{\textbf{Skip-Gran}}\end{tabular}} & \multicolumn{1}{c|}{\checkmark}   &              & \multicolumn{2}{c|}{Branch N/A}                  & \multicolumn{1}{c|}{71.30 $\pm$ 1.99}          & \multicolumn{1}{c|}{71.46 $\pm$ 2.14}   & \multicolumn{1}{c|}{71.30 $\pm$ 1.99} & 12.94 $\pm$ 1.17         & \multicolumn{1}{c|}{70.76 $\pm$ 2.67}          & \multicolumn{1}{c|}{71.08 $\pm$ 2.84}   & \multicolumn{1}{c|}{70.76 $\pm$ 2.67} & 21.34 $\pm$ 2.35         \\ \cline{2-13} 
                                                                                                                  & \multicolumn{1}{c|}{\checkmark}   & \checkmark   & \multicolumn{2}{c|}{Branch N/A}                  & \multicolumn{1}{c|}{70.87 $\pm$ 1.69}          & \multicolumn{1}{c|}{71.03 $\pm$ 1.72}   & \multicolumn{1}{c|}{70.87 $\pm$ 1.69} & 12.95 $\pm$ 0.51         & \multicolumn{1}{c|}{70.49 $\pm$ 1.89}          & \multicolumn{1}{c|}{70.76 $\pm$ 1.62}   & \multicolumn{1}{c|}{70.49 $\pm$ 1.89} & 21.48 $\pm$ 1.24         \\ \cline{2-13} 
                                                                                                                  & \multicolumn{1}{c|}{\checkmark}   &              & \multicolumn{1}{c|}{\checkmark}   &              & \multicolumn{1}{c|}{71.83 $\pm$ 2.11}          & \multicolumn{1}{c|}{72.08 $\pm$ 2.01}   & \multicolumn{1}{c|}{71.83 $\pm$ 2.11} & 15.43 $\pm$ 0.66         & \multicolumn{1}{c|}{72.06 $\pm$ 2.03}          & \multicolumn{1}{c|}{72.40 $\pm$ 1.89}   & \multicolumn{1}{c|}{72.06 $\pm$ 2.03} & 26.31 $\pm$ 1.73         \\ \cline{2-13} 
                                                                                                                  & \multicolumn{1}{c|}{\checkmark}   &              & \multicolumn{1}{c|}{\checkmark}   & \checkmark   & \multicolumn{1}{c|}{75.03 $\pm$ 2.32}          & \multicolumn{1}{c|}{75.38 $\pm$ 2.49}   & \multicolumn{1}{c|}{75.03 $\pm$ 2.32} & 15.90 $\pm$ 0.50         & \multicolumn{1}{c|}{74.85 $\pm$ 1.55}          & \multicolumn{1}{c|}{75.34 $\pm$ 1.80}   & \multicolumn{1}{c|}{74.85 $\pm$ 1.55} & 28.46 $\pm$ 1.61         \\ \cline{2-13} 
                                                                                                                  & \multicolumn{1}{c|}{\checkmark}   & \checkmark   & \multicolumn{1}{c|}{\checkmark}   &              & \multicolumn{1}{c|}{71.66 $\pm$ 2.43}          & \multicolumn{1}{c|}{72.07 $\pm$ 2.52}   & \multicolumn{1}{c|}{71.66 $\pm$ 2.43} & 16.17 $\pm$ 0.94         & \multicolumn{1}{c|}{72.10 $\pm$ 1.35}          & \multicolumn{1}{c|}{72.55 $\pm$ 1.43}   & \multicolumn{1}{c|}{72.10 $\pm$ 1.35} & 28.02 $\pm$ 1.96         \\ \cline{2-13} 
                                                                                                                  & \multicolumn{1}{c|}{\checkmark}   & \checkmark   & \multicolumn{1}{c|}{\checkmark}   & \checkmark   & \multicolumn{1}{c|}{74.63 $\pm$ 1.52}          & \multicolumn{1}{c|}{75.16 $\pm$ 1.65}   & \multicolumn{1}{c|}{74.63 $\pm$ 1.52} & 16.23 $\pm$ 0.94         & \multicolumn{1}{c|}{74.65 $\pm$ 2.42}          & \multicolumn{1}{c|}{75.28 $\pm$ 2.74}   & \multicolumn{1}{c|}{74.65 $\pm$ 2.42} & 29.80 $\pm$ 1.20         \\ \hline
\multirow{6}{*}{\textsc{\textbf{GloVe}}}                                                                          & \multicolumn{1}{c|}{\checkmark}   &              & \multicolumn{2}{c|}{Branch N/A}                  & \multicolumn{1}{c|}{74.99 $\pm$ 2.56}          & \multicolumn{1}{c|}{75.92 $\pm$ 2.25}   & \multicolumn{1}{c|}{74.99 $\pm$ 2.56} & 15.92 $\pm$ 0.89         & \multicolumn{1}{c|}{72.75 $\pm$ 2.58}          & \multicolumn{1}{c|}{73.86 $\pm$ 2.56}   & \multicolumn{1}{c|}{72.75 $\pm$ 2.58} & 27.24 $\pm$ 1.10         \\ \cline{2-13} 
                                                                                                                  & \multicolumn{1}{c|}{\checkmark}   & \checkmark   & \multicolumn{2}{c|}{Branch N/A}                  & \multicolumn{1}{c|}{74.38 $\pm$ 1.50}          & \multicolumn{1}{c|}{75.98 $\pm$ 1.19}   & \multicolumn{1}{c|}{74.38 $\pm$ 1.50} & 15.91 $\pm$ 1.66         & \multicolumn{1}{c|}{72.39 $\pm$ 2.21}          & \multicolumn{1}{c|}{73.91 $\pm$ 1.94}   & \multicolumn{1}{c|}{72.39 $\pm$ 2.21} & 27.17 $\pm$ 0.81         \\ \cline{2-13} 
                                                                                                                  & \multicolumn{1}{c|}{\checkmark}   &              & \multicolumn{1}{c|}{\checkmark}   &              & \multicolumn{1}{c|}{74.54 $\pm$ 2.17}          & \multicolumn{1}{c|}{75.14 $\pm$ 2.33}   & \multicolumn{1}{c|}{74.54 $\pm$ 2.17} & 17.81 $\pm$ 1.31         & \multicolumn{1}{c|}{75.59 $\pm$ 2.05}          & \multicolumn{1}{c|}{76.16 $\pm$ 1.78}   & \multicolumn{1}{c|}{75.59 $\pm$ 2.05} & 32.95 $\pm$ 2.97         \\ \cline{2-13} 
                                                                                                                  & \multicolumn{1}{c|}{\checkmark}   &              & \multicolumn{1}{c|}{\checkmark}   & \checkmark   & \multicolumn{1}{c|}{76.51 $\pm$ 1.94}          & \multicolumn{1}{c|}{77.58 $\pm$ 1.95}   & \multicolumn{1}{c|}{76.51 $\pm$ 1.94} & 17.79 $\pm$ 1.12         & \multicolumn{1}{c|}{76.80 $\pm$ 2.35}          & \multicolumn{1}{c|}{77.55 $\pm$ 2.32}   & \multicolumn{1}{c|}{76.80 $\pm$ 2.35} & 33.21 $\pm$ 2.46         \\ \cline{2-13} 
                                                                                                                  & \multicolumn{1}{c|}{\checkmark}   & \checkmark   & \multicolumn{1}{c|}{\checkmark}   &              & \multicolumn{1}{c|}{73.76 $\pm$ 2.22}          & \multicolumn{1}{c|}{75.36 $\pm$ 2.69}   & \multicolumn{1}{c|}{73.76 $\pm$ 2.22} & 18.39 $\pm$ 1.11         & \multicolumn{1}{c|}{73.83 $\pm$ 1.81}          & \multicolumn{1}{c|}{75.38 $\pm$ 2.04}   & \multicolumn{1}{c|}{73.83 $\pm$ 1.81} & 33.03 $\pm$ 1.52         \\ \cline{2-13} 
                                                                                                                  & \multicolumn{1}{c|}{\checkmark}   & \checkmark   & \multicolumn{1}{c|}{\checkmark}   & \checkmark   & \multicolumn{1}{c|}{77.09 $\pm$ 2.09}          & \multicolumn{1}{c|}{77.91 $\pm$ 2.31}   & \multicolumn{1}{c|}{77.09 $\pm$ 2.09} & 19.88 $\pm$ 1.86         & \multicolumn{1}{c|}{75.55 $\pm$ 2.68}          & \multicolumn{1}{c|}{77.22 $\pm$ 1.61}   & \multicolumn{1}{c|}{75.55 $\pm$ 2.68} & 34.22 $\pm$ 2.67         \\ \hline
\multirow{6}{*}{\textsc{\textbf{Mittens}}}                                                                        & \multicolumn{1}{c|}{\checkmark}   &              & \multicolumn{2}{c|}{Branch N/A}                  & \multicolumn{1}{c|}{73.40 $\pm$ 1.31}          & \multicolumn{1}{c|}{74.20 $\pm$ 1.27}   & \multicolumn{1}{c|}{73.40 $\pm$ 1.31} & 16.72 $\pm$ 1.00         & \multicolumn{1}{c|}{73.04 $\pm$ 1.27}          & \multicolumn{1}{c|}{73.97 $\pm$ 1.50}   & \multicolumn{1}{c|}{73.04 $\pm$ 1.27} & 28.38 $\pm$ 2.27         \\ \cline{2-13} 
                                                                                                                  & \multicolumn{1}{c|}{\checkmark}   & \checkmark   & \multicolumn{2}{c|}{Branch N/A}                  & \multicolumn{1}{c|}{72.17 $\pm$ 2.06}          & \multicolumn{1}{c|}{73.30 $\pm$ 2.34}   & \multicolumn{1}{c|}{72.17 $\pm$ 2.06} & 16.59 $\pm$ 1.09         & \multicolumn{1}{c|}{72.26 $\pm$ 2.27}          & \multicolumn{1}{c|}{73.73 $\pm$ 1.90}   & \multicolumn{1}{c|}{72.26 $\pm$ 2.27} & 29.46 $\pm$ 1.31         \\ \cline{2-13} 
                                                                                                                  & \multicolumn{1}{c|}{\checkmark}   &              & \multicolumn{1}{c|}{\checkmark}   &              & \multicolumn{1}{c|}{74.79 $\pm$ 1.99}          & \multicolumn{1}{c|}{75.63 $\pm$ 1.81}   & \multicolumn{1}{c|}{74.79 $\pm$ 1.99} & 19.29 $\pm$ 0.49         & \multicolumn{1}{c|}{74.77 $\pm$ 2.24}          & \multicolumn{1}{c|}{75.58 $\pm$ 2.42}   & \multicolumn{1}{c|}{74.77 $\pm$ 2.24} & 36.06 $\pm$ 2.95         \\ \cline{2-13} 
                                                                                                                  & \multicolumn{1}{c|}{\checkmark}   &              & \multicolumn{1}{c|}{\checkmark}   & \checkmark   & \multicolumn{1}{c|}{\resultlowclr{\textbf{77.90 $\pm$ 1.18}}} & \multicolumn{1}{c|}{78.56 $\pm$ 1.18}   & \multicolumn{1}{c|}{77.90 $\pm$ 1.18} & 19.56 $\pm$ 1.45         & \multicolumn{1}{c|}{\resultlowclr{\textbf{77.44 $\pm$ 1.47}}} & \multicolumn{1}{c|}{78.81 $\pm$ 1.19}   & \multicolumn{1}{c|}{77.44 $\pm$ 1.47} & 35.29 $\pm$ 3.09         \\ \cline{2-13} 
                                                                                                                  & \multicolumn{1}{c|}{\checkmark}   & \checkmark   & \multicolumn{1}{c|}{\checkmark}   &              & \multicolumn{1}{c|}{75.46 $\pm$ 1.81}          & \multicolumn{1}{c|}{76.30 $\pm$ 1.78}   & \multicolumn{1}{c|}{75.46 $\pm$ 1.81} & 19.65 $\pm$ 0.81         & \multicolumn{1}{c|}{73.62 $\pm$ 2.15}          & \multicolumn{1}{c|}{74.86 $\pm$ 2.21}   & \multicolumn{1}{c|}{73.62 $\pm$ 2.15} & 35.32 $\pm$ 2.26         \\ \cline{2-13} 
                                                                                                                  & \multicolumn{1}{c|}{\checkmark}   & \checkmark   & \multicolumn{1}{c|}{\checkmark}   & \checkmark   & \multicolumn{1}{c|}{76.49 $\pm$ 1.92}          & \multicolumn{1}{c|}{77.35 $\pm$ 2.08}   & \multicolumn{1}{c|}{76.49 $\pm$ 1.92} & 20.21 $\pm$ 1.14         & \multicolumn{1}{c|}{76.73 $\pm$ 2.63}          & \multicolumn{1}{c|}{77.88 $\pm$ 2.09}   & \multicolumn{1}{c|}{76.73 $\pm$ 2.63} & 36.01 $\pm$ 2.54         \\ \hline
\end{tabular}
}
\end{table*}

When using the GRU as the RNN cell, the best model, i.e., the one that employs a GRU layer for Text Branch and a GRU layer followed by a CNN layer for the Social Branch, obtains an accuracy of 77.70\%.
When using BiGRU for the same configuration, the accuracy increases to 78.64\%.
When using LSTM as RNN cells, the accuracy of the best model, i.e., the one that employs an LSTM layer for Text Branch and an LSTM layer followed by a CNN layer for the Social Branch, obtains an accuracy of 77.90\%.
The best model with BiLSTM, i.e., the one that employs a BiLSTM layer for Text Branch and a BiLSTM layer followed by a CNN layer for the Social Branch, obtains an accuracy of 77.44\%.

We observe that similar patterns emerge as in the case of the BuzzFace dataset and that the Social Branch, although measuring other network features, improves the overall performance of the models when trying to detect fake news.

\subsubsection{Twitter16 results}

Tables \ref{tab:results_gru_twitter16} and~\ref{tab:results_lstm_twitter16} present
the experimental results obtained on Twitter16.
For this set of experiments, we observe that the models that have the overall best performance when employing [Bi]GRU as [Bi]RNN cells for creating the Network Embedding are the ones that use \textsc{FastText Skip-Gram} as the embeddings model.
For the models that employ [Bi]LSTM as [Bi]RNN cells, the overall best performance is obtained when using the \textsc{FastText CBOW} word embedding model.
In this case, \textsc{Mittens} is outperformed by \textsc{FastText} because of the size of the vocabulary and maximum length of the tweets (Table \ref{tab:data_stats}).
Both these characteristics influence the quality of the learned word embeddings.
\textsc{FastText}, by using character n-grams, manages to learn sub-words embeddings that better represent small length documents within small size datasets.
In contrast, \textsc{Mittens}, as well as \textsc{GloVe}, uses global corpus statistics to learn word representations.
For small size datasets with small length documents, these global corpus statistics impact negatively the quality of the embeddings.

\begin{table*}[!htbp]
\centering
\caption{Ablation testing on the Twitter16 dataset~\cite{Ma2017} using [Bi]GRU as [Bi]RNN units (bold marks the best accuracy w.r.t. word embedding)}
\label{tab:results_gru_twitter16}
\resizebox{1\textwidth}{!}{%
\begin{tabular}{|c|cc|cc|cccc|cccc|}
\hline
\multicolumn{1}{|l|}{\multirow{3}{*}{\textbf{\begin{tabular}[c]{@{}c@{}}Word\\ Embedding\end{tabular}}}} & \multicolumn{4}{c|}{\textbf{Network Embedding}}  & \multicolumn{4}{c|}{\multirow{2}{*}{\textbf{GRU Cell}}}   & \multicolumn{4}{c|}{\multirow{2}{*}{\textbf{BiGRU Cell}}}                                                                                                                                           \\ \cline{2-5}
\multicolumn{1}{|l|}{}                               & \multicolumn{2}{c|}{\textbf{Text Branch}}     & \multicolumn{2}{c|}{\textbf{Social Branch}}     & \multicolumn{4}{c|}{} & \multicolumn{4}{c|}{}                                                                                                                                       \\ \cline{2-13} 
\multicolumn{1}{|l|}{}                                                                                                                 & \multicolumn{1}{c|}{\textbf{RNN}} & \textbf{CNN} & \multicolumn{1}{c|}{\textbf{RNN}} & \textbf{CNN} & \multicolumn{1}{c|}{\textbf{Accuracy}}         & \multicolumn{1}{c|}{\textbf{Precision}} & \multicolumn{1}{c|}{\textbf{Recall}}  & \multicolumn{1}{c|}{\textbf{Runtime(s)}} & \multicolumn{1}{c|}{\textbf{Accuracy}}         & \multicolumn{1}{c|}{\textbf{Precision}} & \multicolumn{1}{c|}{\textbf{Recall}}  & \multicolumn{1}{c|}{\textbf{Runtime(s)}} \\ \hline
\multirow{6}{*}{\begin{tabular}[c]{@{}c@{}}\textsc{\textbf{Word2Vec}}\\ \textbf{CBOW}\end{tabular}}               & \multicolumn{1}{c|}{\checkmark}   &              & \multicolumn{2}{c|}{Branch N/A}                  & \multicolumn{1}{c|}{71.30 $\pm$ 2.02}          & \multicolumn{1}{c|}{71.40 $\pm$ 2.17}   & \multicolumn{1}{c|}{71.30 $\pm$ 2.02} & 8.43 $\pm$ 0.97          & \multicolumn{1}{c|}{69.43 $\pm$ 2.49}          & \multicolumn{1}{c|}{69.56 $\pm$ 2.63}   & \multicolumn{1}{c|}{69.43 $\pm$ 2.49} & 14.06 $\pm$ 1.13         \\ \cline{2-13} 
                                                                                                                  & \multicolumn{1}{c|}{\checkmark}   & \checkmark   & \multicolumn{2}{c|}{Branch N/A}                  & \multicolumn{1}{c|}{70.24 $\pm$ 2.30}          & \multicolumn{1}{c|}{70.42 $\pm$ 2.40}   & \multicolumn{1}{c|}{70.24 $\pm$ 2.30} & 9.24 $\pm$ 0.78          & \multicolumn{1}{c|}{67.93 $\pm$ 2.14}          & \multicolumn{1}{c|}{67.99 $\pm$ 2.41}   & \multicolumn{1}{c|}{67.93 $\pm$ 2.14} & 13.08 $\pm$ 0.75         \\ \cline{2-13} 
                                                                                                                  & \multicolumn{1}{c|}{\checkmark}   &              & \multicolumn{1}{c|}{\checkmark}   &              & \multicolumn{1}{c|}{73.66 $\pm$ 1.20}          & \multicolumn{1}{c|}{73.78 $\pm$ 1.30}   & \multicolumn{1}{c|}{73.66 $\pm$ 1.20} & 10.89 $\pm$ 0.86         & \multicolumn{1}{c|}{72.32 $\pm$ 2.92}          & \multicolumn{1}{c|}{72.47 $\pm$ 3.28}   & \multicolumn{1}{c|}{72.32 $\pm$ 2.92} & 18.98 $\pm$ 1.31         \\ \cline{2-13} 
                                                                                                                  & \multicolumn{1}{c|}{\checkmark}   &              & \multicolumn{1}{c|}{\checkmark}   & \checkmark   & \multicolumn{1}{c|}{75.53 $\pm$ 2.38}          & \multicolumn{1}{c|}{75.84 $\pm$ 2.55}   & \multicolumn{1}{c|}{75.53 $\pm$ 2.38} & 11.68 $\pm$ 0.78         & \multicolumn{1}{c|}{75.37 $\pm$ 1.48}          & \multicolumn{1}{c|}{75.56 $\pm$ 1.45}   & \multicolumn{1}{c|}{75.37 $\pm$ 1.48} & 18.59 $\pm$ 1.20         \\ \cline{2-13} 
                                                                                                                  & \multicolumn{1}{c|}{\checkmark}   & \checkmark   & \multicolumn{1}{c|}{\checkmark}   &              & \multicolumn{1}{c|}{73.78 $\pm$ 1.42}          & \multicolumn{1}{c|}{73.90 $\pm$ 1.38}   & \multicolumn{1}{c|}{73.78 $\pm$ 1.42} & 11.36 $\pm$ 1.01         & \multicolumn{1}{c|}{71.91 $\pm$ 1.79}          & \multicolumn{1}{c|}{72.12 $\pm$ 1.79}   & \multicolumn{1}{c|}{71.91 $\pm$ 1.79} & 17.94 $\pm$ 0.95         \\ \cline{2-13} 
                                                                                                                  & \multicolumn{1}{c|}{\checkmark}   & \checkmark   & \multicolumn{1}{c|}{\checkmark}   & \checkmark   & \multicolumn{1}{c|}{77.20 $\pm$ 1.54}          & \multicolumn{1}{c|}{77.47 $\pm$ 1.52}   & \multicolumn{1}{c|}{77.20 $\pm$ 1.54} & 11.77 $\pm$ 1.15         & \multicolumn{1}{c|}{75.61 $\pm$ 1.76}          & \multicolumn{1}{c|}{75.91 $\pm$ 1.73}   & \multicolumn{1}{c|}{75.61 $\pm$ 1.76} & 18.73 $\pm$ 0.88         \\ \hline
\multirow{6}{*}{\begin{tabular}[c]{@{}c@{}}\textsc{\textbf{Word2Vec}}\\ \textsc{\textbf{Skip-Gran}}\end{tabular}} & \multicolumn{1}{c|}{\checkmark}   &              & \multicolumn{2}{c|}{Branch N/A}                  & \multicolumn{1}{c|}{71.79 $\pm$ 3.00}          & \multicolumn{1}{c|}{71.77 $\pm$ 3.19}   & \multicolumn{1}{c|}{71.79 $\pm$ 3.00} & 8.38 $\pm$ 0.59          & \multicolumn{1}{c|}{71.54 $\pm$ 2.80}          & \multicolumn{1}{c|}{71.76 $\pm$ 3.15}   & \multicolumn{1}{c|}{71.54 $\pm$ 2.80} & 13.63 $\pm$ 0.89         \\ \cline{2-13} 
                                                                                                                  & \multicolumn{1}{c|}{\checkmark}   & \checkmark   & \multicolumn{2}{c|}{Branch N/A}                  & \multicolumn{1}{c|}{71.63 $\pm$ 2.85}          & \multicolumn{1}{c|}{71.86 $\pm$ 3.07}   & \multicolumn{1}{c|}{71.63 $\pm$ 2.85} & 8.62 $\pm$ 0.98          & \multicolumn{1}{c|}{69.35 $\pm$ 1.69}          & \multicolumn{1}{c|}{69.52 $\pm$ 1.93}   & \multicolumn{1}{c|}{69.35 $\pm$ 1.69} & 13.46 $\pm$ 0.76         \\ \cline{2-13} 
                                                                                                                  & \multicolumn{1}{c|}{\checkmark}   &              & \multicolumn{1}{c|}{\checkmark}   &              & \multicolumn{1}{c|}{73.05 $\pm$ 2.01}          & \multicolumn{1}{c|}{73.17 $\pm$ 2.12}   & \multicolumn{1}{c|}{73.05 $\pm$ 2.01} & 10.99 $\pm$ 0.81         & \multicolumn{1}{c|}{73.21 $\pm$ 2.74}          & \multicolumn{1}{c|}{73.38 $\pm$ 2.89}   & \multicolumn{1}{c|}{73.21 $\pm$ 2.74} & 18.89 $\pm$ 0.71         \\ \cline{2-13} 
                                                                                                                  & \multicolumn{1}{c|}{\checkmark}   &              & \multicolumn{1}{c|}{\checkmark}   & \checkmark   & \multicolumn{1}{c|}{74.96 $\pm$ 2.36}          & \multicolumn{1}{c|}{75.15 $\pm$ 2.39}   & \multicolumn{1}{c|}{74.96 $\pm$ 2.36} & 10.96 $\pm$ 0.47         & \multicolumn{1}{c|}{75.28 $\pm$ 1.89}          & \multicolumn{1}{c|}{75.36 $\pm$ 1.93}   & \multicolumn{1}{c|}{75.28 $\pm$ 1.89} & 18.80 $\pm$ 0.71         \\ \cline{2-13} 
                                                                                                                  & \multicolumn{1}{c|}{\checkmark}   & \checkmark   & \multicolumn{1}{c|}{\checkmark}   &              & \multicolumn{1}{c|}{73.29 $\pm$ 2.13}          & \multicolumn{1}{c|}{73.47 $\pm$ 2.32}   & \multicolumn{1}{c|}{73.29 $\pm$ 2.13} & 11.56 $\pm$ 0.85         & \multicolumn{1}{c|}{72.64 $\pm$ 2.23}          & \multicolumn{1}{c|}{73.20 $\pm$ 2.32}   & \multicolumn{1}{c|}{72.64 $\pm$ 2.23} & 18.33 $\pm$ 0.96         \\ \cline{2-13} 
                                                                                                                  & \multicolumn{1}{c|}{\checkmark}   & \checkmark   & \multicolumn{1}{c|}{\checkmark}   & \checkmark   & \multicolumn{1}{c|}{75.85 $\pm$ 2.00}          & \multicolumn{1}{c|}{76.00 $\pm$ 2.03}   & \multicolumn{1}{c|}{75.85 $\pm$ 2.00} & 11.98 $\pm$ 0.83         & \multicolumn{1}{c|}{75.00 $\pm$ 1.28}          & \multicolumn{1}{c|}{75.32 $\pm$ 1.20}   & \multicolumn{1}{c|}{75.00 $\pm$ 1.28} & 19.30 $\pm$ 1.24         \\ \hline
\multirow{6}{*}{\begin{tabular}[c]{@{}c@{}}\textsc{\textbf{FastText}}\\ \textbf{CBOW}\end{tabular}}               & \multicolumn{1}{c|}{\checkmark}   &              & \multicolumn{2}{c|}{Branch N/A}                  & \multicolumn{1}{c|}{67.85 $\pm$ 4.22}          & \multicolumn{1}{c|}{68.29 $\pm$ 4.25}   & \multicolumn{1}{c|}{67.85 $\pm$ 4.22} & 8.80 $\pm$ 0.96          & \multicolumn{1}{c|}{66.59 $\pm$ 4.08}          & \multicolumn{1}{c|}{67.02 $\pm$ 4.11}   & \multicolumn{1}{c|}{66.59 $\pm$ 4.08} & 13.73 $\pm$ 1.18         \\ \cline{2-13} 
                                                                                                                  & \multicolumn{1}{c|}{\checkmark}   & \checkmark   & \multicolumn{2}{c|}{Branch N/A}                  & \multicolumn{1}{c|}{67.64 $\pm$ 3.46}          & \multicolumn{1}{c|}{67.98 $\pm$ 3.60}   & \multicolumn{1}{c|}{67.64 $\pm$ 3.46} & 9.17 $\pm$ 0.99          & \multicolumn{1}{c|}{64.27 $\pm$ 3.29}          & \multicolumn{1}{c|}{64.68 $\pm$ 3.50}   & \multicolumn{1}{c|}{64.27 $\pm$ 3.29} & 12.31 $\pm$ 0.93         \\ \cline{2-13} 
                                                                                                                  & \multicolumn{1}{c|}{\checkmark}   &              & \multicolumn{1}{c|}{\checkmark}   &              & \multicolumn{1}{c|}{70.98 $\pm$ 2.81}          & \multicolumn{1}{c|}{71.21 $\pm$ 2.93}   & \multicolumn{1}{c|}{70.98 $\pm$ 2.81} & 11.32 $\pm$ 0.65         & \multicolumn{1}{c|}{68.82 $\pm$ 3.01}          & \multicolumn{1}{c|}{68.95 $\pm$ 3.15}   & \multicolumn{1}{c|}{68.82 $\pm$ 3.01} & 18.57 $\pm$ 1.28         \\ \cline{2-13} 
                                                                                                                  & \multicolumn{1}{c|}{\checkmark}   &              & \multicolumn{1}{c|}{\checkmark}   & \checkmark   & \multicolumn{1}{c|}{73.17 $\pm$ 3.82}          & \multicolumn{1}{c|}{73.67 $\pm$ 4.10}   & \multicolumn{1}{c|}{73.17 $\pm$ 3.82} & 11.49 $\pm$ 0.63         & \multicolumn{1}{c|}{71.99 $\pm$ 2.54}          & \multicolumn{1}{c|}{72.37 $\pm$ 2.72}   & \multicolumn{1}{c|}{71.99 $\pm$ 2.54} & 18.86 $\pm$ 1.01         \\ \cline{2-13} 
                                                                                                                  & \multicolumn{1}{c|}{\checkmark}   & \checkmark   & \multicolumn{1}{c|}{\checkmark}   &              & \multicolumn{1}{c|}{70.65 $\pm$ 2.46}          & \multicolumn{1}{c|}{70.97 $\pm$ 2.43}   & \multicolumn{1}{c|}{70.65 $\pm$ 2.46} & 11.25 $\pm$ 0.75         & \multicolumn{1}{c|}{68.74 $\pm$ 3.99}          & \multicolumn{1}{c|}{68.92 $\pm$ 4.00}   & \multicolumn{1}{c|}{68.74 $\pm$ 3.99} & 18.33 $\pm$ 1.20         \\ \cline{2-13} 
                                                                                                                  & \multicolumn{1}{c|}{\checkmark}   & \checkmark   & \multicolumn{1}{c|}{\checkmark}   & \checkmark   & \multicolumn{1}{c|}{74.55 $\pm$ 2.74}          & \multicolumn{1}{c|}{75.09 $\pm$ 2.79}   & \multicolumn{1}{c|}{74.55 $\pm$ 2.74} & 11.79 $\pm$ 0.76         & \multicolumn{1}{c|}{73.25 $\pm$ 2.83}          & \multicolumn{1}{c|}{73.53 $\pm$ 2.84}   & \multicolumn{1}{c|}{73.25 $\pm$ 2.83} & 18.86 $\pm$ 1.05         \\ \hline
\multirow{6}{*}{\begin{tabular}[c]{@{}c@{}}\textsc{\textbf{FastText}}\\ \textsc{\textbf{Skip-Gran}}\end{tabular}} & \multicolumn{1}{c|}{\checkmark}   &              & \multicolumn{2}{c|}{Branch N/A}                  & \multicolumn{1}{c|}{72.40 $\pm$ 2.32}          & \multicolumn{1}{c|}{72.76 $\pm$ 2.48}   & \multicolumn{1}{c|}{72.40 $\pm$ 2.32} & 8.51 $\pm$ 0.75          & \multicolumn{1}{c|}{70.41 $\pm$ 1.95}          & \multicolumn{1}{c|}{70.49 $\pm$ 1.71}   & \multicolumn{1}{c|}{70.41 $\pm$ 1.95} & 13.53 $\pm$ 0.63         \\ \cline{2-13} 
                                                                                                                  & \multicolumn{1}{c|}{\checkmark}   & \checkmark   & \multicolumn{2}{c|}{Branch N/A}                  & \multicolumn{1}{c|}{71.38 $\pm$ 2.15}          & \multicolumn{1}{c|}{71.51 $\pm$ 2.02}   & \multicolumn{1}{c|}{71.38 $\pm$ 2.15} & 8.31 $\pm$ 0.68          & \multicolumn{1}{c|}{69.63 $\pm$ 2.02}          & \multicolumn{1}{c|}{69.91 $\pm$ 2.01}   & \multicolumn{1}{c|}{69.63 $\pm$ 2.02} & 13.32 $\pm$ 0.59         \\ \cline{2-13} 
                                                                                                                  & \multicolumn{1}{c|}{\checkmark}   &              & \multicolumn{1}{c|}{\checkmark}   &              & \multicolumn{1}{c|}{75.12 $\pm$ 1.90}          & \multicolumn{1}{c|}{75.53 $\pm$ 1.85}   & \multicolumn{1}{c|}{75.12 $\pm$ 1.90} & 10.93 $\pm$ 0.86         & \multicolumn{1}{c|}{75.53 $\pm$ 2.75}          & \multicolumn{1}{c|}{75.74 $\pm$ 2.82}   & \multicolumn{1}{c|}{75.53 $\pm$ 2.75} & 18.79 $\pm$ 1.25         \\ \cline{2-13} 
                                                                                                                  & \multicolumn{1}{c|}{\checkmark}   &              & \multicolumn{1}{c|}{\checkmark}   & \checkmark   & \multicolumn{1}{c|}{76.95 $\pm$ 2.36}          & \multicolumn{1}{c|}{77.44 $\pm$ 2.27}   & \multicolumn{1}{c|}{76.95 $\pm$ 2.36} & 10.93 $\pm$ 1.04         & \multicolumn{1}{c|}{77.14 $\pm$ 2.48}          & \multicolumn{1}{c|}{77.34 $\pm$ 2.53}   & \multicolumn{1}{c|}{77.15 $\pm$ 2.48} & 19.09 $\pm$ 1.31         \\ \cline{2-13} 
                                                                                                                  & \multicolumn{1}{c|}{\checkmark}   & \checkmark   & \multicolumn{1}{c|}{\checkmark}   &              & \multicolumn{1}{c|}{74.27 $\pm$ 2.43}          & \multicolumn{1}{c|}{74.54 $\pm$ 2.26}   & \multicolumn{1}{c|}{74.27 $\pm$ 2.43} & 11.03 $\pm$ 1.14         & \multicolumn{1}{c|}{74.15 $\pm$ 2.16}          & \multicolumn{1}{c|}{74.40 $\pm$ 2.24}   & \multicolumn{1}{c|}{74.15 $\pm$ 2.16} & 18.85 $\pm$ 1.08         \\ \cline{2-13} 
                                                                                                                  & \multicolumn{1}{c|}{\checkmark}   & \checkmark   & \multicolumn{1}{c|}{\checkmark}   & \checkmark   & \multicolumn{1}{c|}{\resultlowclr{\textbf{77.40 $\pm$ 1.93}}} & \multicolumn{1}{c|}{77.80 $\pm$ 1.94}   & \multicolumn{1}{c|}{77.40 $\pm$ 1.93} & 11.40 $\pm$ 1.03         & \multicolumn{1}{c|}{\resultlowclr{\textbf{77.15 $\pm$ 2.62}}} & \multicolumn{1}{c|}{77.55 $\pm$ 2.65}   & \multicolumn{1}{c|}{77.15 $\pm$ 2.62} & 19.36 $\pm$ 1.55         \\ \hline
\multirow{6}{*}{\textsc{\textbf{GloVe}}}                                                                          & \multicolumn{1}{c|}{\checkmark}   &              & \multicolumn{2}{c|}{Branch N/A}                  & \multicolumn{1}{c|}{69.67 $\pm$ 2.83}          & \multicolumn{1}{c|}{73.18 $\pm$ 2.47}   & \multicolumn{1}{c|}{69.67 $\pm$ 2.83} & 9.53 $\pm$ 0.74          & \multicolumn{1}{c|}{68.78 $\pm$ 5.00}          & \multicolumn{1}{c|}{71.29 $\pm$ 5.24}   & \multicolumn{1}{c|}{68.78 $\pm$ 5.00} & 16.10 $\pm$ 1.89         \\ \cline{2-13} 
                                                                                                                  & \multicolumn{1}{c|}{\checkmark}   & \checkmark   & \multicolumn{2}{c|}{Branch N/A}                  & \multicolumn{1}{c|}{67.76 $\pm$ 5.22}          & \multicolumn{1}{c|}{71.15 $\pm$ 4.23}   & \multicolumn{1}{c|}{67.76 $\pm$ 5.22} & 9.75 $\pm$ 0.50          & \multicolumn{1}{c|}{69.84 $\pm$ 4.84}          & \multicolumn{1}{c|}{71.92 $\pm$ 4.82}   & \multicolumn{1}{c|}{69.84 $\pm$ 4.84} & 15.88 $\pm$ 1.67         \\ \cline{2-13} 
                                                                                                                  & \multicolumn{1}{c|}{\checkmark}   &              & \multicolumn{1}{c|}{\checkmark}   &              & \multicolumn{1}{c|}{72.44 $\pm$ 3.78}          & \multicolumn{1}{c|}{75.04 $\pm$ 4.32}   & \multicolumn{1}{c|}{72.44 $\pm$ 3.78} & 11.98 $\pm$ 0.52         & \multicolumn{1}{c|}{72.85 $\pm$ 2.87}          & \multicolumn{1}{c|}{75.01 $\pm$ 3.51}   & \multicolumn{1}{c|}{72.85 $\pm$ 2.87} & 21.27 $\pm$ 0.78         \\ \cline{2-13} 
                                                                                                                  & \multicolumn{1}{c|}{\checkmark}   &              & \multicolumn{1}{c|}{\checkmark}   & \checkmark   & \multicolumn{1}{c|}{73.54 $\pm$ 2.45}          & \multicolumn{1}{c|}{76.47 $\pm$ 3.08}   & \multicolumn{1}{c|}{73.54 $\pm$ 2.45} & 12.14 $\pm$ 1.05         & \multicolumn{1}{c|}{72.85 $\pm$ 2.92}          & \multicolumn{1}{c|}{75.40 $\pm$ 2.37}   & \multicolumn{1}{c|}{72.85 $\pm$ 2.92} & 21.28 $\pm$ 1.03         \\ \cline{2-13} 
                                                                                                                  & \multicolumn{1}{c|}{\checkmark}   & \checkmark   & \multicolumn{1}{c|}{\checkmark}   &              & \multicolumn{1}{c|}{71.75 $\pm$ 5.38}          & \multicolumn{1}{c|}{75.35 $\pm$ 6.20}   & \multicolumn{1}{c|}{71.75 $\pm$ 5.38} & 12.56 $\pm$ 0.55         & \multicolumn{1}{c|}{71.38 $\pm$ 3.20}          & \multicolumn{1}{c|}{73.34 $\pm$ 3.57}   & \multicolumn{1}{c|}{71.38 $\pm$ 3.20} & 21.40 $\pm$ 0.53         \\ \cline{2-13} 
                                                                                                                  & \multicolumn{1}{c|}{\checkmark}   & \checkmark   & \multicolumn{1}{c|}{\checkmark}   & \checkmark   & \multicolumn{1}{c|}{70.08 $\pm$ 3.10}          & \multicolumn{1}{c|}{74.41 $\pm$ 3.23}   & \multicolumn{1}{c|}{70.08 $\pm$ 3.10} & 12.72 $\pm$ 0.52         & \multicolumn{1}{c|}{71.42 $\pm$ 4.27}          & \multicolumn{1}{c|}{74.97 $\pm$ 3.18}   & \multicolumn{1}{c|}{71.42 $\pm$ 4.27} & 21.89 $\pm$ 0.86         \\ \hline
\multirow{6}{*}{\textsc{\textbf{Mittens}}}                                                                        & \multicolumn{1}{c|}{\checkmark}   &              & \multicolumn{2}{c|}{Branch N/A}                  & \multicolumn{1}{c|}{69.88 $\pm$ 2.28}          & \multicolumn{1}{c|}{72.12 $\pm$ 2.69}   & \multicolumn{1}{c|}{69.88 $\pm$ 2.28} & 10.00 $\pm$ 0.96         & \multicolumn{1}{c|}{70.33 $\pm$ 4.30}          & \multicolumn{1}{c|}{72.70 $\pm$ 4.35}   & \multicolumn{1}{c|}{70.33 $\pm$ 4.30} & 15.59 $\pm$ 1.30         \\ \cline{2-13} 
                                                                                                                  & \multicolumn{1}{c|}{\checkmark}   & \checkmark   & \multicolumn{2}{c|}{Branch N/A}                  & \multicolumn{1}{c|}{69.80 $\pm$ 4.04}          & \multicolumn{1}{c|}{72.30 $\pm$ 4.14}   & \multicolumn{1}{c|}{69.80 $\pm$ 4.04} & 10.49 $\pm$ 0.71         & \multicolumn{1}{c|}{66.38 $\pm$ 3.94}          & \multicolumn{1}{c|}{68.13 $\pm$ 3.63}   & \multicolumn{1}{c|}{66.38 $\pm$ 3.94} & 15.49 $\pm$ 1.61         \\ \cline{2-13} 
                                                                                                                  & \multicolumn{1}{c|}{\checkmark}   &              & \multicolumn{1}{c|}{\checkmark}   &              & \multicolumn{1}{c|}{73.66 $\pm$ 1.98}          & \multicolumn{1}{c|}{74.99 $\pm$ 2.02}   & \multicolumn{1}{c|}{73.66 $\pm$ 1.98} & 13.00 $\pm$ 1.17         & \multicolumn{1}{c|}{74.67 $\pm$ 1.66}          & \multicolumn{1}{c|}{75.94 $\pm$ 1.93}   & \multicolumn{1}{c|}{74.67 $\pm$ 1.66} & 21.30 $\pm$ 0.76         \\ \cline{2-13} 
                                                                                                                  & \multicolumn{1}{c|}{\checkmark}   &              & \multicolumn{1}{c|}{\checkmark}   & \checkmark   & \multicolumn{1}{c|}{74.23 $\pm$ 3.54}          & \multicolumn{1}{c|}{76.06 $\pm$ 3.52}   & \multicolumn{1}{c|}{74.23 $\pm$ 3.54} & 12.50 $\pm$ 0.94         & \multicolumn{1}{c|}{75.20 $\pm$ 3.12}          & \multicolumn{1}{c|}{76.50 $\pm$ 3.09}   & \multicolumn{1}{c|}{75.20 $\pm$ 3.12} & 22.32 $\pm$ 3.30         \\ \cline{2-13} 
                                                                                                                  & \multicolumn{1}{c|}{\checkmark}   & \checkmark   & \multicolumn{1}{c|}{\checkmark}   &              & \multicolumn{1}{c|}{72.89 $\pm$ 2.51}          & \multicolumn{1}{c|}{75.00 $\pm$ 2.68}   & \multicolumn{1}{c|}{72.89 $\pm$ 2.51} & 12.31 $\pm$ 0.65         & \multicolumn{1}{c|}{73.29 $\pm$ 2.15}          & \multicolumn{1}{c|}{75.14 $\pm$ 2.67}   & \multicolumn{1}{c|}{73.29 $\pm$ 2.15} & 22.03 $\pm$ 1.18         \\ \cline{2-13} 
                                                                                                                  & \multicolumn{1}{c|}{\checkmark}   & \checkmark   & \multicolumn{1}{c|}{\checkmark}   & \checkmark   & \multicolumn{1}{c|}{72.93 $\pm$ 3.71}          & \multicolumn{1}{c|}{75.07 $\pm$ 4.21}   & \multicolumn{1}{c|}{72.93 $\pm$ 3.71} & 12.74 $\pm$ 0.79         & \multicolumn{1}{c|}{73.37 $\pm$ 2.55}          & \multicolumn{1}{c|}{75.64 $\pm$ 2.06}   & \multicolumn{1}{c|}{73.37 $\pm$ 2.55} & 22.48 $\pm$ 1.04         \\ \hline
\end{tabular}
}
\end{table*}


\begin{table*}[!htbp]
\centering
\caption{Ablation testing on the Twitter16 dataset~\cite{Ma2017} using [Bi]LSTM as [Bi]RNN units (bold marks the best accuracy w.r.t. word embedding)}
\label{tab:results_lstm_twitter16}
\resizebox{1\textwidth}{!}{%
\begin{tabular}{|c|cc|cc|cccc|cccc|}
\hline
\multicolumn{1}{|l|}{\multirow{3}{*}{\textbf{\begin{tabular}[c]{@{}c@{}}Word\\ Embedding\end{tabular}}}} & \multicolumn{4}{c|}{\textbf{Network Embedding}}  & \multicolumn{4}{c|}{\multirow{2}{*}{\textbf{GRU Cell}}}   & \multicolumn{4}{c|}{\multirow{2}{*}{\textbf{BiGRU Cell}}}                                                                                                                                           \\ \cline{2-5}
\multicolumn{1}{|l|}{}                               & \multicolumn{2}{c|}{\textbf{Text Branch}}     & \multicolumn{2}{c|}{\textbf{Social Branch}}     & \multicolumn{4}{c|}{} & \multicolumn{4}{c|}{}                                                                                                                                       \\ \cline{2-13} 
\multicolumn{1}{|l|}{}                                                                                                                 & \multicolumn{1}{c|}{\textbf{RNN}} & \textbf{CNN} & \multicolumn{1}{c|}{\textbf{RNN}} & \textbf{CNN} & \multicolumn{1}{c|}{\textbf{Accuracy}}         & \multicolumn{1}{c|}{\textbf{Precision}} & \multicolumn{1}{c|}{\textbf{Recall}}  & \multicolumn{1}{c|}{\textbf{Runtime(s)}} & \multicolumn{1}{c|}{\textbf{Accuracy}}         & \multicolumn{1}{c|}{\textbf{Precision}} & \multicolumn{1}{c|}{\textbf{Recall}}  & \multicolumn{1}{c|}{\textbf{Runtime(s)}} \\ \hline
\multirow{6}{*}{\begin{tabular}[c]{@{}c@{}}\textsc{\textbf{Word2Vec}}\\ \textbf{CBOW}\end{tabular}}               & \multicolumn{1}{c|}{\checkmark}   &              & \multicolumn{2}{c|}{Branch N/A}                  & \multicolumn{1}{c|}{75.20 $\pm$ 2.89}          & \multicolumn{1}{c|}{75.72 $\pm$ 2.79}   & \multicolumn{1}{c|}{75.20 $\pm$ 2.89} & 8.56 $\pm$ 0.40          & \multicolumn{1}{c|}{74.43 $\pm$ 2.67}          & \multicolumn{1}{c|}{74.66 $\pm$ 2.68}   & \multicolumn{1}{c|}{74.43 $\pm$ 2.67} & 14.35 $\pm$ 0.93         \\ \cline{2-13} 
                                                                                                                  & \multicolumn{1}{c|}{\checkmark}   & \checkmark   & \multicolumn{2}{c|}{Branch N/A}                  & \multicolumn{1}{c|}{75.08 $\pm$ 2.25}          & \multicolumn{1}{c|}{75.36 $\pm$ 2.12}   & \multicolumn{1}{c|}{75.08 $\pm$ 2.25} & 9.01 $\pm$ 1.17          & \multicolumn{1}{c|}{74.84 $\pm$ 1.59}          & \multicolumn{1}{c|}{75.09 $\pm$ 1.52}   & \multicolumn{1}{c|}{74.84 $\pm$ 1.59} & 14.01 $\pm$ 0.79         \\ \cline{2-13} 
                                                                                                                  & \multicolumn{1}{c|}{\checkmark}   &              & \multicolumn{1}{c|}{\checkmark}   &              & \multicolumn{1}{c|}{76.02 $\pm$ 2.28}          & \multicolumn{1}{c|}{76.57 $\pm$ 2.08}   & \multicolumn{1}{c|}{76.02 $\pm$ 2.28} & 10.89 $\pm$ 0.84         & \multicolumn{1}{c|}{75.37 $\pm$ 2.36}          & \multicolumn{1}{c|}{75.82 $\pm$ 2.26}   & \multicolumn{1}{c|}{75.37 $\pm$ 2.36} & 19.28 $\pm$ 0.78         \\ \cline{2-13} 
                                                                                                                  & \multicolumn{1}{c|}{\checkmark}   &              & \multicolumn{1}{c|}{\checkmark}   & \checkmark   & \multicolumn{1}{c|}{76.79 $\pm$ 3.18}          & \multicolumn{1}{c|}{77.22 $\pm$ 3.08}   & \multicolumn{1}{c|}{76.79 $\pm$ 3.18} & 11.17 $\pm$ 0.83         & \multicolumn{1}{c|}{77.03 $\pm$ 2.14}          & \multicolumn{1}{c|}{77.49 $\pm$ 2.10}   & \multicolumn{1}{c|}{77.03 $\pm$ 2.14} & 19.50 $\pm$ 0.70         \\ \cline{2-13} 
                                                                                                                  & \multicolumn{1}{c|}{\checkmark}   & \checkmark   & \multicolumn{1}{c|}{\checkmark}   &              & \multicolumn{1}{c|}{74.92 $\pm$ 3.04}          & \multicolumn{1}{c|}{75.32 $\pm$ 2.64}   & \multicolumn{1}{c|}{74.92 $\pm$ 3.04} & 11.13 $\pm$ 0.89         & \multicolumn{1}{c|}{75.16 $\pm$ 1.24}          & \multicolumn{1}{c|}{75.32 $\pm$ 1.19}   & \multicolumn{1}{c|}{75.16 $\pm$ 1.24} & 19.41 $\pm$ 0.64         \\ \cline{2-13} 
                                                                                                                  & \multicolumn{1}{c|}{\checkmark}   & \checkmark   & \multicolumn{1}{c|}{\checkmark}   & \checkmark   & \multicolumn{1}{c|}{77.72 $\pm$ 2.29}          & \multicolumn{1}{c|}{78.27 $\pm$ 2.22}   & \multicolumn{1}{c|}{77.72 $\pm$ 2.29} & 11.97 $\pm$ 1.02         & \multicolumn{1}{c|}{75.81 $\pm$ 1.93}          & \multicolumn{1}{c|}{76.51 $\pm$ 1.76}   & \multicolumn{1}{c|}{75.81 $\pm$ 1.93} & 20.27 $\pm$ 1.02         \\ \hline
\multirow{6}{*}{\begin{tabular}[c]{@{}c@{}}\textsc{\textbf{Word2Vec}}\\ \textsc{\textbf{Skip-Gran}}\end{tabular}} & \multicolumn{1}{c|}{\checkmark}   &              & \multicolumn{2}{c|}{Branch N/A}                  & \multicolumn{1}{c|}{73.70 $\pm$ 1.95}          & \multicolumn{1}{c|}{74.06 $\pm$ 2.14}   & \multicolumn{1}{c|}{73.70 $\pm$ 1.95} & 8.26 $\pm$ 1.00          & \multicolumn{1}{c|}{73.41 $\pm$ 2.83}          & \multicolumn{1}{c|}{73.51 $\pm$ 2.95}   & \multicolumn{1}{c|}{73.41 $\pm$ 2.83} & 13.56 $\pm$ 0.96         \\ \cline{2-13} 
                                                                                                                  & \multicolumn{1}{c|}{\checkmark}   & \checkmark   & \multicolumn{2}{c|}{Branch N/A}                  & \multicolumn{1}{c|}{73.37 $\pm$ 2.38}          & \multicolumn{1}{c|}{73.73 $\pm$ 2.71}   & \multicolumn{1}{c|}{73.37 $\pm$ 2.38} & 8.47 $\pm$ 0.73          & \multicolumn{1}{c|}{74.76 $\pm$ 2.05}          & \multicolumn{1}{c|}{74.86 $\pm$ 1.99}   & \multicolumn{1}{c|}{74.76 $\pm$ 2.05} & 13.63 $\pm$ 0.86         \\ \cline{2-13} 
                                                                                                                  & \multicolumn{1}{c|}{\checkmark}   &              & \multicolumn{1}{c|}{\checkmark}   &              & \multicolumn{1}{c|}{74.96 $\pm$ 2.57}          & \multicolumn{1}{c|}{75.04 $\pm$ 2.80}   & \multicolumn{1}{c|}{74.96 $\pm$ 2.57} & 10.55 $\pm$ 0.83         & \multicolumn{1}{c|}{74.51 $\pm$ 1.87}          & \multicolumn{1}{c|}{74.77 $\pm$ 1.95}   & \multicolumn{1}{c|}{74.51 $\pm$ 1.87} & 18.61 $\pm$ 0.86         \\ \cline{2-13} 
                                                                                                                  & \multicolumn{1}{c|}{\checkmark}   &              & \multicolumn{1}{c|}{\checkmark}   & \checkmark   & \multicolumn{1}{c|}{75.16 $\pm$ 2.13}          & \multicolumn{1}{c|}{75.72 $\pm$ 2.37}   & \multicolumn{1}{c|}{75.16 $\pm$ 2.13} & 10.80 $\pm$ 1.08         & \multicolumn{1}{c|}{76.38 $\pm$ 2.70}          & \multicolumn{1}{c|}{76.52 $\pm$ 3.00}   & \multicolumn{1}{c|}{76.38 $\pm$ 2.70} & 19.02 $\pm$ 0.57         \\ \cline{2-13} 
                                                                                                                  & \multicolumn{1}{c|}{\checkmark}   & \checkmark   & \multicolumn{1}{c|}{\checkmark}   &              & \multicolumn{1}{c|}{73.37 $\pm$ 2.42}          & \multicolumn{1}{c|}{73.60 $\pm$ 2.81}   & \multicolumn{1}{c|}{73.37 $\pm$ 2.42} & 10.73 $\pm$ 0.83         & \multicolumn{1}{c|}{74.07 $\pm$ 1.75}          & \multicolumn{1}{c|}{74.21 $\pm$ 1.98}   & \multicolumn{1}{c|}{74.07 $\pm$ 1.75} & 18.66 $\pm$ 0.51         \\ \cline{2-13} 
                                                                                                                  & \multicolumn{1}{c|}{\checkmark}   & \checkmark   & \multicolumn{1}{c|}{\checkmark}   & \checkmark   & \multicolumn{1}{c|}{75.37 $\pm$ 2.81}          & \multicolumn{1}{c|}{75.70 $\pm$ 2.82}   & \multicolumn{1}{c|}{75.37 $\pm$ 2.81} & 11.87 $\pm$ 0.84         & \multicolumn{1}{c|}{75.49 $\pm$ 2.56}          & \multicolumn{1}{c|}{75.87 $\pm$ 2.84}   & \multicolumn{1}{c|}{75.49 $\pm$ 2.56} & 20.02 $\pm$ 1.26         \\ \hline
\multirow{6}{*}{\begin{tabular}[c]{@{}c@{}}\textsc{\textbf{FastText}}\\ \textbf{CBOW}\end{tabular}}               & \multicolumn{1}{c|}{\checkmark}   &              & \multicolumn{2}{c|}{Branch N/A}                  & \multicolumn{1}{c|}{75.98 $\pm$ 2.09}          & \multicolumn{1}{c|}{76.31 $\pm$ 2.13}   & \multicolumn{1}{c|}{75.98 $\pm$ 2.09} & 8.51 $\pm$ 0.65          & \multicolumn{1}{c|}{76.54 $\pm$ 1.69}          & \multicolumn{1}{c|}{76.85 $\pm$ 1.45}   & \multicolumn{1}{c|}{76.54 $\pm$ 1.69} & 14.44 $\pm$ 0.99         \\ \cline{2-13} 
                                                                                                                  & \multicolumn{1}{c|}{\checkmark}   & \checkmark   & \multicolumn{2}{c|}{Branch N/A}                  & \multicolumn{1}{c|}{74.55 $\pm$ 1.90}          & \multicolumn{1}{c|}{74.79 $\pm$ 1.70}   & \multicolumn{1}{c|}{74.55 $\pm$ 1.90} & 8.98 $\pm$ 0.59          & \multicolumn{1}{c|}{74.51 $\pm$ 2.67}          & \multicolumn{1}{c|}{74.67 $\pm$ 2.85}   & \multicolumn{1}{c|}{74.51 $\pm$ 2.67} & 14.25 $\pm$ 0.94         \\ \cline{2-13} 
                                                                                                                  & \multicolumn{1}{c|}{\checkmark}   &              & \multicolumn{1}{c|}{\checkmark}   &              & \multicolumn{1}{c|}{75.93 $\pm$ 1.91}          & \multicolumn{1}{c|}{76.28 $\pm$ 2.12}   & \multicolumn{1}{c|}{75.93 $\pm$ 1.91} & 11.29 $\pm$ 0.87         & \multicolumn{1}{c|}{75.73 $\pm$ 1.43}          & \multicolumn{1}{c|}{76.25 $\pm$ 1.51}   & \multicolumn{1}{c|}{75.73 $\pm$ 1.43} & 19.87 $\pm$ 1.11         \\ \cline{2-13} 
                                                                                                                  & \multicolumn{1}{c|}{\checkmark}   &              & \multicolumn{1}{c|}{\checkmark}   & \checkmark   & \multicolumn{1}{c|}{75.77 $\pm$ 2.48}          & \multicolumn{1}{c|}{76.19 $\pm$ 2.17}   & \multicolumn{1}{c|}{75.77 $\pm$ 2.48} & 11.66 $\pm$ 0.37         & \multicolumn{1}{c|}{77.56 $\pm$ 2.56}          & \multicolumn{1}{c|}{77.98 $\pm$ 2.83}   & \multicolumn{1}{c|}{77.56 $\pm$ 2.56} & 20.44 $\pm$ 0.81         \\ \cline{2-13} 
                                                                                                                  & \multicolumn{1}{c|}{\checkmark}   & \checkmark   & \multicolumn{1}{c|}{\checkmark}   &              & \multicolumn{1}{c|}{74.59 $\pm$ 1.15}          & \multicolumn{1}{c|}{74.81 $\pm$ 1.16}   & \multicolumn{1}{c|}{74.59 $\pm$ 1.15} & 12.03 $\pm$ 0.73         & \multicolumn{1}{c|}{75.49 $\pm$ 1.44}          & \multicolumn{1}{c|}{75.66 $\pm$ 1.40}   & \multicolumn{1}{c|}{75.49 $\pm$ 1.44} & 19.66 $\pm$ 0.97         \\ \cline{2-13} 
                                                                                                                  & \multicolumn{1}{c|}{\checkmark}   & \checkmark   & \multicolumn{1}{c|}{\checkmark}   & \checkmark   & \multicolumn{1}{c|}{\resultlowclr{\textbf{77.95 $\pm$ 1.93}}} & \multicolumn{1}{c|}{78.54 $\pm$ 2.02}   & \multicolumn{1}{c|}{77.95 $\pm$ 1.93} & 12.11 $\pm$ 0.66         & \multicolumn{1}{c|}{\resultlowclr{\textbf{77.68 $\pm$ 1.76}}} & \multicolumn{1}{c|}{78.03 $\pm$ 1.47}   & \multicolumn{1}{c|}{77.68 $\pm$ 1.76} & 20.77 $\pm$ 1.12         \\ \hline
\multirow{6}{*}{\begin{tabular}[c]{@{}c@{}}\textsc{\textbf{FastText}}\\ \textsc{\textbf{Skip-Gran}}\end{tabular}} & \multicolumn{1}{c|}{\checkmark}   &              & \multicolumn{2}{c|}{Branch N/A}                  & \multicolumn{1}{c|}{73.78 $\pm$ 1.98}          & \multicolumn{1}{c|}{73.98 $\pm$ 2.39}   & \multicolumn{1}{c|}{73.78 $\pm$ 1.98} & 8.04 $\pm$ 0.68          & \multicolumn{1}{c|}{73.58 $\pm$ 3.01}          & \multicolumn{1}{c|}{74.13 $\pm$ 2.99}   & \multicolumn{1}{c|}{73.58 $\pm$ 3.01} & 13.27 $\pm$ 0.52         \\ \cline{2-13} 
                                                                                                                  & \multicolumn{1}{c|}{\checkmark}   & \checkmark   & \multicolumn{2}{c|}{Branch N/A}                  & \multicolumn{1}{c|}{74.19 $\pm$ 2.17}          & \multicolumn{1}{c|}{74.57 $\pm$ 2.24}   & \multicolumn{1}{c|}{74.19 $\pm$ 2.17} & 8.48 $\pm$ 0.97          & \multicolumn{1}{c|}{73.17 $\pm$ 1.91}          & \multicolumn{1}{c|}{73.40 $\pm$ 1.99}   & \multicolumn{1}{c|}{73.17 $\pm$ 1.91} & 13.48 $\pm$ 0.44         \\ \cline{2-13} 
                                                                                                                  & \multicolumn{1}{c|}{\checkmark}   &              & \multicolumn{1}{c|}{\checkmark}   &              & \multicolumn{1}{c|}{74.47 $\pm$ 2.35}          & \multicolumn{1}{c|}{75.14 $\pm$ 2.19}   & \multicolumn{1}{c|}{74.47 $\pm$ 2.35} & 11.01 $\pm$ 0.62         & \multicolumn{1}{c|}{74.35 $\pm$ 1.79}          & \multicolumn{1}{c|}{74.67 $\pm$ 1.78}   & \multicolumn{1}{c|}{74.35 $\pm$ 1.79} & 18.26 $\pm$ 0.60         \\ \cline{2-13} 
                                                                                                                  & \multicolumn{1}{c|}{\checkmark}   &              & \multicolumn{1}{c|}{\checkmark}   & \checkmark   & \multicolumn{1}{c|}{75.37 $\pm$ 2.45}          & \multicolumn{1}{c|}{75.77 $\pm$ 2.59}   & \multicolumn{1}{c|}{75.37 $\pm$ 2.45} & 11.10 $\pm$ 0.71         & \multicolumn{1}{c|}{75.00 $\pm$ 1.93}          & \multicolumn{1}{c|}{75.22 $\pm$ 2.02}   & \multicolumn{1}{c|}{75.00 $\pm$ 1.93} & 19.21 $\pm$ 0.79         \\ \cline{2-13} 
                                                                                                                  & \multicolumn{1}{c|}{\checkmark}   & \checkmark   & \multicolumn{1}{c|}{\checkmark}   &              & \multicolumn{1}{c|}{75.08 $\pm$ 2.17}          & \multicolumn{1}{c|}{75.44 $\pm$ 2.08}   & \multicolumn{1}{c|}{75.08 $\pm$ 2.17} & 11.27 $\pm$ 0.85         & \multicolumn{1}{c|}{73.01 $\pm$ 1.71}          & \multicolumn{1}{c|}{73.30 $\pm$ 1.77}   & \multicolumn{1}{c|}{73.01 $\pm$ 1.71} & 18.57 $\pm$ 0.47         \\ \cline{2-13} 
                                                                                                                  & \multicolumn{1}{c|}{\checkmark}   & \checkmark   & \multicolumn{1}{c|}{\checkmark}   & \checkmark   & \multicolumn{1}{c|}{76.99 $\pm$ 1.92}          & \multicolumn{1}{c|}{77.11 $\pm$ 1.95}   & \multicolumn{1}{c|}{76.99 $\pm$ 1.92} & 12.42 $\pm$ 0.60         & \multicolumn{1}{c|}{75.65 $\pm$ 2.01}          & \multicolumn{1}{c|}{75.98 $\pm$ 2.05}   & \multicolumn{1}{c|}{75.65 $\pm$ 2.01} & 20.90 $\pm$ 0.91         \\ \hline
\multirow{6}{*}{\textsc{\textbf{GloVe}}}                                                                          & \multicolumn{1}{c|}{\checkmark}   &              & \multicolumn{2}{c|}{Branch N/A}                  & \multicolumn{1}{c|}{72.89 $\pm$ 3.87}          & \multicolumn{1}{c|}{75.85 $\pm$ 4.27}   & \multicolumn{1}{c|}{72.89 $\pm$ 3.87} & 10.09 $\pm$ 0.86         & \multicolumn{1}{c|}{74.31 $\pm$ 3.26}          & \multicolumn{1}{c|}{77.05 $\pm$ 3.42}   & \multicolumn{1}{c|}{74.31 $\pm$ 3.26} & 17.42 $\pm$ 1.68         \\ \cline{2-13} 
                                                                                                                  & \multicolumn{1}{c|}{\checkmark}   & \checkmark   & \multicolumn{2}{c|}{Branch N/A}                  & \multicolumn{1}{c|}{71.34 $\pm$ 3.64}          & \multicolumn{1}{c|}{73.49 $\pm$ 3.68}   & \multicolumn{1}{c|}{71.34 $\pm$ 3.64} & 9.97 $\pm$ 0.56          & \multicolumn{1}{c|}{73.33 $\pm$ 4.43}          & \multicolumn{1}{c|}{75.58 $\pm$ 5.08}   & \multicolumn{1}{c|}{73.33 $\pm$ 4.43} & 17.18 $\pm$ 0.82         \\ \cline{2-13} 
                                                                                                                  & \multicolumn{1}{c|}{\checkmark}   &              & \multicolumn{1}{c|}{\checkmark}   &              & \multicolumn{1}{c|}{75.53 $\pm$ 3.79}          & \multicolumn{1}{c|}{77.95 $\pm$ 3.45}   & \multicolumn{1}{c|}{75.53 $\pm$ 3.79} & 12.18 $\pm$ 0.74         & \multicolumn{1}{c|}{74.31 $\pm$ 3.89}          & \multicolumn{1}{c|}{75.87 $\pm$ 4.29}   & \multicolumn{1}{c|}{74.31 $\pm$ 3.89} & 22.34 $\pm$ 0.99         \\ \cline{2-13} 
                                                                                                                  & \multicolumn{1}{c|}{\checkmark}   &              & \multicolumn{1}{c|}{\checkmark}   & \checkmark   & \multicolumn{1}{c|}{75.33 $\pm$ 3.44}          & \multicolumn{1}{c|}{78.71 $\pm$ 2.61}   & \multicolumn{1}{c|}{75.33 $\pm$ 3.44} & 12.28 $\pm$ 0.67         & \multicolumn{1}{c|}{75.61 $\pm$ 4.27}          & \multicolumn{1}{c|}{77.85 $\pm$ 4.01}   & \multicolumn{1}{c|}{75.61 $\pm$ 4.27} & 22.50 $\pm$ 1.47         \\ \cline{2-13} 
                                                                                                                  & \multicolumn{1}{c|}{\checkmark}   & \checkmark   & \multicolumn{1}{c|}{\checkmark}   &              & \multicolumn{1}{c|}{73.90 $\pm$ 3.89}          & \multicolumn{1}{c|}{75.66 $\pm$ 4.58}   & \multicolumn{1}{c|}{73.90 $\pm$ 3.89} & 13.10 $\pm$ 1.22         & \multicolumn{1}{c|}{73.46 $\pm$ 3.75}          & \multicolumn{1}{c|}{75.94 $\pm$ 3.53}   & \multicolumn{1}{c|}{73.46 $\pm$ 3.75} & 23.63 $\pm$ 2.32         \\ \cline{2-13} 
                                                                                                                  & \multicolumn{1}{c|}{\checkmark}   & \checkmark   & \multicolumn{1}{c|}{\checkmark}   & \checkmark   & \multicolumn{1}{c|}{72.48 $\pm$ 3.10}          & \multicolumn{1}{c|}{75.47 $\pm$ 2.85}   & \multicolumn{1}{c|}{72.48 $\pm$ 3.10} & 12.47 $\pm$ 0.39         & \multicolumn{1}{c|}{71.46 $\pm$ 2.71}          & \multicolumn{1}{c|}{75.05 $\pm$ 3.12}   & \multicolumn{1}{c|}{71.46 $\pm$ 2.71} & 22.43 $\pm$ 0.89         \\ \hline
\multirow{6}{*}{\textsc{\textbf{Mittens}}}                                                                        & \multicolumn{1}{c|}{\checkmark}   &              & \multicolumn{2}{c|}{Branch N/A}                  & \multicolumn{1}{c|}{73.05 $\pm$ 2.74}          & \multicolumn{1}{c|}{75.22 $\pm$ 1.60}   & \multicolumn{1}{c|}{73.05 $\pm$ 2.74} & 10.35 $\pm$ 1.77         & \multicolumn{1}{c|}{73.46 $\pm$ 3.41}          & \multicolumn{1}{c|}{74.47 $\pm$ 3.26}   & \multicolumn{1}{c|}{73.46 $\pm$ 3.41} & 16.82 $\pm$ 0.96         \\ \cline{2-13} 
                                                                                                                  & \multicolumn{1}{c|}{\checkmark}   & \checkmark   & \multicolumn{2}{c|}{Branch N/A}                  & \multicolumn{1}{c|}{74.31 $\pm$ 2.28}          & \multicolumn{1}{c|}{75.65 $\pm$ 2.09}   & \multicolumn{1}{c|}{74.31 $\pm$ 2.28} & 10.39 $\pm$ 0.59         & \multicolumn{1}{c|}{72.60 $\pm$ 2.74}          & \multicolumn{1}{c|}{73.75 $\pm$ 2.31}   & \multicolumn{1}{c|}{72.60 $\pm$ 2.74} & 18.07 $\pm$ 1.56         \\ \cline{2-13} 
                                                                                                                  & \multicolumn{1}{c|}{\checkmark}   &              & \multicolumn{1}{c|}{\checkmark}   &              & \multicolumn{1}{c|}{75.12 $\pm$ 2.65}          & \multicolumn{1}{c|}{76.56 $\pm$ 2.81}   & \multicolumn{1}{c|}{75.12 $\pm$ 2.65} & 12.50 $\pm$ 0.52         & \multicolumn{1}{c|}{75.53 $\pm$ 2.66}          & \multicolumn{1}{c|}{76.57 $\pm$ 2.44}   & \multicolumn{1}{c|}{75.53 $\pm$ 2.66} & 23.31 $\pm$ 2.10         \\ \cline{2-13} 
                                                                                                                  & \multicolumn{1}{c|}{\checkmark}   &              & \multicolumn{1}{c|}{\checkmark}   & \checkmark   & \multicolumn{1}{c|}{76.75 $\pm$ 1.87}          & \multicolumn{1}{c|}{78.68 $\pm$ 1.68}   & \multicolumn{1}{c|}{76.75 $\pm$ 1.87} & 12.70 $\pm$ 0.62         & \multicolumn{1}{c|}{75.53 $\pm$ 4.11}          & \multicolumn{1}{c|}{76.92 $\pm$ 3.59}   & \multicolumn{1}{c|}{75.53 $\pm$ 4.11} & 23.00 $\pm$ 1.54         \\ \cline{2-13} 
                                                                                                                  & \multicolumn{1}{c|}{\checkmark}   & \checkmark   & \multicolumn{1}{c|}{\checkmark}   &              & \multicolumn{1}{c|}{74.23 $\pm$ 2.85}          & \multicolumn{1}{c|}{75.54 $\pm$ 2.73}   & \multicolumn{1}{c|}{74.23 $\pm$ 2.85} & 13.20 $\pm$ 1.74         & \multicolumn{1}{c|}{73.94 $\pm$ 2.18}          & \multicolumn{1}{c|}{75.13 $\pm$ 2.03}   & \multicolumn{1}{c|}{73.94 $\pm$ 2.18} & 22.85 $\pm$ 0.73         \\ \cline{2-13} 
                                                                                                                  & \multicolumn{1}{c|}{\checkmark}   & \checkmark   & \multicolumn{1}{c|}{\checkmark}   & \checkmark   & \multicolumn{1}{c|}{74.84 $\pm$ 2.51}          & \multicolumn{1}{c|}{76.56 $\pm$ 3.02}   & \multicolumn{1}{c|}{74.84 $\pm$ 2.51} & 13.65 $\pm$ 1.21         & \multicolumn{1}{c|}{74.59 $\pm$ 2.18}          & \multicolumn{1}{c|}{77.07 $\pm$ 2.60}   & \multicolumn{1}{c|}{74.59 $\pm$ 2.18} & 23.87 $\pm$ 1.66         \\ \hline
\end{tabular}
}
\end{table*}

Regardless of the [Bi]RNN cell employed, the models that obtain the best results use a [Bi]RNN layer followed by a CNN layer for both Text and Social Branches. 
The overall best accuracy scores are 77.40\%, 77.15\%, 77.95\%, 77.68\% when using GRU, BiGRU, LSTM, and BiLSTM as [Bi]RNN cells, respectively.
For this dataset, the overall best performing model uses LSTM as RNN cells.

\subsection{Scalability}

We observe that the runtime performance (Tables \ref{tab:results_gru_buzzface} to~\ref{tab:results_lstm_twitter16}) of the models scales with the size of the dataset, length of the documents, size of the vocabulary, [Bi]RNN cells employed, and the number of layers.

The models converge faster to a result depending on the size of the dataset, length of the documents, and size of the vocabulary.
Thus, the models trained on Twitter16 converge with approximately 40\% faster than the ones trained on BuzzFace w.r.t. the word embedding and [Bi]RNN cells employed.

The models that employ RNN cells are faster than the ones that employ BiRNN cells.
Also, as LSTM is a more complex cell than GRU, the models that employ [Bi]LSTM cells are slower that the ones that employ [Bi]GRU cells.

Regardless of the word embedding and dataset, we observe that the models that have more layers converge slower than the models with fewer layers.

\subsection{Comparison with state-of-the-art models.}

To show the efficiency of our proposed  solution, we compare our results with state-of-the-art models from the literature. 
For BuzzFace, an in-depth analysis of detection models is presented in \cite{Santia2019}, where the best performing models are: 
(1) Multinomial Naïve Bayes with TF-IDF which obtains a precision score of 70.63\% and a recall score of 49.66\%; and
(2) Random Forest with extracted feature (i.e., average response time, average comment length, innovation rate, maximum daily comments, number of links, thread deviation) obtain a precision score of 76.83\% and a recall score of 50.83\%.
When using only the text features, in \cite{Ozbay2020} the best performing model is J48 with the accuracy, precision, and recall scores of 65.5\%, 65.5\%, and 68.1\%, respectively.
We obtain the overall best results on BuzzFace (i.e., accuracy, precision, and recall scores of 79.74\%, 74.57\%, and 79.74\%, respectively) with the model that employs an LSTM layer for the Text Branch and an LSTM layer followed by a CNN layer for the Social Branch with \textsc{Mittens} word embeddings. 

On the Twitter datasets the best performing model is TD-RvNN \cite{Ma2017} which obtains an accuracy score of 72.3\% on Twitter15 and 73.7\% on Twitter16.
We obtain the overall best results on Twitter15 (i.e., accuracy, precision, and recall scores of 77.90\%, 78.56\%, and 77.90\%, respectively) with the model that employs an LSTM layer for the Text Branch and an LSTM layer followed by a CNN layer for the Social Branch with \textsc{Mittens} word embeddings. 
For Twitter16, we obtain the overall best results (i.e., accuracy, precision, and recall scores of 77.95\%, 78.54\%, and 77.95\%, respectively) with the model that employs for both the Text and Social Branches an LSTM layer followed by a CNN layer with \textsc{FastText CBOW} word embeddings. 

\section{Discussion and Limitations}\label{sec:discussion}

The ablation testing shows that we obtain the best performance with the Network Embedding that employs \textsc{Mittens}, a domain-specific embedding that leverages corpus specific global information, for medium to large dataset, i.e., BuzzFace and Twitter15.
For small size datasets, i.e., Twitter16, \textsc{FastText}, a character n-gram model, manages to learn better word embeddings by considering sub-words. 
DANES learns to discriminate based on these complex and specific features using the novel proposed Network Embedding.
Besides, we observe that the Feed Forward layers used when creating the Network Embeddings perform slightly better than the Bidirectional ones.
As our results are not conclusive, we advise using a data-driven approach to decide on which configuration of DANES to use.
Further, our experiments show that the best configuration of DANES to create the novel Network Embedding and to detect with high accuracy the veracity of news articles: 
\begin{itemize}
    \item[\textit{(1)}] uses only the [Bi]LSTM or the [Bi]GRU on the Text Branch for medium to large datasets;
    \item[\textit{(2)}] uses the [Bi]LSTM or the [Bi]GRU layer followed by a CNN layer on the Text Branch for small datasets;
    \item[\textit{(3)}] employs a CNN layer for the Social Branch regardless of the [Bi]RNN cells used by the previous layer. 
\end{itemize}

DANES impact is four-fold.
From a social perspective, DANES is an automatic model for fake news detection that helps users to make informed decisions in near real-time regarding the social media content that they consume.
From a research perspective, DANES is a novel supervised model that creates a new Network Embedding which considers both the textual content and the social context.
Also from a research perspective, DANES is a novel supervised model that uses both social and textual context to combat online misinformation.
Finally, from an education perspective, DANES can be used to engage users and make them aware of the content they are consuming and the users they are following, as well as increase their media literacy.

The limitations of this study lay in the datasets used.
This is also a shortcoming in the current literature, as other datasets for which we could collect the social context data are no longer available on the original platforms where the posts were present. 
For example, the FakeNewsNet~\cite{Shu2017} dataset only provides tweets' ids, most of which have been removed from Twitter, and hence cannot be collected.
Overall, such datasets are small in terms of the number of records and may become obsolete, as they provide hyperlinks rather than the posts' text, a limitation also discussed in~\cite{Nguyen2020}.

We conclude the discussion section by answering one final question: \textit{Does social and textual context solve the problem of fake news detection?} The short answer is \textit{No}. Although the proposed \textit{DANES} architecture shows promising results in terms of accuracy when taking into account both social and textual context, there are still a lot of challenges that should be tackled to help users make informed decisions when reading social networks' news feeds. The main challenge that we identified while surveying the current literature is the need for a shift from a Deep/Machine Learning-centric perspective to a data-centric perspective consisting of:
\begin{itemize}
    \item[(\textit{1})] Multi-modal datasets that create richer feature representations that encapsulate multiple perspectives, i.e., visual, audio, information diffusion, network immunization, etc.;
    \item[(\textit{2})] Stable dataset which stores all the metadata information rather than (unstable) URLs and unique IDs;
    \item[(\textit{3})] Source verification and author credibility to improve the feature selection and weighting~\cite{parikh2018media}.
\end{itemize}
Most of these shortcomings are due to a lack of complete, high-quality, human-curated, and well-labeled (large) corpora and the time-consuming work that is required to obtain such data.

Furthermore, we want to emphasize that no approach for fake news detection and mitigation actually solves the overall problem.
Each approach in the current literature solves parts of the problem.
In this paper, we improve fake news detection by introducing both social and textual context into our model, an approach never tried before.
Although our experimental results are significant for the current research in the field of fake news detection and show a novel take for solving this task, the problem remains open.

\section{Conclusions}\label{sec:conclusions}

We introduced DANES, a new hybrid approach for fake news detection that considers both textual content and social context using a novel Network Embedding.
Based on the experimental results and the comparison with the results obtained by the state-of-the-art models, we can draw the following conclusions.

Firstly, the novel Network Embedding that employs domain-specific word embeddings as input for the Text Branch, i.e., \textsc{Mittens}, better encodes medium to large size corpora of textual data than generic pre-trained word embeddings, while the Network Embedding that uses word embeddings that employ character n-grams based on sub-word as input for the Text Branch, i.e., \textsc{FastText}, works better on small to large size datasets.
Secondly, the social context embeddings manage to encode the overall user interaction within the social network improving the performance of the proposed Network Embedding.
Put together to create a new Network Embedding rather than taken separately, the text content and the social context embeddings manage to improve the detection of fake news  --- answering \textit{RQ1}.
Thirdly, DANES obtains good results on small training datasets and outperform current state-of-the-art models --- answering \textit{RQ2}.
Finally, we should consider a data-driven approach when deciding whether or not to employ a Bidirectional layer or GRU over LSTM.

In the future, we plan to mitigate the current shortcoming regarding the dataset size and obsolescence due to broken hyperlinks. That is, we aim to collect and create a new multi-modal social contextual dataset.
Furthermore, we aim to integrate into our solution and analyze the different combinations of linguistic features, i.e., sentiment polarity, text subjectivity, etc., as well as complexity, stylistic, and psychology features~\cite{Horne2017} and global features, i.e., topic modeling, to determine if the accuracy of fake news detection improves.

\bibliographystyle{plainnat}  
\bibliography{main}

\end{document}